\documentclass[pmlr,twocolumn,10pt]{jmlr} 

\usepackage{booktabs}
\usepackage{amsmath}
\usepackage{algpseudocode}
\usepackage{algorithm}
\usepackage{multirow} 
\usepackage{xcolor}
\usepackage[utf8]{inputenc}
\usepackage{enumitem}

\usepackage{siunitx}

\algnewcommand\algorithmicforeach{\textbf{for each}}
\algdef{S}[FOR]{ForEach}[1]{\algorithmicforeach\ #1\ \algorithmicdo}

\algnewcommand\algorithmicnot{\textbf{not}}
\algdef{SE}[IF]{IfNot}{EndIf}[1]{\algorithmicif\ \algorithmicnot\ #1\ \algorithmicthen}{\algorithmicend\ \algorithmicif}%

\newcommand{\hi}{hi}
\newcommand{\fl}{fl}
\newcommand{\Enct}{\mathtt{Enc}^{t}}
\newcommand{\Ence}{\mathtt{Enc}^{e}}
\newcommand{\Encstruc}{\mathtt{Enc}_{struc}}
\newcommand{\Decstruc}{\mathtt{Dec}_{struc}}
\newcommand{\Embhi}{\textbf{E}_{\hi}}
\newcommand{\Embfl}{\textbf{E}_{\fl}}
\newcommand{\Enchi}{\mathtt{Enc}_{\hi}}
\newcommand{\Encfl}{\mathtt{Enc}_{\fl}}

\newcommand{\equal}[1]{{\hypersetup{linkcolor=black}\thanks{#1}}}

\theorembodyfont{\upshape}
\theoremheaderfont{\scshape}
\theorempostheader{:}
\theoremsep{\newline}

\jmlrvolume{LEAVE UNSET}
\jmlryear{2023}
\jmlrsubmitted{LEAVE UNSET}
\jmlrpublished{LEAVE UNSET}
\jmlrworkshop{Conference on Health, Inference, and Learning (CHIL) 2023} 

\title{Rediscovery of CNN's Versatility\\for Text-based Encoding of Raw Electronic Health Records}

\author{%
  \Name{Eunbyeol Cho}\equal{These authors contributed equally} \Email{eunbyeol.cho@kaist.ac.kr}\\
  \Name{Minjae Lee}\footnotemark[1] \Email{mjbooo@kaist.ac.kr}\\
  \Name{Kyunghoon Hur} \Email{pacesun@kaist.ac.kr}\\
  \Name{Jiyoun Kim} \Email{jiyoun.kim@kaist.ac.kr}\\
  \addr KAIST, Republic of Korea
  \AND
  \Name{Jinsung Yoon} \Email{jinsungyoon@google.com}\\
  \addr Google Cloud AI Research, USA
  \AND
  \Name{Edward Choi} \Email{edwardchoi@kaist.ac.kr}\\
  \addr KAIST, Republic of Korea
  }

\begin{document}

\maketitle

\begin{abstract}
Making the most use of abundant information in electronic health records (EHR) is rapidly becoming an important topic in the medical domain.
Recent work presented a promising framework that embeds entire features in raw EHR data regardless of its form and medical code standards.
The framework, however, only focuses on encoding EHR with minimal preprocessing and fails to consider how to learn efficient EHR representation in terms of computation and memory usage.
In this paper, we search for a versatile encoder not only reducing the large data into a manageable size but also well preserving the core information of patients to perform diverse clinical tasks. 
We found that hierarchically structured Convolutional Neural Network (CNN) often outperforms the state-of-the-art model on diverse tasks such as reconstruction, prediction, and generation, even with fewer parameters and less training time.
Moreover, it turns out that making use of the inherent hierarchy of EHR data can boost the performance of any kind of backbone models and clinical tasks performed.
Through extensive experiments, we present concrete evidence to generalize our research findings into real-world practice.
We give a clear guideline on building the encoder based on the research findings captured while exploring numerous settings.
\end{abstract}

\paragraph*{Data and Code Availability}
This paper uses the MIMIC-III and eICU datasets, which are available on PhysioNet repository \citep{johnson2016mimic, pollard2018eicu}.
The source code is available at the Github repository. \footnote{\url{https://github.com/eunbyeol-cho/versatile-ehr-encoder}}

\paragraph*{Institutional Review Board (IRB)}
This research does not require IRB approval.

\section{Introduction}
\label{sec:intro}
The widespread introduction of electronic health record (EHR) systems brings tremendous opportunities to apply a data-driven approach to the healthcare domain.
Through EHR systems, millions of patients' data are now collected in a systematic manner across diverse healthcare institutions.
Using this rapidly growing EHR dataset, many researchers have found applications such as predicting clinical outcomes, learning representations of cohorts to get medical insights, and synthesizing clinical data.

To perform EHR-related tasks, conventional frameworks employed various encoding architectures such as Recurrent Neural Networks~\cite{lipton2015lstm, edward2015doctorai, rajkomar2018scalable}, Convolutional Neural Networks~\cite{miotto2016deep, nguyen2016deepr, landi2020}, and Transformer-based models~\cite{yoon2022ehr, edward2019gct, shang2020gbert, rasmy2020medbert, li2020behrt, song2018sand}.
Despite the diversity of proposed works, all these attempts have clear limitations in that they are applicable exclusively to their own EHR system.
For example, they cannot use both MIMIC-III~\cite{johnson2016mimic, johnson2016physionet, goldberger2000physiobank} and eICU~\cite{pollard2018eicu, johnson2019eicu} datasets together for training, unless involving manual harmonization of incompatible features.
In addition, inherent disparities in the standards of medical codes and database schemas prevent data from being sourced from multiple healthcare institutions.
None of previous works thus can fully utilize information in multiple diverse EHR systems.

A recent work~\cite{hur2020descemb, hur2022unihpf} presented a text-based encoding approach to address this conventional limitation.
Specifically, \cite{hur2022unihpf} proposed a universal framework that can embed entire features in raw EHR data regardless of its schema and medical code standards.
Moreover, the framework showed comparable, if not better, performance on various tasks, even without relying on domain knowledge-based preprocessing.
However, it comes at a steep price of the embedded data having tens to hundreds of times larger in size than existing methods, even if it covers only a few hours of medical trajectory of a patient in hospital.
Therefore, a structure that efficiently extracts the core information from the large input is needed. 

In this study, we search for a versatile model architecture to encode the raw EHR input into a low-dimensional space under the universal text-based encoding framework for diverse tasks such as reconstruction (\textit{i.e., autoencoding}), prediction, and generation.
Throughout the paper, we make the following contributions:

\begin{itemize}
  \item
  To the best of our knowledge, this is the first work to search for a versatile encoder not only reduces the large raw EHR into a manageable size but also preserves patients' core information to perform diverse clinical tasks.
  \item
  We conduct extensive experiments with multiple variables to tune model architectures and various clinical tasks (\textit{i.e., reconstruction, prediction, generation}).
  Furthermore, by experimenting with two representative datasets in the EHR domain (\textit{i.e., MIMIC-III, eICU}), we present concrete evidence for generalizing our research findings to a wide variety of EHR systems.
  We capture the core tendencies while exploring these numerous settings and systematically summarize the findings to give a clear guideline on building the encoder.
  \item
  The encoder that we found is widely applicable in real-world practice.  
  Even with fewer parameters and less training time, a hierarchically structured CNN often outperforms the state-of-the-art model on widely accepted tasks in the field.
\end{itemize}

\section{Related work}
\paragraph{Feature-selection-based encoder for EHR}
Many researchers have applied a data-driven approach to the healthcare domain.
By making use of EHR datasets, they predict medical outcomes, learn the representation of patients for various downstream tasks, and synthesize medical data.
To perform the clinical tasks, they employed various encoding backbones such as Recurrent Neural Network~\cite{lipton2015lstm, edward2015doctorai, rajkomar2018scalable}, Convolutional Neural Network~\cite{miotto2016deep, nguyen2016deepr, landi2020}, and Transformer~\cite{yoon2022ehr, edward2019gct, shang2020gbert, rasmy2020medbert, li2020behrt, song2018sand}.
However, all such works have invested a lot in preprocessing the raw EHR for standardizing the EHR schema, or feature engineering specifically for the given task.
This is a clear limitation in that extensive human labor and clinical domain knowledge are required to produce satisfactory model performance.

\paragraph{Universal Healthcare Predictive Framework}
Recently,~\cite{hur2022unihpf} presented UniHPF, a universal framework that can embed entire features of raw EHR regardless of schema and medical code standard used in the database.
Specifically, UniHPF views EHR data as pure text and flattened the EHR tables (\textit{e.g.} prescriptions, lab results) to feed them to Transformer-based text encoders.
This text-based approach showed comparable if not better performance compared to conventional approaches on various predictive tasks, even without relying on medical domain knowledge.
Ironically, because it can handle raw EHR data without any preprocessing or feature selection, UniHPF generates extremely long embedded data, given how it encodes entire EHR data in a text-encoding fashion \footnote{
\figureref{apd:tokenhist} shows the extent to which the number of tokens increases over the observation time.}.
This imposes strong computational limits on the framework and necessitates an additional module for compressing the embedded data to a smaller size.

\section{Method}
\label{sec:math}

\begin{figure*}[!h]
\floatconts
  {fig:mainfigure}
  {
  \caption{EHR structure and encoder framework.}
  \label{EHR_structure}
  }
  {%
    \subfigure[Illustration of EHR tables and patient representation]{\label{fig:EHRstructure}
      \includegraphics[width=0.9\textwidth]{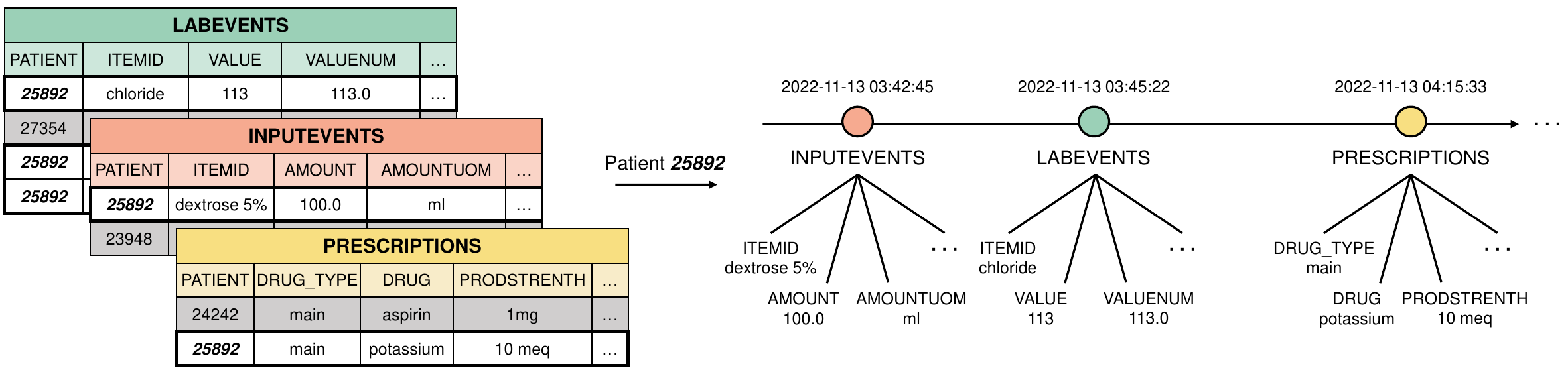}
      }
      
    \subfigure[Encoder framework and examples of hierarchical input $\textbf{x}_{hi}$ and flattened input $\textbf{x}_{fl}$]{\label{fig:EncoderStructure}
      \includegraphics[width=0.9\textwidth]{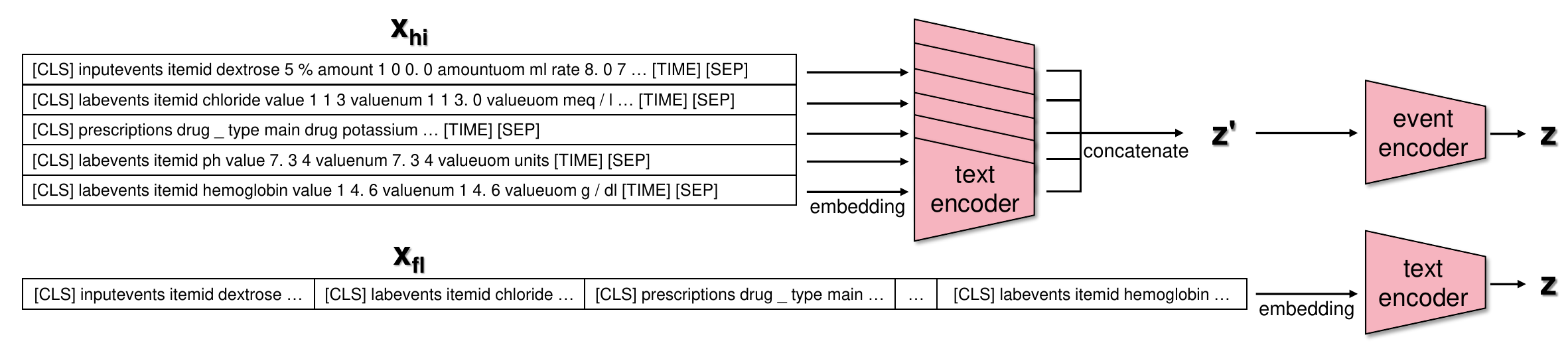}
      }
  }
\end{figure*}

In order to search for a versatile encoder for raw text-based EHR, we examine different encoder designs.
In \sectionref{subsec:EHRSerial}, we first show how raw tabular EHR is converted into natural text, following the embedding strategy of UniHPF.
In \sectionref{subsec:EncodingFramework}, we construct both hierarchical and flattened encoder structures to confirm the effectiveness of reflecting EHR hierarchy. Also, two highly scalable model classes of CNN and Transformer are considered as backbone models \footnote{We do not consider RNNs in this work, as maximum input can be up to 8,000 tokens.}.

As shown in \figureref{fig:Pipeline}, we evaluate the significance of the encoder in downstream tasks such as prediction, reconstruction, and generation. Accordingly, we present decoder architectures for reconstruction and generation in \sectionref{subsec:TaskSpecificModels} and explanations for other task-specific models (e.g., classifier) in \sectionref{sec:expdesign}.

\subsection{EHR serialization}
\label{subsec:EHRSerial}

\paragraph{EHR structure}
Once a patient is admitted to the intensive care unit (ICU), a series of medical events occur during the ICU stay.
Each event is recorded in one of several tables, such as \textit{prescription, diagnosis, lab events} and \textit{input events} in the hospital database. 
Each table consists of multiple rows (e.g., lab events), which in turn consist of multiple columns representing different feature variables (e.g., lab date, lab name, lab value).
The type of cell values can be categorized as textual, numeric, and itemized values (e.g., lab test ID) that can be textualized using a description from the definition table. 
\figureref{fig:EHRstructure} provides an overview of EHR structure. 

\paragraph{EHR serialization}
In order to construct the input data, we first extract each patient's records from the multiple tables in the hospital database and sort all events chronologically.
Following the UniHPF-strategy, we formulate patient representation as
\[\textbf{x} = f(\texttt{column name} \otimes g(\texttt{cell}) \otimes \texttt{timegap})\]
where $f$ is a tokenizer, $g$ is a mapping function that converts the type of cell value to text, $\otimes$ is the concatenate function, and $\texttt{timegap}$ represents the quantized time interval between consecutive events.
In detail, $g$ converts an itemized value to a corresponding free text description (e.g., lab test ID 51385 to ``Atypical Lymphocyte'') and numeric values to text separated by space (e.g., 123.1 to ``1 2 3 . 1'').
For $f$, resulting texts are tokenized into the sub-word level.
In this way, with minimal pre-processing, we construct the patient representation $\textbf{x}$ as a sequence of discrete tokens.

In addition, as shown in \figureref{fig:EHRstructure}, the patient representation has two levels of hierarchy; event and feature levels.
Specifically, it is described by a sequence of events where each event consists of multiple features. 
Making use of this hierarchical relationship between events and their corresponding features, we define hierarchical input $\textbf{x}_{\hi} \in \mathbb{R}^{n_e \times n_{t/e}}$, and flattened input $\textbf{x}_{\fl} \in \mathbb{R}^{n_t}$, where $n_e$, $n_{t/e}$, and $n_t$ are, respectively, the number of events, the number of text tokens per event, and the number of flattened text tokens.

Note that padding and truncation approaches are used to make fixed $n_e$, $n_{t/e}$, and $n_t$.
Since $\textbf{x}_{\fl}$ is an unfolded version of $\textbf{x}_{\hi}$ with padding removed and events concatenated, $n_t$ is less than $n_e \times n_{t/e}$ although $\textbf{x}_{\fl} $ and $\textbf{x}_{\hi}$ contain exactly the same amount of information.
Examples of $\textbf{x}_{\hi}$ and $\textbf{x}_{\fl}$ are shown on the left part in \figureref{fig:EncoderStructure}.

\subsection{Encoding framework}
\label{subsec:EncodingFramework}

\paragraph{Embedding}
Before feeding the input sequence to an encoder, we embed tokens according to the UniHPF embedding approach. Details are specified in \appendixref{apd:AdditionalEmbedding}. 
We denote the hierarchical input embedding by $\textbf{E}_{\hi} = \mathtt{Emb}(\textbf{x}_{\hi}) \in \mathbb{R}^{n_e \times n_{t/e} \times d}$, and the flattened input embedding by $\textbf{E}_{\fl} = \mathtt{Emb}(\textbf{x}_{\fl}) \in \mathbb{R}^{n_{t} \times d}$, where $d$ is the token embedding dimension.

\paragraph{Encoder}
To encode $\textbf{E}_{\hi}$ and $\mathbf{E}_{\fl}$ into latent vector $\textbf{z}$, we construct $\Encstruc$ reflecting the input structure.
\figureref{fig:EncoderStructure} summarizes the overall encoding process.

For hierarchical embedding $\textbf{E}_{\hi}$, we use a two-stage encoder $\mathtt{Enc}_{\hi}$ consisting of a text encoder $\Enct$ for token embedding within an event, followed by an event encoder $\Ence$ for aggregating and compressing encoded events $\textbf{z}'$ into $\textbf{z}$.
Specifically, $\Enct$ takes per-event token embeddings $\{\textbf{e}_{\hi}^i\}_{i=1}^{n_e} \subset \mathbb{R}^{n_{t/e} \times d}$ as input and produces $\textbf{z}' \in \mathbb{R}^{n_e \times d'}$ with the following operation:
\[\textbf{z}' = \mathtt{Concat}(\mathtt{Flatten}([\Enct(\textbf{e}_{\hi}^1),..., \Enct(\textbf{e}_{\hi}^{n_e})]))\]
where $\mathtt{Flatten}$ is an element-wise operation that expands each $\Enct(\textbf{e}_{\hi}^i)$ into a 1D vector with $d'$ dimension and $\mathtt{Concat}$ is a function that concatenates events in a chronological order.
Note that all $\Enct$ are shared across events. 
Afterwards, $\Ence$ compresses $\textbf{z}'$ into a latent vector $\textbf{z} \in \mathbb{R}^{t \times c}$ where $t$ and $c$ are the desired temporal and channel dimension, respectively. 
Such a two-stage process gives the encoder explicit information about the hierarchical structure between events and their features.

For flattened embedding $\textbf{E}_{\fl}$, we use an one-stage encoder $\mathtt{Enc}_{\fl}$ consisting only of $\Enct$, which directly compress $\textbf{E}_{\fl}$ into $\textbf{z}$.

\subsection{Encoding schemes}
\label{subsec:EncodingSchemes}
In this section, we describe how the input embedding 
$\textbf{E}_{\hi}$ or $\textbf{E}_{\fl}$ is compressed into $\textbf{z}$ of a desired dimension.
Note that the encoding strategy is shared for both $\Enct$ and $\Ence$.
Therefore, we can generalize the input and output of each encoder as $\textbf{E}_{input} \in \mathbb{R}^{n \times d}$ with $n$ sequence length and $d$ dimension, and $\textbf{E}_{output} \in \mathbb{R}^{n' \times d'}$ where $n'$ and $d'$ denote the desired length and hidden dimension.

We consider using CNN and Transformer as the backbone of the encoder and develop a custom encoding scheme for each of them.
We examine several Transformer-based encoding schemes summarized in \appendixref{apd:TransfEncoding} and decide on the encoding strategy for each backbone. 
\algorithmref{alg:CNNlayernum} and \ref{alg:CNNlayerorder} summarize the CNN encoding scheme, while \algorithmref{alg:Transformer} summarizes Transformer encoding scheme.
Specific examples of each algorithm are provided in \tableref{tab:algCNN} and \tableref{tab:algTransf}.

\begin{figure}[!t]
    \vspace*{-6mm}
    \includegraphics[width=\linewidth]{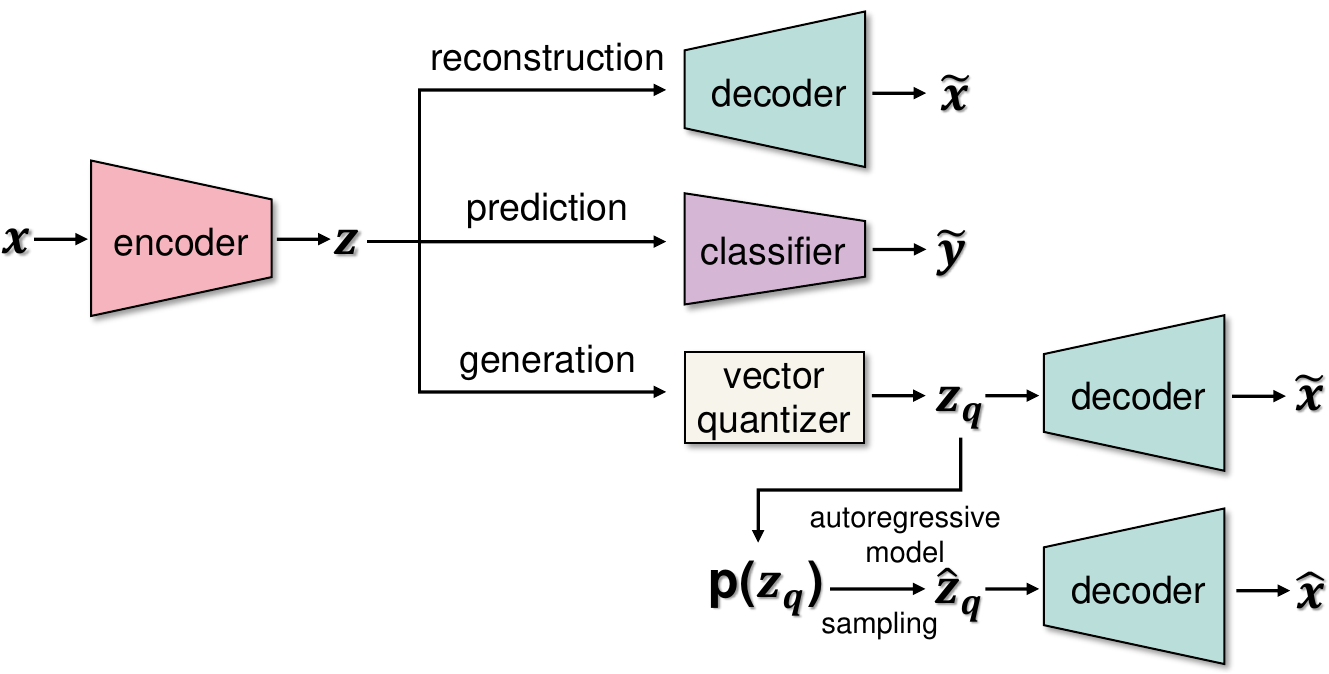}
    \centering
    \caption{Overall pipeline. We search for a versatile encoder design and validate it in three downstream tasks: reconstruction, prediction, and generation. }
    \label{fig:Pipeline}
    \vspace*{-3mm}
\end{figure}

\subsection{Decoder}
\label{subsec:TaskSpecificModels}
We employ CNN-based and Transformer-based decoder architectures for the reconstruction and generation tasks (shown in \figureref{fig:Pipeline}).

The CNN-based decoder is designed with a symmetric structure of the CNN-based encoder.
Specifically, while the encoder layers either compress the dimension in half or leave it unchanged, the decoder layers inversely expand the dimension by two or leave it unchanged, respectively.

For the Transformer-based decoder, we reconstruct the original input $\textbf{x}$ by using cross attention between the latent vector  $\textbf{z}$ and a learnable placeholder embedding with the identical length of $\textbf{x}$.
Specifically, the placeholder embeddings are randomly initialized at first, and while passing through each layer of the decoder, the embeddings are decompressed at the channel level with cross attention applied.
As shown in the decoder of \figureref{fig:transformerdecoder}, the decoder block consists of a Transformer decoder layer in which cross-attention is applied and a linear layer for channel increase.
Between the encoder and decoder, we add linear layers in order to match the latent and placeholder dimensions before applying cross attention. This is because the placeholder embedding dimension is increased with each decoder layer.
An exploration of various Transformer-based decoder architectures can be found in \appendixref{apd:TransfDec}.

\section{Experiments}
\subsection{Dataset description}
\paragraph{Source data}

We experimented with two representative datasets in the EHR domain; MIMIC-III and eICU.
The MIMIC-III dataset comprises deidentified clinical data from more than 40,000 patients admitted to the ICU of Beth Israel Deaconess Medical Center.
The eICU is populated with data from around 140,000 patients who were admitted to a combination of many critical care units throughout the continental United States.
To fully represent patients' medical trajectories, the datasets contain many medical events, such as laboratory test, prescriptions, and input events (e.g., fluid injections) with temporal information.
In accordance with the UniHPF configuration, we employed solely three tables from both datasets - laboratory test, prescriptions, and input events.

\begin{figure}[t]
    \includegraphics[width=\linewidth]
    {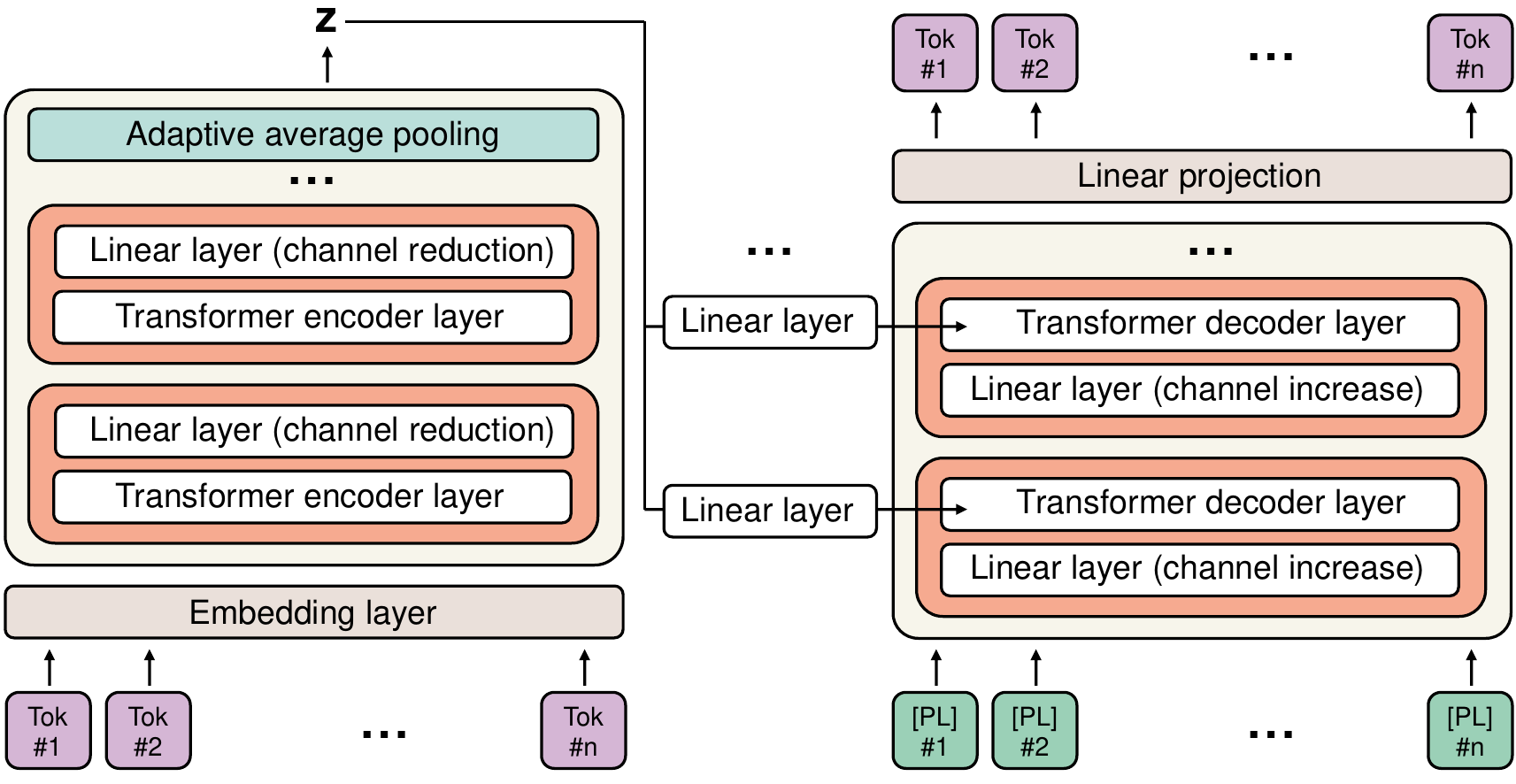}
    \centering
    \caption{Proposed Transformer-based encoder and decoder architecture.}
    \label{fig:transformerdecoder}
    \vspace*{-3mm}
\end{figure}

\paragraph{Cohort definition}
To build cohorts of patients from MIMIC-III and eICU databases, we follow the criteria on which the universal framework~\cite{hur2022unihpf} is based. 
We retrieve records of patients over 18 years old who stayed in the ICU for over 24 hours.
We filter out ICU stays with less than five medical events and use the first 12 hours of events of the first ICU stay for each hospital stay.
For both datasets, we divide patients into separate train, validation, and test sets with a ratio of 8:1:1.

\subsection{Experimental design and model}
\label{sec:expdesign}

We perform three downstream tasks: reconstruction, prediction, and generation.
For each downstream task, we explore multiple settings for the encoder $\Encstruc$ (or additionally decoder $\Decstruc$), where $struc \in \{\hi, \fl\}$.
We use the same $struc$ for both $\Encstruc$ and $\Decstruc$.
Their backbone can either be CNN or Transformer, denoted by $C$ and $T$, respectively.

\paragraph{Reconstruction task}
In order to evaluate the efficacy of the encoder in preserving patient information, we build an autoencoder as follows: 
$$ \mathbf{\Tilde{x}} = \mathtt{Dec}_{struc}(\mathtt{Enc}_{struc}(\mathbf{x}))$$
We train the autoencoder with a cross-entropy loss between $\mathbf{x}$ and $\mathbf{\tilde{x}}$.
We experimented $(\mathtt{Enc}_{struc}, \mathtt{Dec}_{struc})$ on $(C,C), (T,C) \text{ and } (T,T)$ for $\hi$ and $\fl$.
We evaluate the reconstruction performance via token-level accuracy, the ratio of correct tokens to the total number of tokens.

\paragraph{Prediction task}
\label{para:prediction}
To perform clinical outcome prediction, we formulate the task as binary, multi-class or multi-label classification: 
$$\tilde{\textbf{y}}=\mathtt{Classifier}(\mathtt{Enc}_{struc}(\mathbf{x}))$$
The encoder and classifier are trained with the binary cross-entropy loss (cross-entropy loss for multi-class) between the predicted probabilities $\tilde{\textbf{y}}$ and the true label $\textbf{y}$.
For $\mathtt{Classifier}$, we use Transformer due to its permutation invariant nature, rather than a purely MLP-based classifier\footnote{Detailed discussion regarding the choice of classifier architecture is provided in \appendixref{apd:BuildingClassifier}}.
After linearly projecting $\textbf{x}$ into $\textbf{z}$, we pass $\textbf{z}$ through the Transformer layers.
The output is then averaged and linearly mapped to generate logits for each class.
We adopted six clinically meaningful prediction tasks following \cite{hur2022unihpf}; Diagnosis, Mortality, Final acuity Imminent Discharge, Length-Of-Stay for cases of three and seven days.
Further details of each clinical task are provided in the \appendixref{apd:DefClinicTask}.
We evaluated all prediction tasks in terms of AUROC.

\paragraph{Generation task} 
For the unconditional synthesis of textualized EHR $\textbf{x}$, autoregressive modeling of $\textbf{x}$ is challenging due to the high modeling capacity and memory inefficiency required by the considerable sequence length of UniHPF syntax.
Therefore, we use the VQ-VAE~\cite{van2017neural} approach to model the discrete latent space autoregressively. 
The VQ-VAE consists of  $\Encstruc$, $\Decstruc$, a vector quantization layer $\mathtt{VQ}$ and a learnable codebook, formulated as follows:
\begin{align*}
\textbf{z}_q &= \mathtt{VQ}(\Encstruc(\textbf{x})), \
\tilde{\textbf{x}}=\Decstruc(\textbf{z}_q) 
\\ \hat{\textbf{x}} &= \Decstruc(\hat{\textbf{z}}_q) \ \text{where} \ \hat{\textbf{z}}_q \sim p(\textbf{z}_q)
\end{align*}
where $\mathtt{VQ}$ maps each vector $\textbf{z}=\Encstruc(\textbf{x})$ into $\textbf{z}_q$ with the nearest code. 
The VQ-VAE is trained in two steps. 
First, $\Encstruc$, $\Decstruc$, and the codebook are trained simultaneously to minimize the distance between $\textbf{x}$ and $\tilde{\textbf{x}}$ and the distance between $\textbf{z}$ and $\textbf{z}_q$.
Second, the Transformer-based autoregressive model is trained to learn the prior distribution over the discrete latent space $p(\textbf{z}_q)$.
Then the sampled latent code sequence $\hat{\textbf{z}}_q$ is passed into the decoder to synthesize $\hat{\textbf{x}}$.
Further details of training VQ-VAE are in \appendixref{apd:VQVAE}. We conduct experiments on $(C,C)$, $(T,C)$ and $(T,T)$ for $(\mathtt{Enc}_{hi}, \mathtt{Dec}_{hi})$.

We assess the quality of synthetic data quantitatively, qualitatively, and from a privacy perspective.

We propose a metric to measure the preservation of table syntax and semantic consistency by comparing triples (\texttt{table}, \texttt{column name}, \texttt{cell}) in generated data with real data.
For example, the lab ID should be on a lab table, not a prescription table.
On a per-event or per-sample basis, we compute \textbf{RCE} (ratio of correct events to total events), \textbf{RUE} (ratio of correct unique events to total unique events), and \textbf{RCS} (ratio of correct samples to total samples).
Specifics regarding the evaluation algorithm and scoring metrics are listed in \appendixref{apd:SyntheticDataScoring}.

To qualitatively compare the distribution of original and synthetic data, we used t-SNE to visualize latent vectors in a two-dimensional space. 

In addition, for the privacy evaluation, we conducted a membership inference attack~\cite{shokri2017membership}, and the task definition and results are shown in \appendixref{apd:MI}.

\subsection{Implementation details}
For the CNN-based encoder (\algorithmref{alg:CNNlayernum} and \ref{alg:CNNlayerorder}), we employed a scheme that compresses the dimension as desired by combining two types of kernels, one with a kernel size of 5, a stride of 2, and a padding size of 2, and the other with a kernel size of 1 and a stride of 1 and no padding.

For the Transformer-based encoder (\algorithmref{alg:Transformer}), we use four Transformer layers for $\Enchi$ and four Performer~\cite{choromanski2020rethinking} layers for $\Encfl$, in which all layers in both $\Enchi$ and $\Encfl$ use 4 attention heads. 

We use 2 Transformer layers with 4 heads and a hidden dimension of 128 for the $\mathtt{Classifier}$.
As for the Transformer-based autoregressive model used for generation, we utilize 4 Transformer layers with 4 heads and a hidden dimension of 256.
The Adam optimizer \cite{kingma2014adam}, along with a learning rate of 5e-4 (or 5e-5 for models that failed to train), is employed for the optimization process. Additionally, we select different batch sizes of 16, 64, and 32 for reconstruction, prediction, and generation, respectively.
Experiments are conducted with different random seed values, with three seeds for reconstruction and prediction, and two seeds for the generation.

\subsection{Searching range}
For input embeddings $\Embhi \in \mathbb{R}^{n_e \times n_{t/e} \times d}$ and $\Embfl \in \mathbb{R}^{n_{t} \times d}$, we employ $n_e=256$, $n_{t/e}=128$ for $\Embhi$, and $n_t=8192$ for $\Embfl$.
The embedding layer $\mathtt{Emb}$ has a fixed dimension $d=256$.
For the compressed latent vector $\textbf{z} \in \mathbb{R}^{t \times c}$ of $\Embhi$ and $\Embfl$, we define the latent dimension $l = t \times c$.

For prediction and reconstruction tasks, we explore its size starting from $l$ = 256 and double it up until $l$ = 4096.
For each $l$, represented as $2^{2i-1}$ or $2^{2i}$, we search five possible cases of $t$ from $2^{i-2}$ to $2^{i+2}$ by increasing $i$.
(\textit{e.g.,} For the latent vector $\textbf{z}$ having $l=2048=2^{2*6-1}$, we search $t$ from $2^{4}$ to $2^{8}$).

However, for the generation task, we use a less compressed searching range of $l$, starting from 4096 up to 32768.
This is because the first stage in VQVAE can be viewed as conducting reconstruction along with vector quantization, in which additional information loss is inevitable.

Compression rate, indicating how many times $\textbf{z}$ is smaller than $\Embhi$ or $\Embfl$, can be calculated by $\frac{n_e\times n_{t/e}\times d}{l}$ and $\frac{n_t\times d}{l}$, respectively.
(\textit{e.g.,} if $\textbf{z}$ has a shape of $(256, 8)$ in hierarchical case, the compression rate is ${\times}4096 = \frac{256 \times 128 \times 256}{2048}$).
Consequently, $\Enchi$ has a compression rate ranging from 2048 to 32768, while $\Encfl$ has a lower rate ranging from 512 to 8192.

\begin{figure*}[!ht]
\vspace*{-3mm}
\floatconts
  {fig:comparison_recon_mimic_eicu}
  {\caption{
    Comparison of reconstruction performance for both MIMIC-III (left) and eICU (right) datasets. 
    }
  \label{recon_summary}
  }
  {
    \subfigure[Reconstruction performances arranged by the latent dimension $l$]{\label{recon_summary_latent}
      \includegraphics[width=0.85\textwidth]{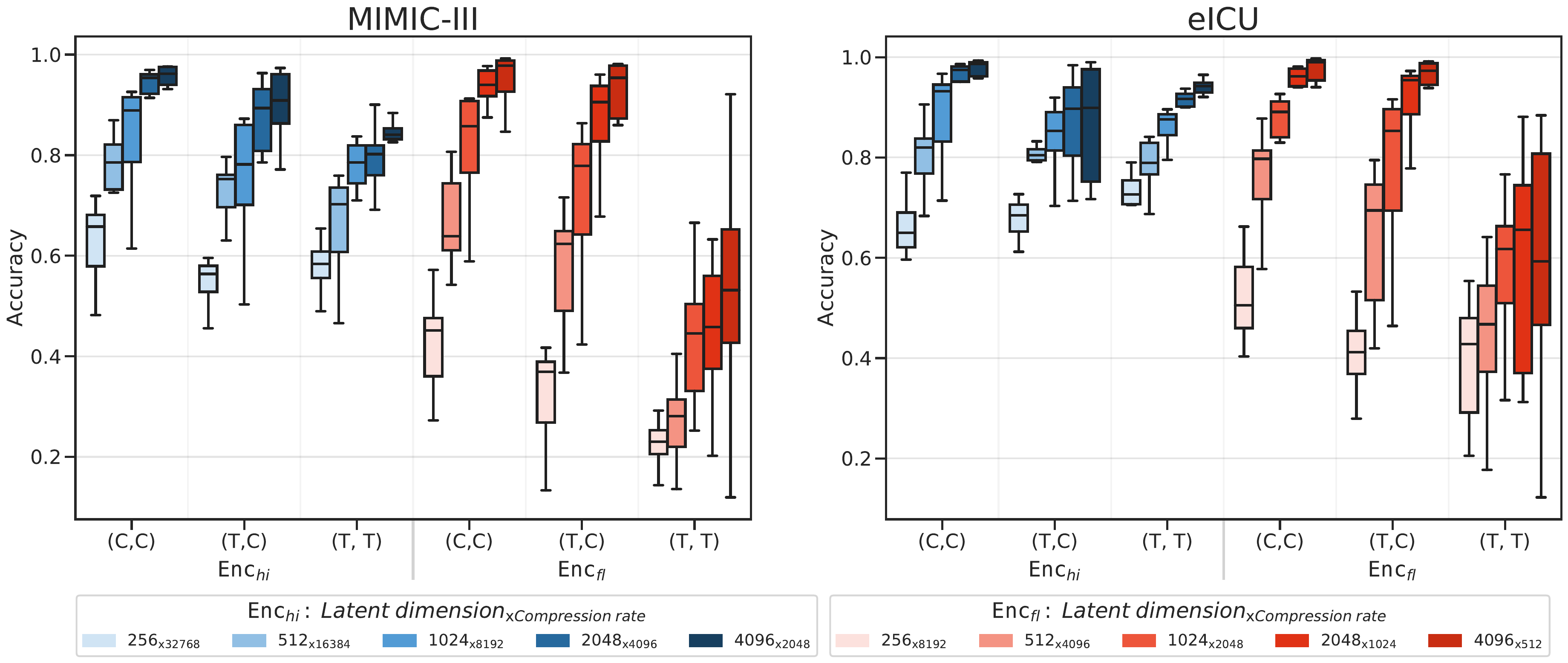}
      }
    \subfigure[Reconstruction performances arranged by the temporal dimension $t$]{\label{recon_summary_temporal}%
      \includegraphics[width=0.85\textwidth]{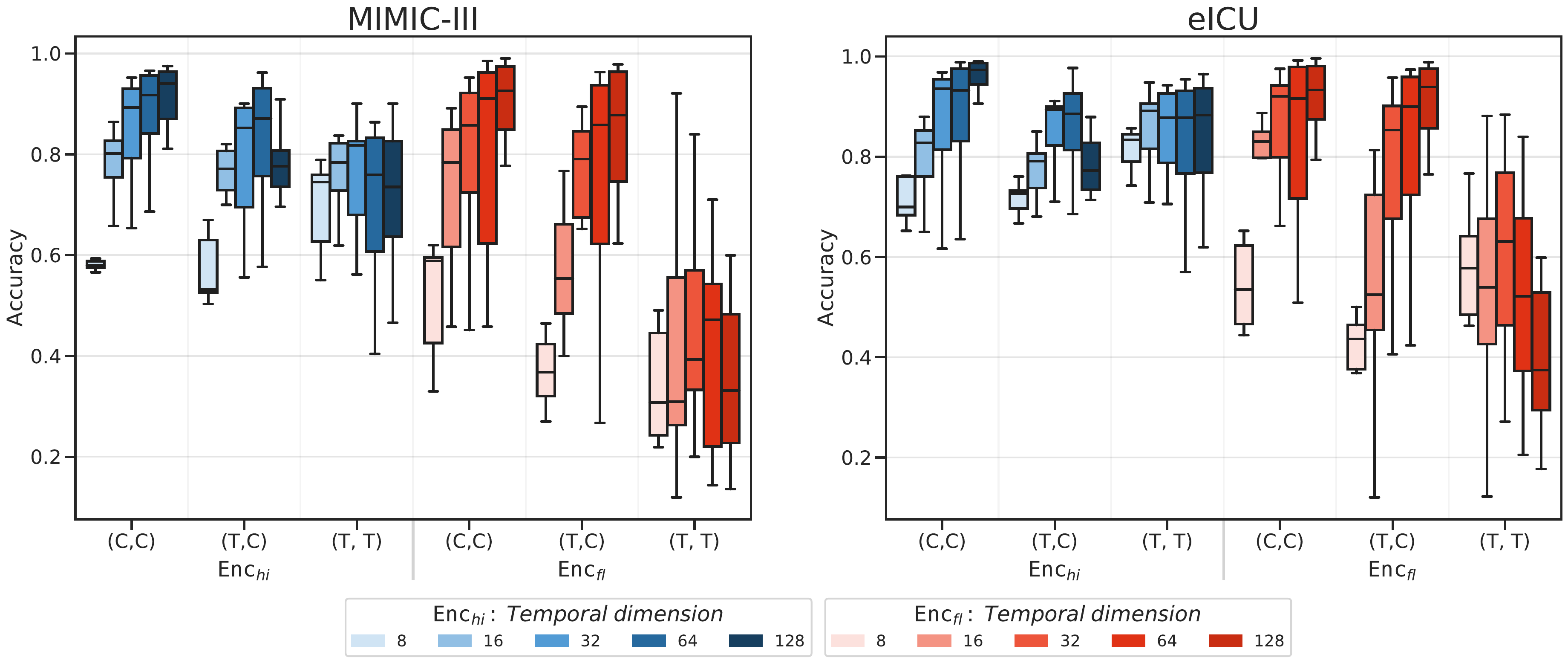}
      }
  }
  \vspace*{-3mm}
\end{figure*}

\begin{figure*}[!ht]
\vspace*{-3mm}
\floatconts
  {fig:comparison_pred_mimic_eicu}
  {\caption{Comparison of prediction performance for both MIMIC-III (left) and eICU (right) datasets. }
  \label{pred_summary}
  }
  {%
    \subfigure[Prediction performances arranged by the latent dimension $l$]{\label{pred_summary_latent}%
      \includegraphics[width=0.85\textwidth]{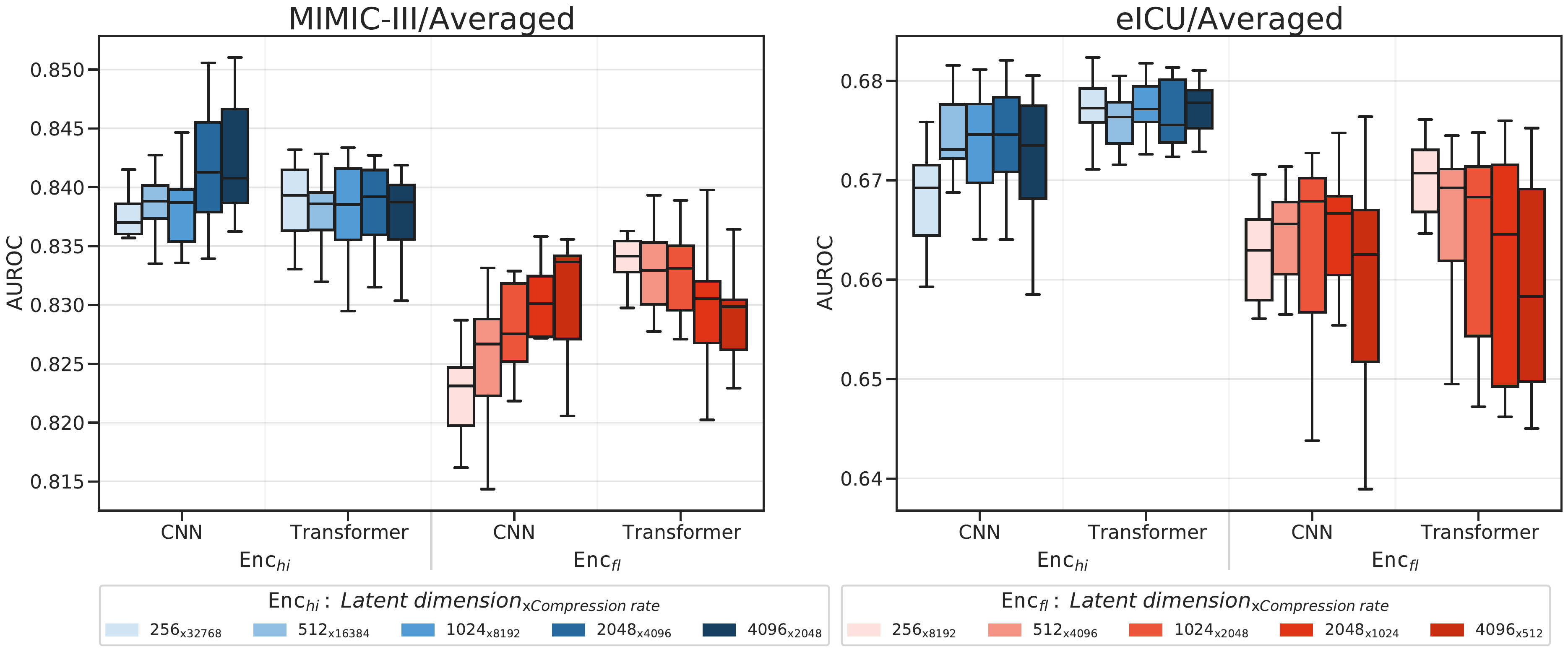}
      }
      
    \subfigure[Prediction performances arranged by the temporal dimension $t$]{\label{pred_summary_temporal}%
      \includegraphics[width=0.85\textwidth]{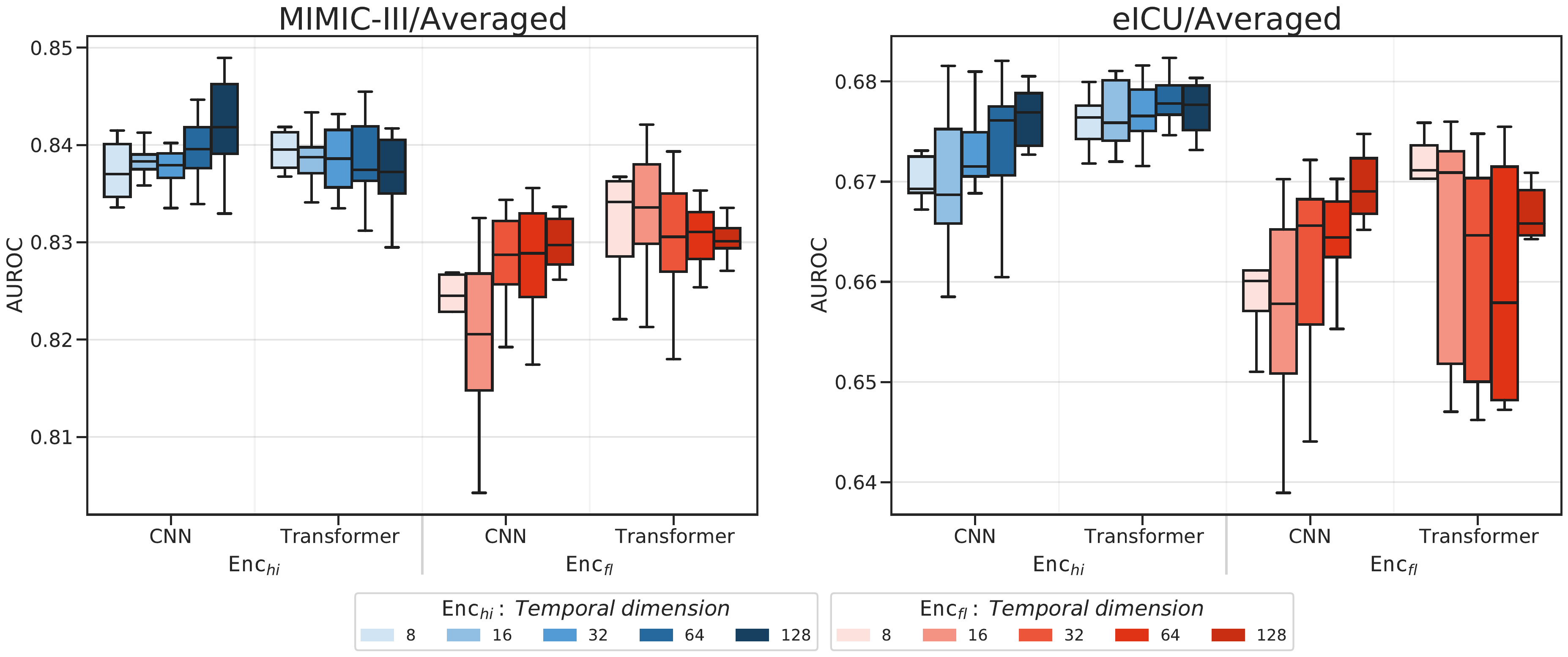}
      }%
      \vspace*{-3mm}
  }
  \vspace*{-3mm}
\end{figure*}

\section{Result}
We conducted extensive experiments on building $\Encstruc$ for the above reconstruction, prediction, and generation tasks.
Specifically, we experiment on different types of backbone models, both $\hi$ and $\fl$ structures, and various combinations of ($t$, $c$) of latent vector $\textbf{z}$ $\in \mathbb{R}^{t \times c}$.
When comparing performances using box plots, it is important to consider both the mean and best performances respectively represented by the box's position and the upper whisker, as each box contains several model variants of the same latent dimension or temporal dimension.\footnote{Ultimately, users will select the model showing best performance from each setting.}

\subsection{Reconstruction}
For the reconstruction task for MIMIC-III and eICU datasets, CNN outperforms Transformer when used as both the encoder and a decoder.
\figureref{recon_summary_latent} shows the reconstruction performances of autoencoders $(C, C)$, $(T, T)$, and $(T, C)$ at the same latent dimension $l$.
We evaluated each backbone as an encoder by comparing models with the different encoders but with the same decoder and vice versa for the decoder case.

Results in \figureref{recon_summary_latent} show that CNN is a better encoder compared to Transformer, as $(C, C)$ shows higher performance compared to $(T, C)$ for both $\hi$ and $\fl$ structures.
When compared in the same manner, CNN as a decoder showed a far better performance than Transformer with a much wider margin than the encoder case.
The conclusion that CNN is better in both aspects is reinforced by the fact that the autoencoder composed solely of CNN performs better than that of Transformers.

\paragraph{Finding} 
Our experiment results show that EHR as a time series dataset has inherent temporal locality; each medical event contained in EHR is mainly correlated with events that happened in a short period.
Specifically, the reconstruction results with CNN, having a local receptive field, outperform that of Transformer, which has a global receptive field.
Moreover, as depicted in \appendixref{apd:LocalEHR}, Transformer mainly attends temporally proximal events represented by the elements near the diagonal line, which shows an entirely different pattern in the case of prediction.
Lastly, by rearranging the results with temporal dimension $t$ as shown in \figureref{recon_summary_temporal}, CNN performs clearly better as $t$ increases, while the performance of Transformer either remains stagnant or decreases.
Thus, in order to preserve patient information with minimal loss, it is better to keep more temporal information even within the same latent dimension.

\subsection{Prediction}
For the prediction tasks, the CNN-based encoder shows comparable performance to the Transformer-based encoder in the hierarchical setting.
\figureref{pred_summary} illustrates the averaged AUROC performance of the six prediction tasks for each model architecture.
As the four models share the same classifier, we compare the results based on different encoder settings.
Specific results for each task are reported in \appendixref{apd:Pred6Tasks}.
For the $\hi$ setting, the CNN-based encoder shows comparable performance to Transformer, whereas for the $\fl$ setting, CNN shows lower performance.
Such results imply that explicit information on EHR hierarchy is more effective for the CNN compared to the Transformer in predictive tasks.

\begin{figure*}[!h]
\floatconts
  {fig:quan_eval_mimic_eicu}
  {
  \vspace*{-3mm}
  \caption{Quantitative evaluation on MIMIC-III and eICU datasets. RCE and RCS represent the ratio of correct events to total events and the ratio of correct samples to total samples, respectively.}
  \label{fig:scores}
  }
  {%
    \subfigure[MIMIC-III/RCE]{\label{}%
      \includegraphics[width=.24\linewidth]{     
      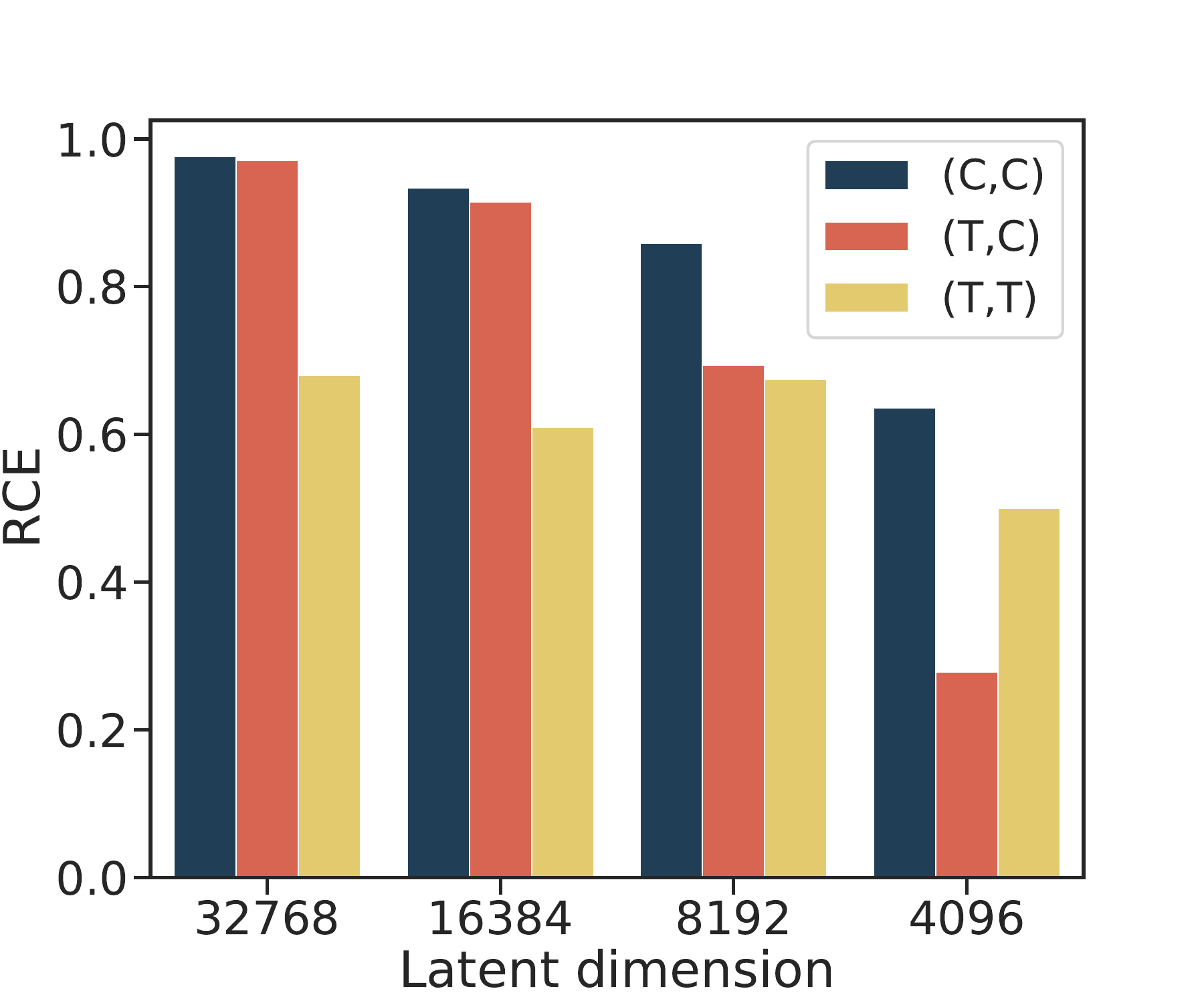}}%
    \subfigure[MIMIC-III/RCS]{\label{}
      \includegraphics[width=.24\linewidth]{     
      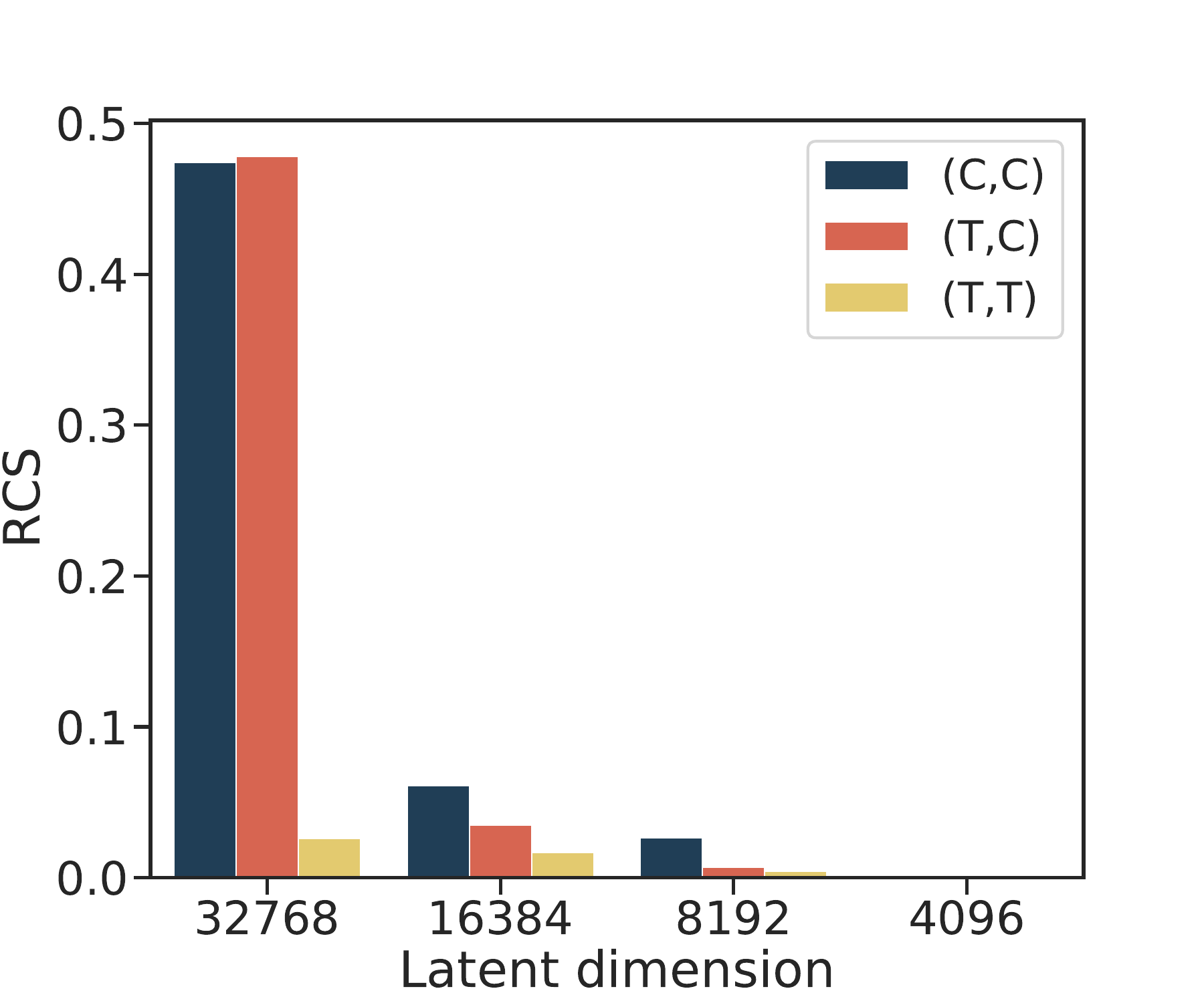}}%
    \subfigure[eICU/RCE]{\label{}%
      \includegraphics[width=.24\linewidth]{
      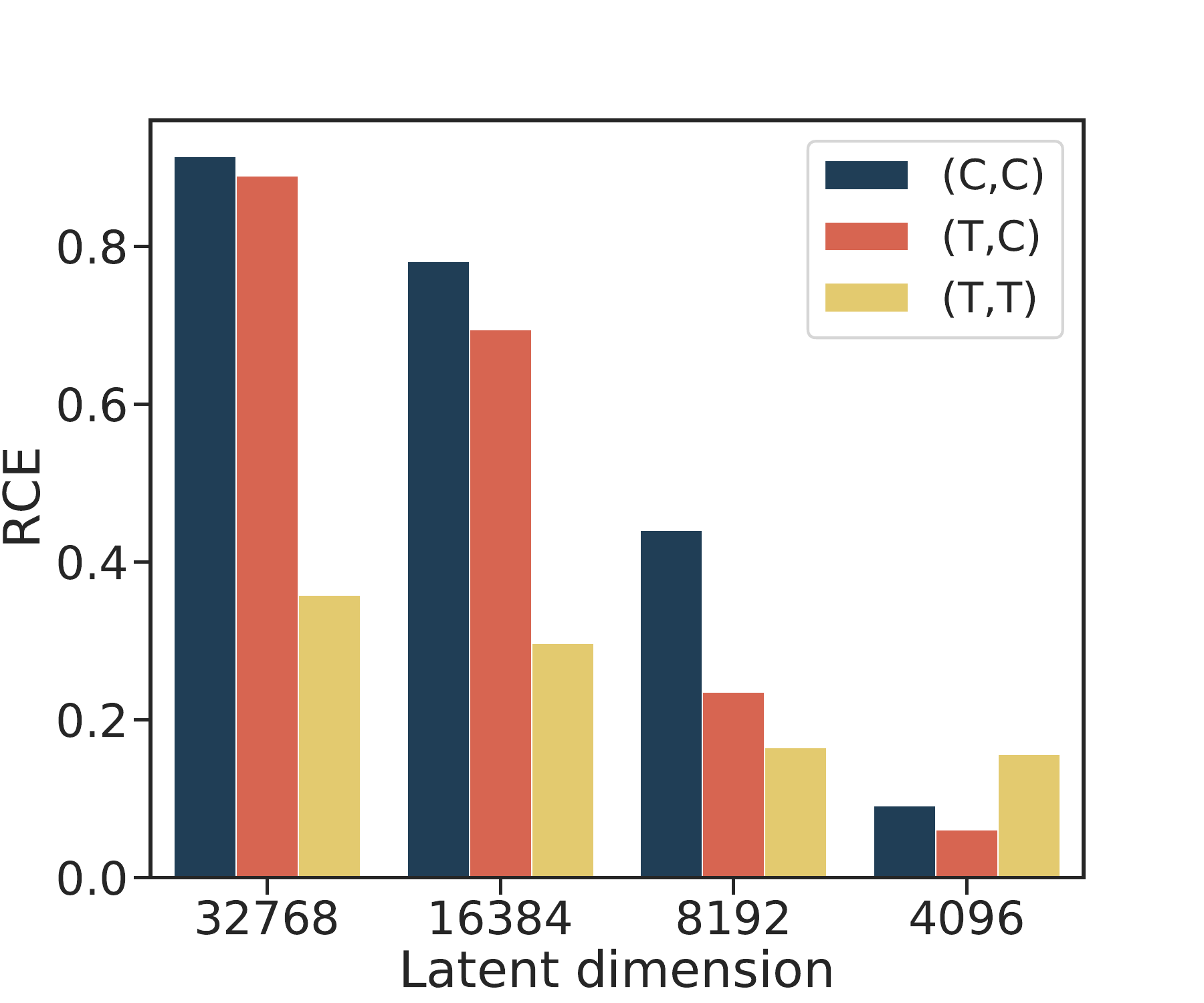}}
    \subfigure[eICU/RCS]{\label{}%
      \includegraphics[width=.24\linewidth]{
      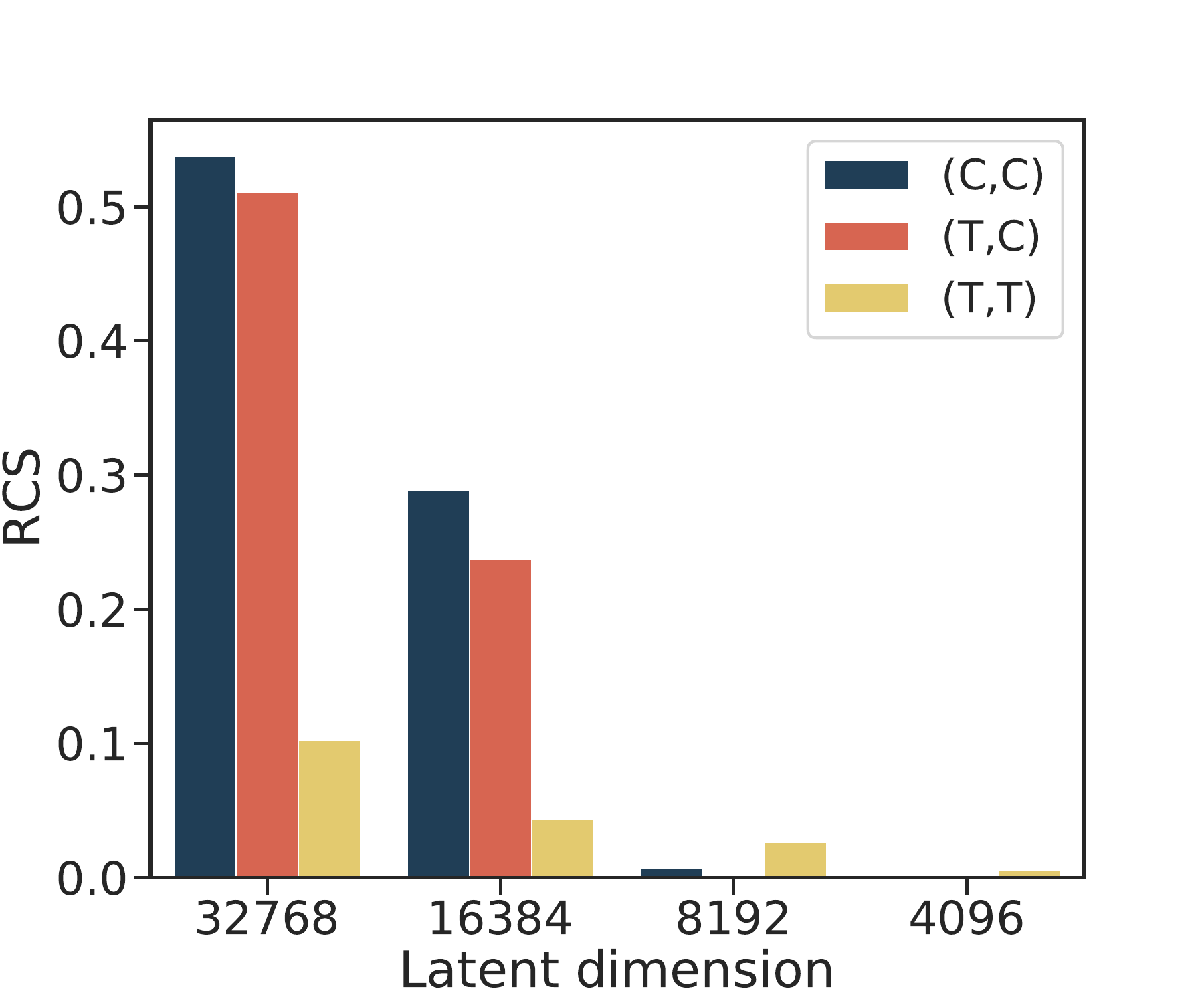}}
  }
\end{figure*}

\begin{figure*}[!ht]
\floatconts
  {fig:tsne_mimic}
  {
  \caption{t-SNE visualization on latent space on MIMIC-III dataset.}
  \label{fig:tsne}
  }
  {%
    \subfigure[(C,C)]{\label{}%
      \includegraphics[width=.25\textwidth]{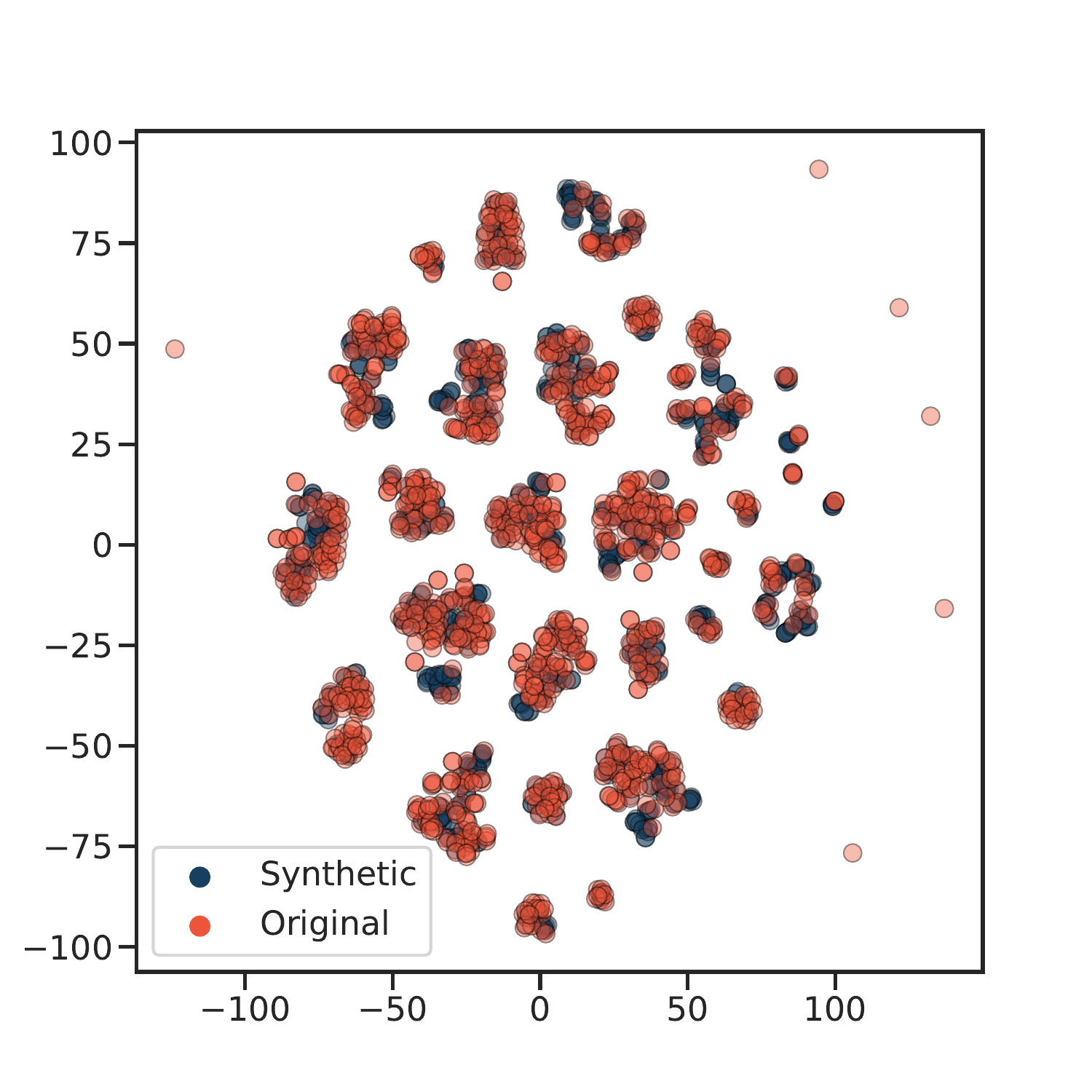}}%
    \subfigure[(T,C)]{\label{}%
      \includegraphics[width=.25\textwidth]{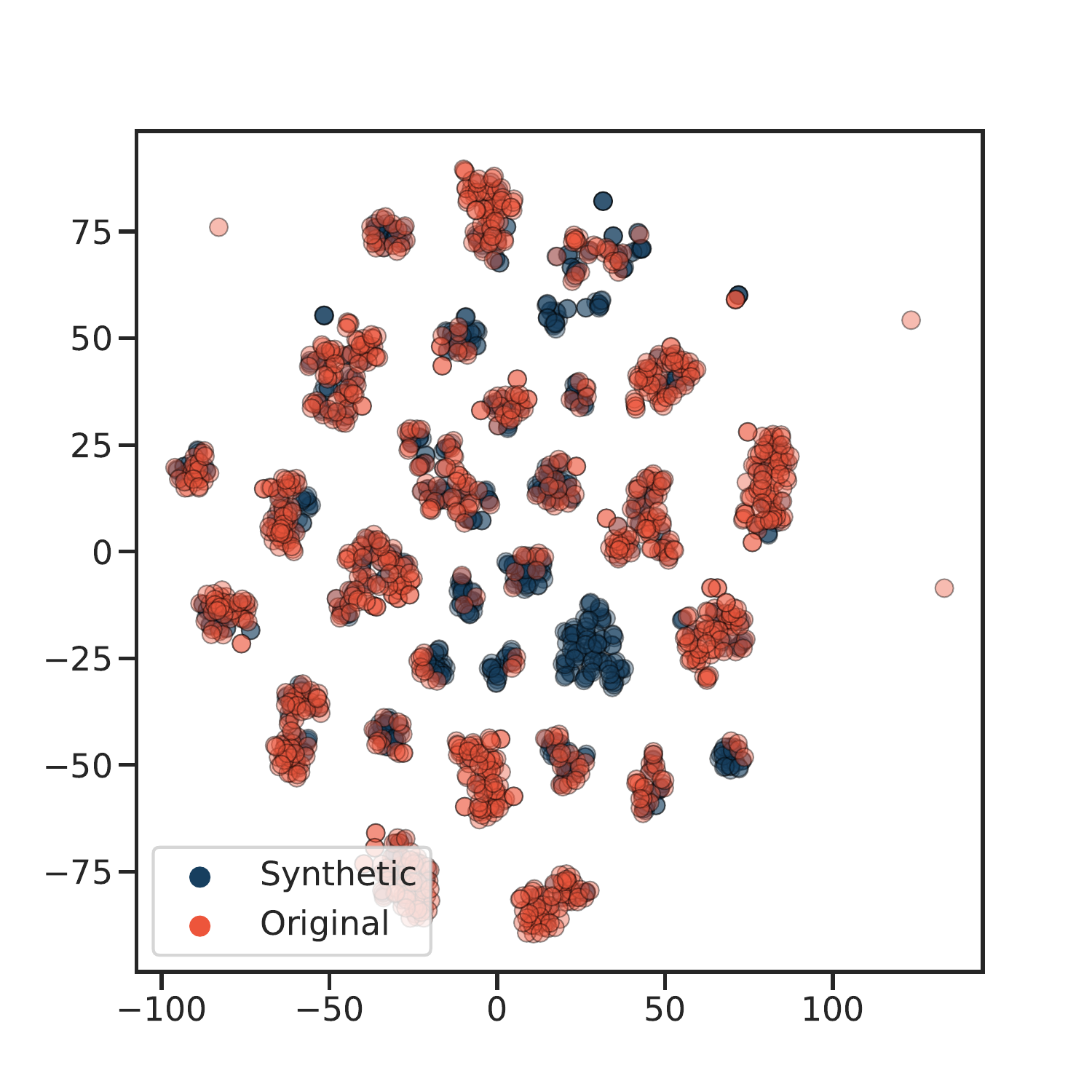}}
    \subfigure[(T,T)]{\label{}%
      \includegraphics[width=.25\textwidth]{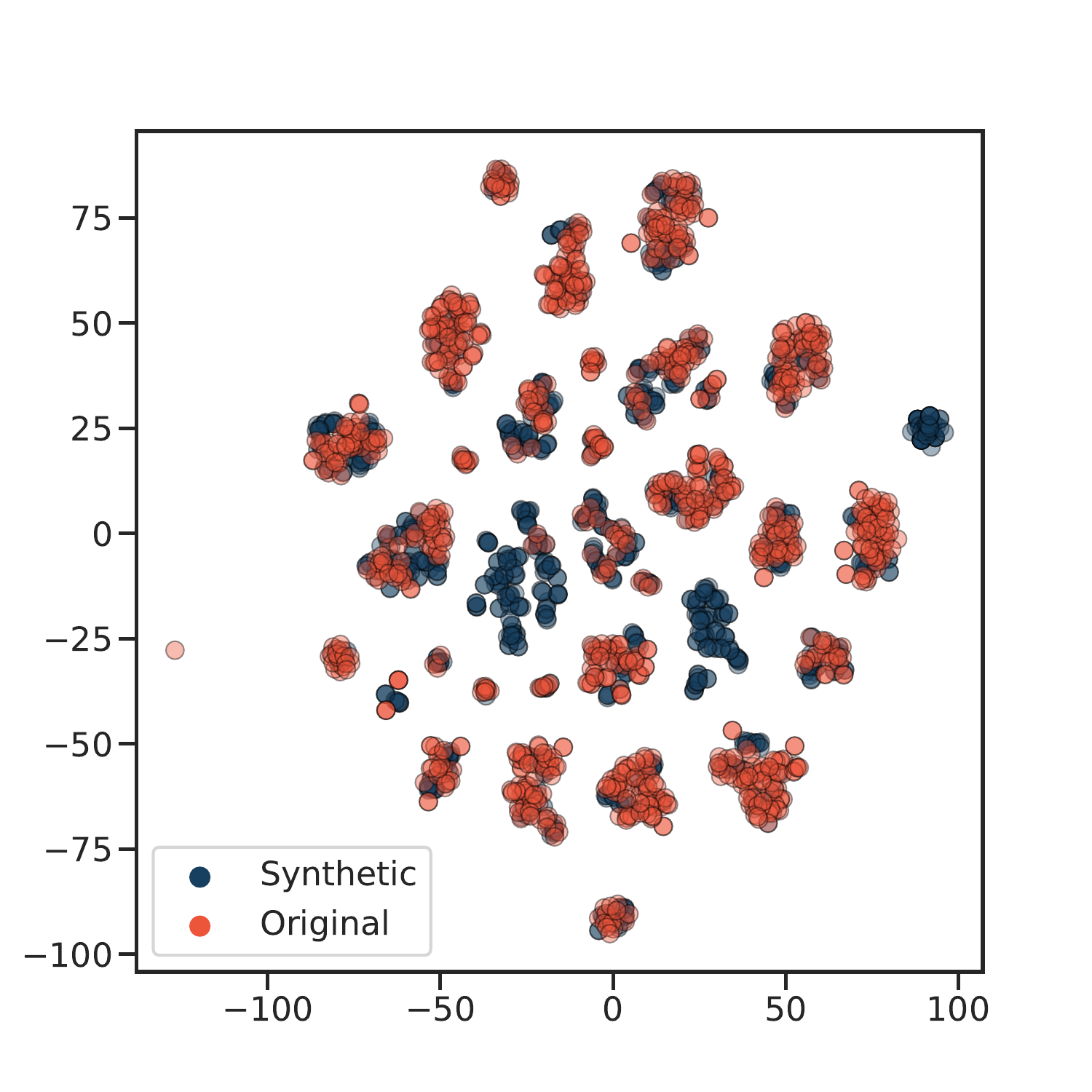}}%
  }
\end{figure*}

\paragraph{Finding} 
For the CNN-based encoder, both reconstruction and prediction tasks showed a significant positive correlation.
As shown in \figureref{apd:pca}, the CNN-based encoder learns similar latent representations for reconstruction and prediction.
Moreover, as shown in \figureref{pred_summary_temporal}, keeping more temporal information by increasing $t$ improves the predictive performance, even within the same $l$.
By increasing $t$, we can thus enhance the CNN-based encoder in both aspects simultaneously; the higher performance for both reconstruction and prediction tasks.

\subsection{Generation}
\label{section:Gen}
\paragraph{Quantitative evaluation}
As shown in \figureref{fig:scores}, we report the results of proposed metrics according to the different $(\mathtt{Enc}_{hi}, \mathtt{Dec}_{hi})$ combinations, and latent dimensions.
High RCE indicates that each generated event follows the table structure while maintaining the semantic consistency of input.
The RCS, a stricter measure, denotes the amount of synthetic patient data readily available.
In the case of latent dimension 32768, $(C,C)$ and $(T,C)$ show comparable performance. 
However, as the compression rate increases, the performance gap between $(C,C)$ and $(T,C)$ is enlarged, with $(C,C)$ showing superior performance. 
Such results indicate that the CNN-based encoder effectively preserves patient information despite a high compression rate.
$(T,T)$, on the other hand, generally shows the lowest performance compared to both $(C,C)$ and $(T,C)$.
We also report RUE on the \appendixref{apd:RUE}.

\paragraph{Qualitative evaluation}
We conduct t-SNE on the encoded latent $\textbf{z} \in \mathbb{R}^{32768}$ of $(C,C)$, $(T,C)$ and $(T,T)$ for both the original and synthetic data, as shown in \figureref{fig:tsne}.
The distributions of both the original and synthetic data surprisingly form in multiple clusters with some outliers.
For the case of $(C,C)$, although the synthetic data distribution does not cover all outliers, it shows clusterings in similar regions to the original.
On the other hand, the synthetic data of $(T,C)$ and especially $(T,T)$ also form condensed clusters that resemble those of the original data. 
However, some clusters do not contain any original data clusters.
Such results show that compared to the Transformer-based encoder, the CNN-based encoder better encodes the input with similar distributions to the original data.
We also included t-SNE visualization results according to perplexity values in \appendixref{apd:tsne_ppl}.
 
Furthermore, the poor performance of membership inference attack in \appendixref{apd:MI} showed a low possibility of privacy leakage.

\subsection{Structure}
We compared the hierarchical structure to the flattened one to measure the effect with all the other variables controlled.
In  \figureref{recon_summary} and \figureref{pred_summary}, the blue histograms are positioned higher than the red ones, indicating that hierarchically structuring the model led to higher performance across tasks and backbone models.
Using the inherent hierarchy of the EHR system can boost the model's performance.

\subsection{Time and Parameter Efficiency}
\begin{figure}[!t]
  {
    \newcommand\effreconwidth{0.22}
    \newcommand\effpredwidth{0.23}
    \subfigure[Model Size vs. Performance]{\label{fig:numparams}%
      \includegraphics[width=\effreconwidth\textwidth]{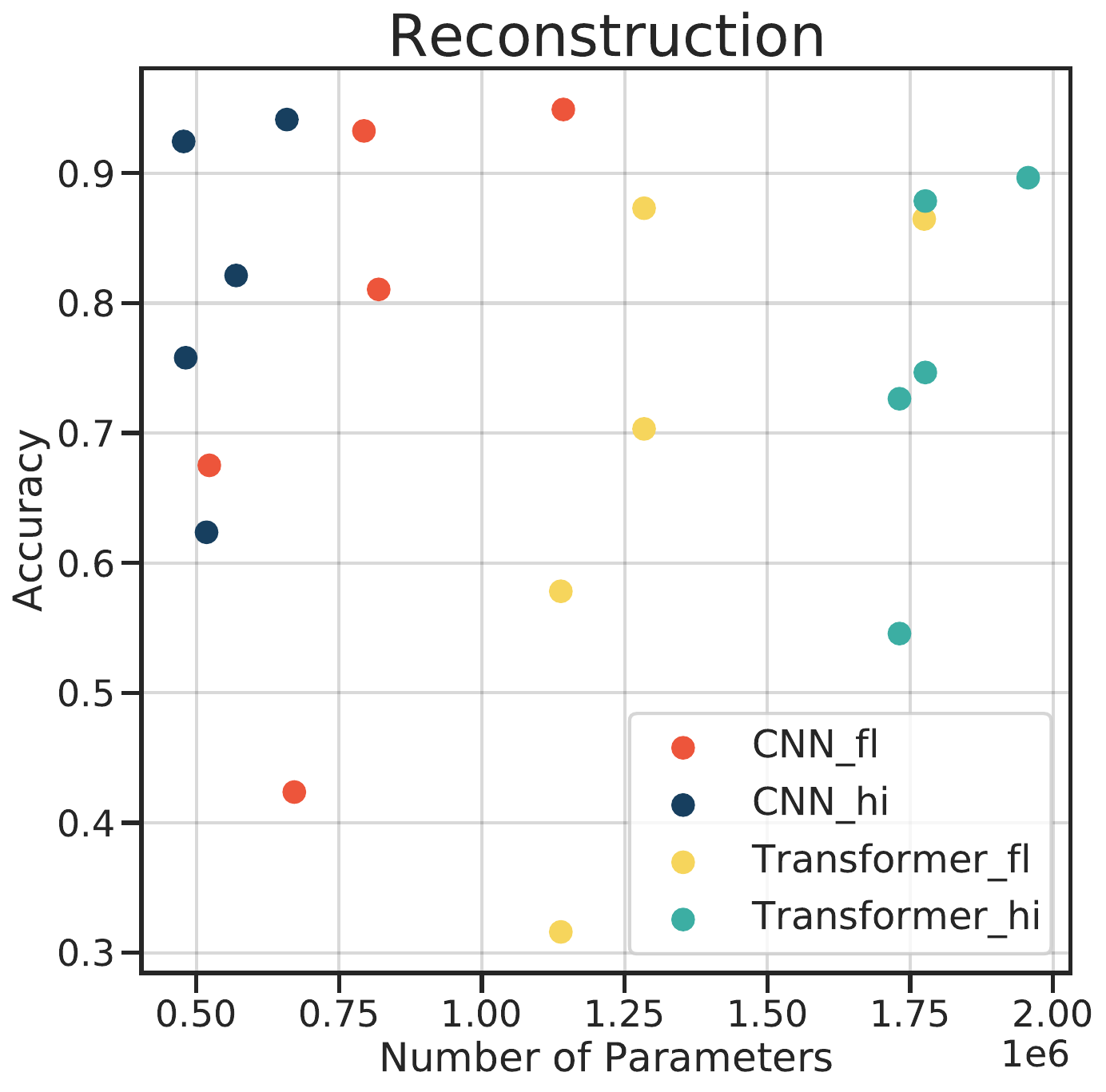}
      \includegraphics[width=\effpredwidth\textwidth]{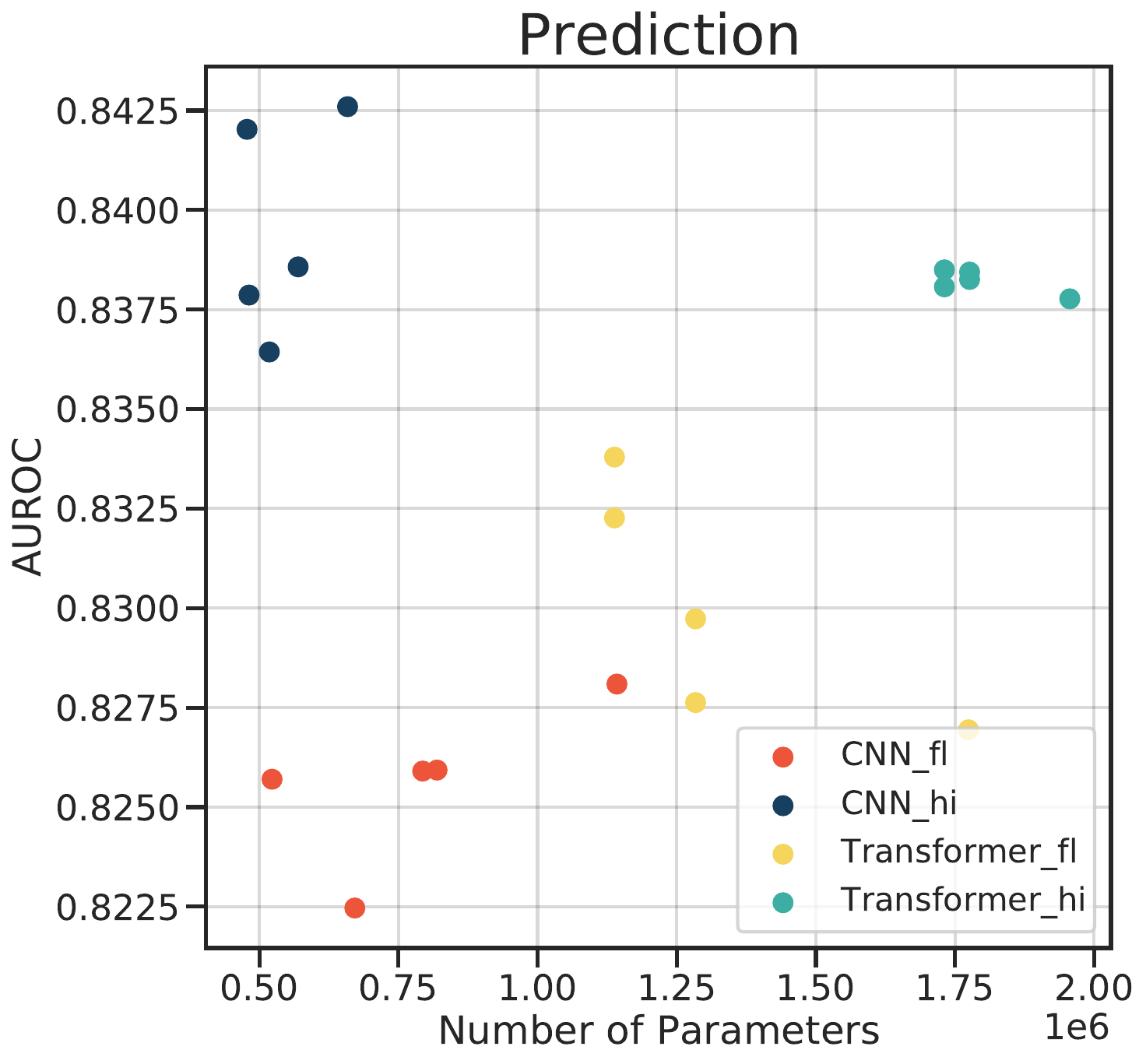}
      }%

    \subfigure[FLOPs vs. Performance]{\label{fig:flops}%
      \includegraphics[width=\effreconwidth\textwidth]{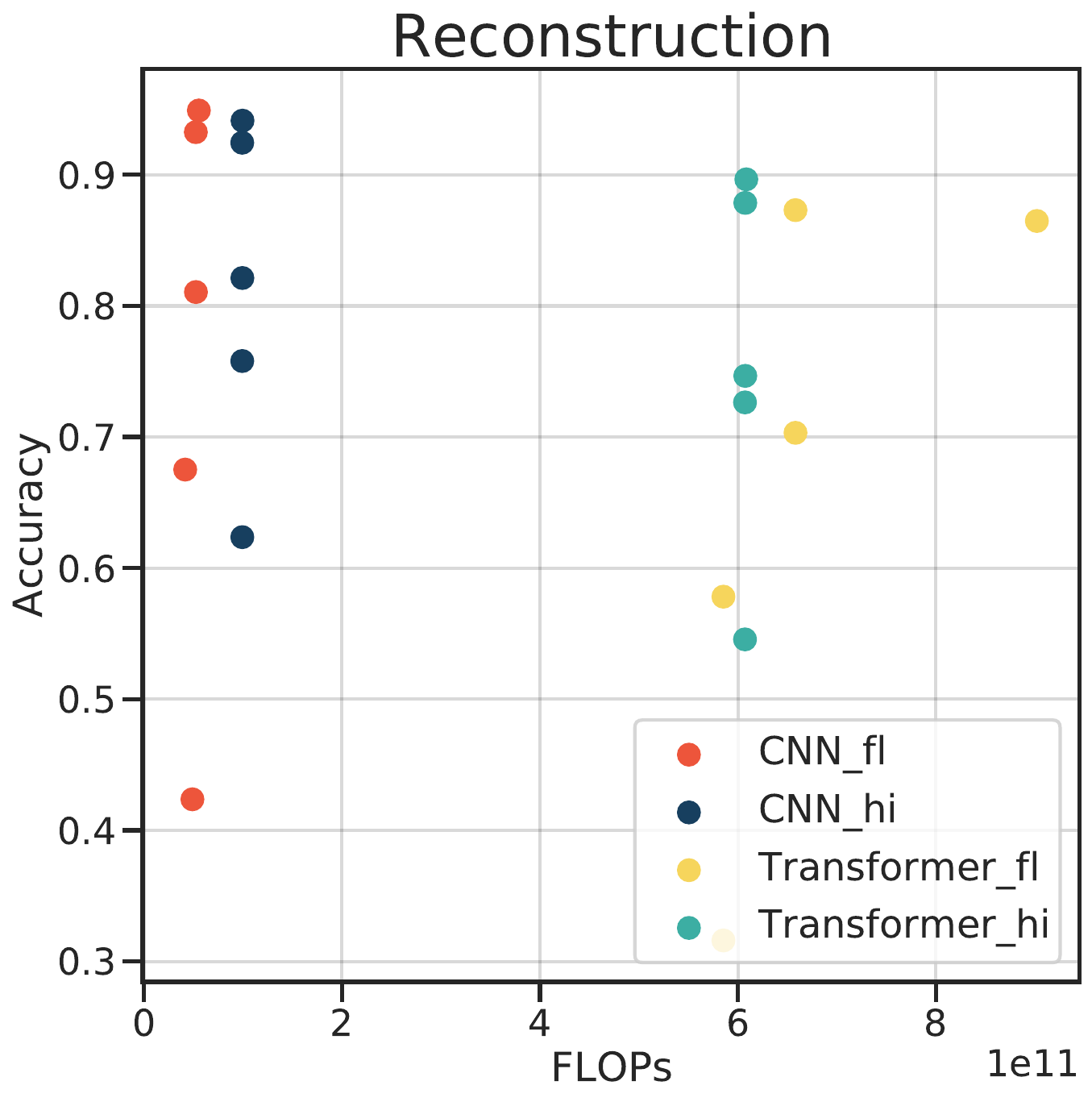}
      \includegraphics[width=\effpredwidth\textwidth]{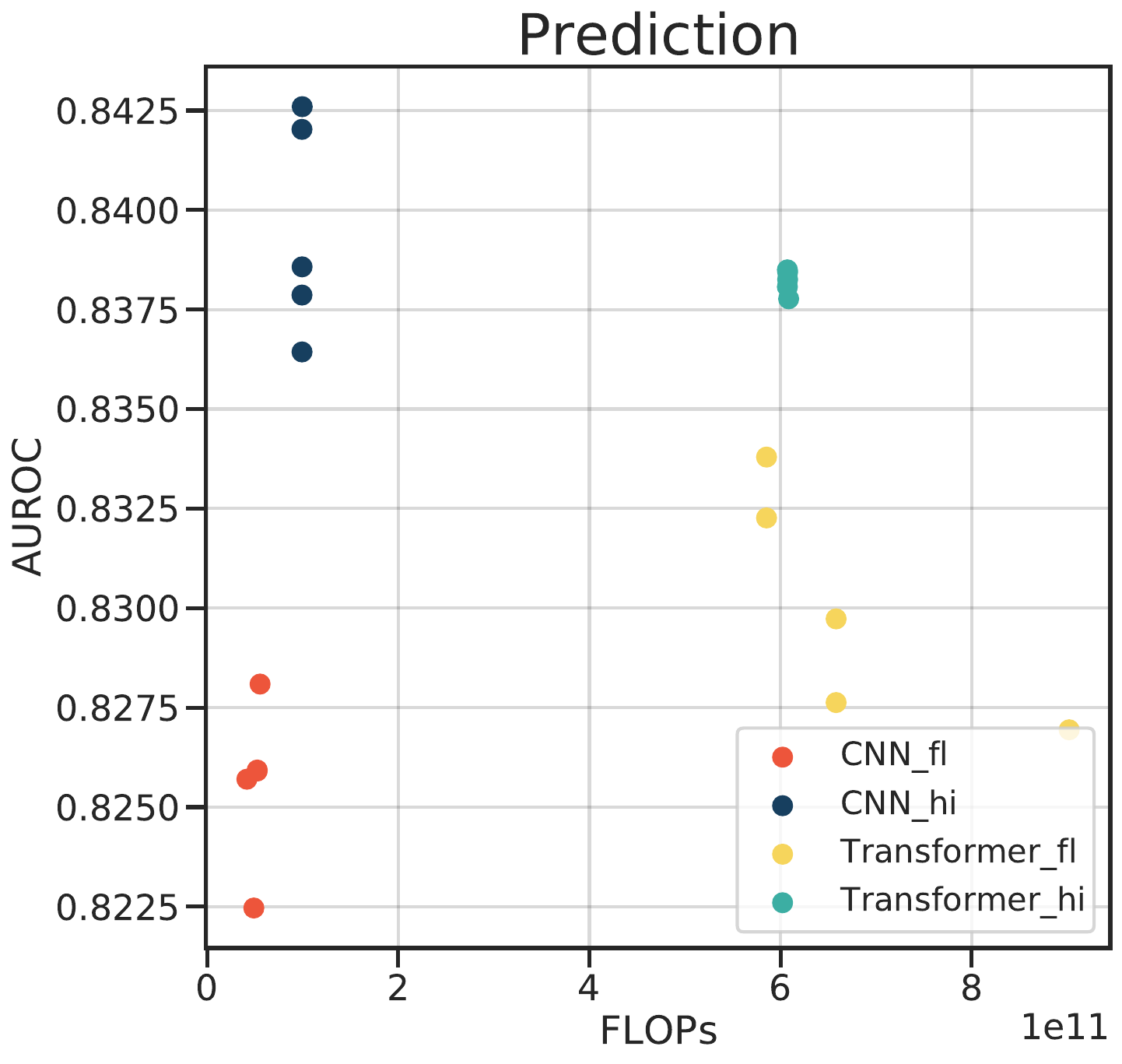}
      }%

    \subfigure[Training time vs. Performance]{\label{fig:time}%
      \includegraphics[width=\effreconwidth\textwidth]{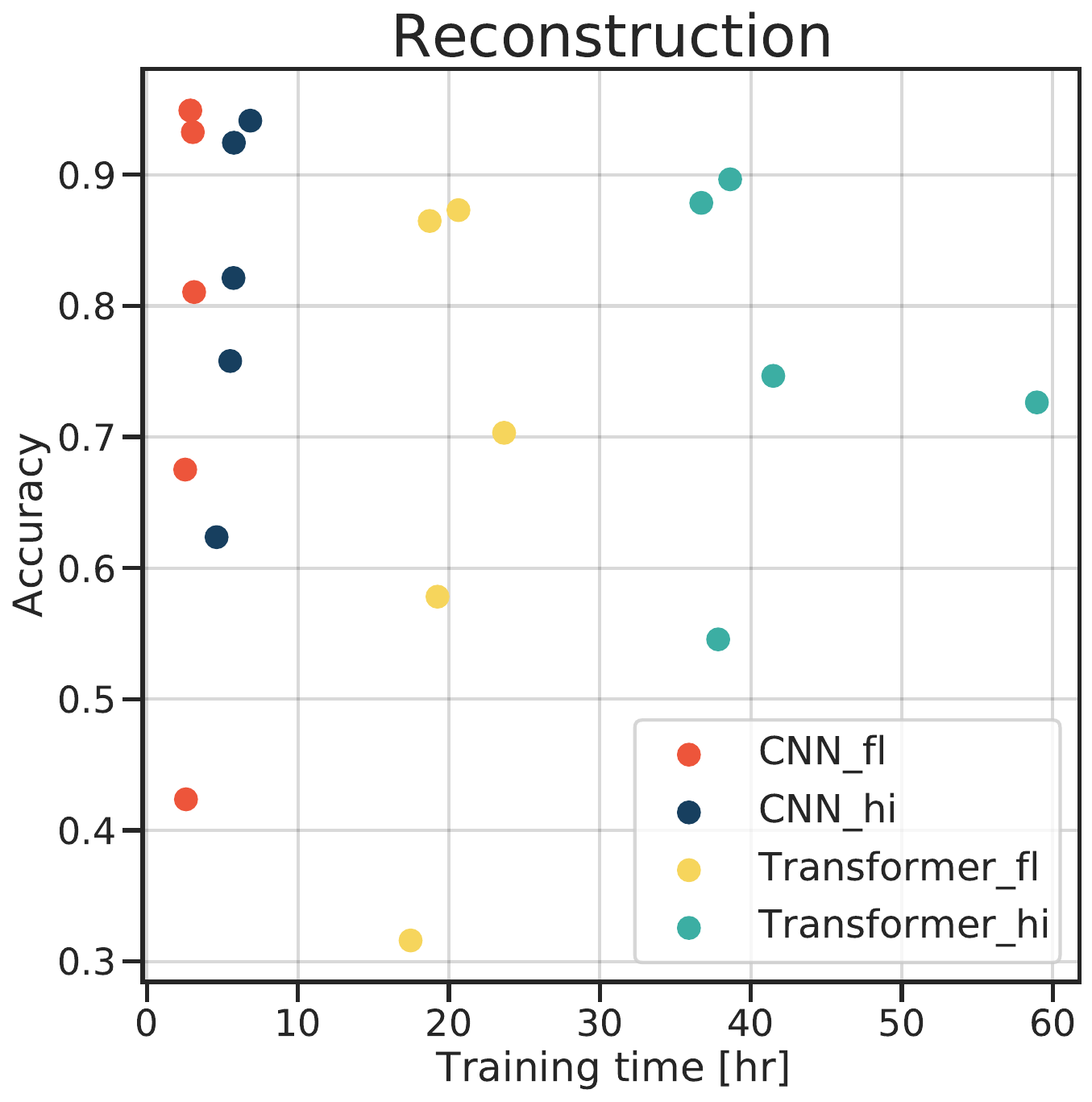}
      \includegraphics[width=\effpredwidth\textwidth]{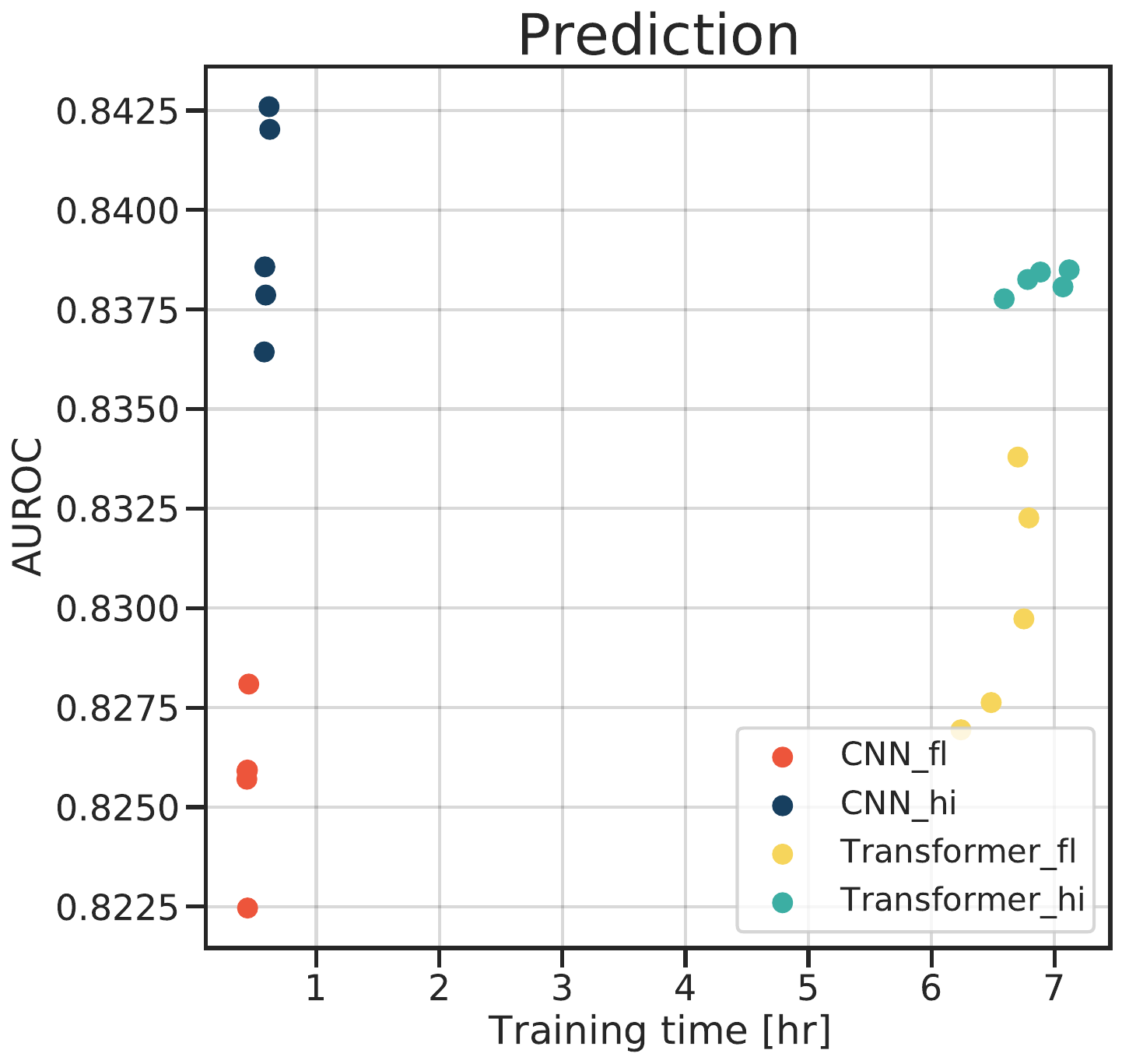}
      }%
  }
    \centering
    \caption{
    Reconstruction (left) and prediction (right) performances on MIMIC-III relative to the resource (the number of parameters, FLOPs, and training time) consumed.
    In all figures, CNN having $\hi$ structure is located in the upper left corner; it shows better performance with lower cost than the others.
    For each model, we grouped variants based on $l=t\times c$ and averaged them.
    As the recipe of building $\Encstruc$ varies by the ratio of $t$ to $c$ even within same $l$, the model size is not strictly proportional to the compression rate.
    }
    \vspace*{-6mm}
    \label{fig:efficiency}
\end{figure}

We compare the resource consumption of each model in our experiments.
\figureref{fig:efficiency} illustrates the parameters-performance, FLOPs-performance, and time-performance curves for the models performing the reconstruction task.
We only consider the encoder without the rest of the model for the number of parameters and time consumed for training.
Compared to the Transformer-based models, the results of the CNN-based models are located in the upper left part for all \ref{fig:numparams}, \ref{fig:flops} and \ref{fig:time}.
Thus, CNN is a better backbone for building an encoder than Transformer, even with notably fewer parameters and lower computational cost.

\section{Conclusion}
We have searched for a versatile architecture to encode the raw EHR input into a low-dimensional space when the input processed by the universal framework is on a large scale.
To the best of our knowledge, this is the first work to search for a versatile encoder not only reducing the large EHR into a manageable size but also well preserving the core information of patients to perform clinical tasks.
Even with fewer parameters and less training time, hierarchical CNN outperforms the state-of-the-art model on widely accepted tasks in the field.
Moreover, it turns out that making use of the inherent hierarchy of the EHR system can boost the performance of any backbone models and clinical tasks performed.
By conducting extensive experiments, we present concrete evidence for generalizing our research findings into real-world practice.
We capture the core tendencies while exploring these numerous settings and systematically summarize the findings to give a clear guideline on building the encoder.

\acks
We are grateful to Seongsu Bae, Sungjin Park, and Jiyoung Lee for their fruitful comments and inspiration.
We acknowledge the support of Google's TPU Research Cloud (TRC), which provided us with Cloud TPUs to conduct the research.
This work was supported by Institute of Information \& Communications Technology Planning \& Evaluation (IITP) grant (No.2019-0-00075), National Research Foundation of Korea (NRF) grant (NRF-2020H1D3A2A03100945), and the Korea Health Industry Development Institute (KHIDI) grant (No.HI21C1138), funded by the Korea government (MSIT, MOHW).

\bibliography{jmlr-sample}

\appendix
\vspace*{-1.5mm}

\section{Input Embedding}
Before being embedded to the encoder, token sequence $\textbf{x}$ is first mapped to a vector representation based on a learnable lookup table.
Next, we add different types of embeddings to the vector representation:
\label{apd:AdditionalEmbedding}
\begin{itemize}[noitemsep]
\item \textbf{Token-type embedding} 
adds structural information of table to the vector representation, considering that the input is based on tabular data.
The types of added embeddings are \texttt{[table name]}, \texttt{[column name]}, \texttt{[column value]}, \texttt{[timegap]}, \texttt{[start token]}, \texttt{[end token]}, and \texttt{[pad token]}.

\item \textbf{Digit-place embedding} 
adds value embeddings for numeric features (e.g. dosage, rate).
Since neural tokenizers are notorious for having difficulty in processing numbers, we use Digit-Place Embedding (DPE) to let the models to recognize numbers naturally.
DPE first splits the numeric values into digits and assigns each digit to its place value.
For example, "123.1" becomes "1 2 3 . 1”, and corresponds to “\texttt{[hundreds]}, \texttt{[tens]}, \texttt{[units]}, \texttt{[decimal point]}, \texttt{[tenth]}".
For tokens that are not a number or decimal point, we add \texttt{[non-digit]}. 

\item \textbf{Positional embedding} adds a time signal by mapping each position of the sequence to the embedding space.
Such embedding is needed for models without recurrence or convolution, such as the Transformer.
Thus, we use sinusoidal positional embedding for the Transformer-based encoder.
\end{itemize}

\section{Algorithm for building encoders and examples}

For CNN encoding scheme, \algorithmref{alg:CNNlayernum} and \ref{alg:CNNlayerorder} respectively summarize the number of layers by type and the order of layers.
To reduce the biased effect on the manifold in low-dimensional space when compressing only one dimension consecutively, we carefully design the order of layers to compress the temporal and channel dimensions alternately.
For Transformer, \algorithmref{alg:Transformer} summarizes encoding scheme.

We provide examples of algorithms for building encoders in \tableref{tab:algCNN} and \ref{tab:algTransf} for some specific cases.
\vspace*{-0.mm}

\newcommand{\dict}{D_{enc}}
\begin{algorithm2e}[bh]
\DontPrintSemicolon
\SetAlgoLined
\caption{CNN encoding scheme: the number of layers by type}
\label{alg:CNNlayernum}

\newcommand{\Ln}{L_n}
\newcommand{\Ld}{L_d}
\newcommand{\Lb}{L_{nd}}
\newcommand{\Einput}{\mathbf{E}_{input}}
\newcommand{\Eoutput}{\mathbf{E}_{output}}
\newcommand{\ninput}{n}
\newcommand{\noutput}{n'}
\newcommand{\dinput}{d}
\newcommand{\doutput}{d'}

\BlankLine
\KwIn{$\Einput \in \mathbb{R}^{n \times d}, \noutput, \doutput$}
\tcp{$(\noutput, \doutput):$ an expected shape of $\Eoutput$}
\KwOut{$\dict$}
\tcp{
$\Ln$ compresses $\ninput$-axis in half\\
$\Ld$ compresses $\dinput$-axis in half\\
$\Lb$ compresses $\ninput$ and $\dinput$ axes in half
} 
\newcommand{\nlayers}{n_l}
\tcp{Determine the number of layers $\nlayers$}

\newcommand{\rn}{r_n}
\newcommand{\rd}{r_d}
$\rn, \rd \leftarrow log_2(n/n'), log_2(d/d')$\;
$n_l\leftarrow max(\rn, \rd)$\;
\BlankLine
\tcp{Decide the number of layers by type}
\uIf{$\rn > \rd$}{
$\dict\leftarrow\{\Lb:\rd,\; \Ln: \nlayers-\rd\}$\;
}
\uElseIf{$\rn < \rd$}{
$\dict\leftarrow\{\Lb:\rn,\;\Ld: \nlayers-\rn\}$\;
}
\uElse{
$\dict\leftarrow\{\Lb: \nlayers\}$ \;
}
\end{algorithm2e}
\vspace*{-2mm}

\begin{algorithm2e}
\SetKw{kwSet}{set}
\SetKw{kwTo}{to}
\DontPrintSemicolon
\caption{Transformer encoding scheme}
\label{alg:Transformer}
\BlankLine
\KwIn{$\textbf{E}_{input} \in \mathbb{R}^{n \times d}, n', d', n_l$}
\KwOut{encoder layers}
$(n', d'):$ an expected shape of $\textbf{E}_{output}$\\
\tcp{
adaptive average pooling: $P_{adpt}(\mathbf{E}_{input}, n') \in \mathbb{R}^{n' \times d'}$
}
$r_c\leftarrow log_2(d/d')$\;
\BlankLine
$q\leftarrow r_d//n_l$\; 
$r\leftarrow r_d\%n_l$\; 
\BlankLine
\tcp{Define the encoder layers}
\kwSet{$L_{d_1}$} \kwTo{} decrease channel dim by $2^{q+1}$ \;
\kwSet{$L_{d_2}$} \kwTo{} decrease channel dim by $2^q$ \;
\BlankLine
\Return $[L_{d_1}] * r + [L_{d_2}] * (n_l-r) + [P_{adpt}]$\;
\end{algorithm2e}

\section{Definition of clinical predictive task}
\label{apd:DefClinicTask}
\begin{enumerate}
\item \textit{Diagnosis} (Dx) (multi-label):
Predict all diagnoses occurred during the entire ICU stay.
By following Clinical Classification Software (CCS) for ICD-9-CM criteria, diagnosis codes are classified into 18 categories. 

\item  \textit{Final Acuity} (Fi\_ac) (multi-class):
At the end of the ICU stay, predict where the patient will be discharged among the various places.

\item  \textit{Imminent Discharge} (Im\_disch) (multi-class):
Predict whether the patient will be discharged within the subsequent prediction window of 48 hours and, if so, where to be discharged.
\end{enumerate}
\vspace*{-3mm}

\begin{algorithm2e}[tbh]
\DontPrintSemicolon
\SetAlgoLined
\caption{CNN encoding scheme: the order of layers}
\label{alg:CNNlayerorder}

\newcommand{\Ln}{L_n}
\newcommand{\Ld}{L_d}
\newcommand{\Lb}{L_{nd}}
\newcommand{\Einput}{\textbf{E}_{input}}
\newcommand{\Eoutput}{\textbf{E}_{output}}
\newcommand{\ninput}{n}
\newcommand{\noutput}{n'}
\newcommand{\dinput}{d}
\newcommand{\doutput}{d'}
\newcommand{\nlayers}{n_l}
\newcommand{\rn}{r_n}
\newcommand{\rd}{r_d}

\BlankLine
\KwIn{$\dict,\nlayers,\rn,\rd$}
\KwOut{encoder layers}
\newcommand{\blocknb}{block_{alt}}
\newcommand{\numn}{num_n^{block}}
\newcommand{\numremainder}{num_n^{rem}}
\newcommand{\blockrep}{block_n}
\newcommand{\blocknbn}{block_{alt'}}
\tcp{$\blocknb$ compresses $n$ and $d$ axes alternatively}
\BlankLine
\uIf{$\Ln\in\dict$}{
$\numn \leftarrow (\nlayers+1)//(\rd+1)$ \;
$\numremainder \leftarrow (\nlayers+1)\%(\rd+1)$ \;
$\blockrep\leftarrow[\Ln]*\text{max}(\numn, 0)$\;
$\blocknb\leftarrow\blockrep+[\Lb]$ \;
$\blocknbn\leftarrow\blocknb+[\Ln]$ \;
\Return ($\blocknb)*(\rd-\numremainder)+(\blocknbn)*(\numremainder)+\blockrep$ \;
}
\uElseIf{$\Ld\in\dict$}{
\newcommand{\altblock}{block_{alt}}
\newcommand{\numalt}{num_{alt}}
$\altblock\leftarrow[\Lb, \Ld]$\; 
\uIf{$\nlayers-2(\rd-\rn)+1<0$}{
\newcommand{\nlodd}{n_{odd}}
$\nlodd\leftarrow\nlayers\%2$\;
\Return $[\Ld]*\nlodd+\altblock*\rn+[\Ld]*(\nlayers-2\rn-\nlodd)$\;
}

\uElse{
\uIf{$2\rn-\rd\geq0$}
{$\numalt\leftarrow\rd-\rn$}
\uElse{$\numalt\leftarrow\text{min}(r_n, r_d//2)$}
\newcommand{\numnonalt}{block_{non-alt}}

\uIf{$(\rn-\numalt)=\rd-2\numalt$}
{$\numnonalt\leftarrow[\Lb]*(\rn-\numalt)$}
\uElse{$\numnonalt\leftarrow[\Ld]*(\rd-2\numalt)$}
\Return $\numnonalt+\altblock*\numalt$\;
}
}
\uElse{
\Return $[\Lb] * \nlayers$ \;
}
\end{algorithm2e}

\begin{enumerate}
\setcounter{enumi}{3}
\item  \textit{Mortality} (Mort) (binary):
Predict whether or not a patient will be discharged with the state ``expired'' within the prediction window of 48 hours.
The discharge state was ``expired'' within the prediction window of 48 hours.

\item  \textit{Length-of-Stay} (binary):
Predict whether the patient’s whole length of stay will be longer than 3 days or not (LOS3), and 7 days or not (LOS7).
\end{enumerate}
\vspace*{-5mm}

\section{Choice of classifier backbone for the prediction tasks}
\label{apd:BuildingClassifier}
We consider using MLP and Transformer as backbone of the classifier for $\textbf{z}$.
When CNN encodes $\textbf{x}$ into $\textbf{z}$, the temporal order of $\textbf{x}$ is preserved (i.e., permutation equivariant).
Thus, the classifier should aggregate $\textbf{z}$ to be permutation invariant for better prediction.

For the MLP-based classifier, $\textbf{z}$ is flattened into a 1D vector, then linearly projected into logits.
However, only specific parameters are used to process $\textbf{z}$ at specific locations, preventing full aggregation of $\textbf{z}$ into logits.

The Transformer-based classifier, on the other hand, passes $\textbf{z}$ into the self-attention layer, resulting in complete aggregation of $\textbf{z}$ (i.e., permutation invariant).

\begin{figure}[hb]
    \includegraphics[width=\linewidth]{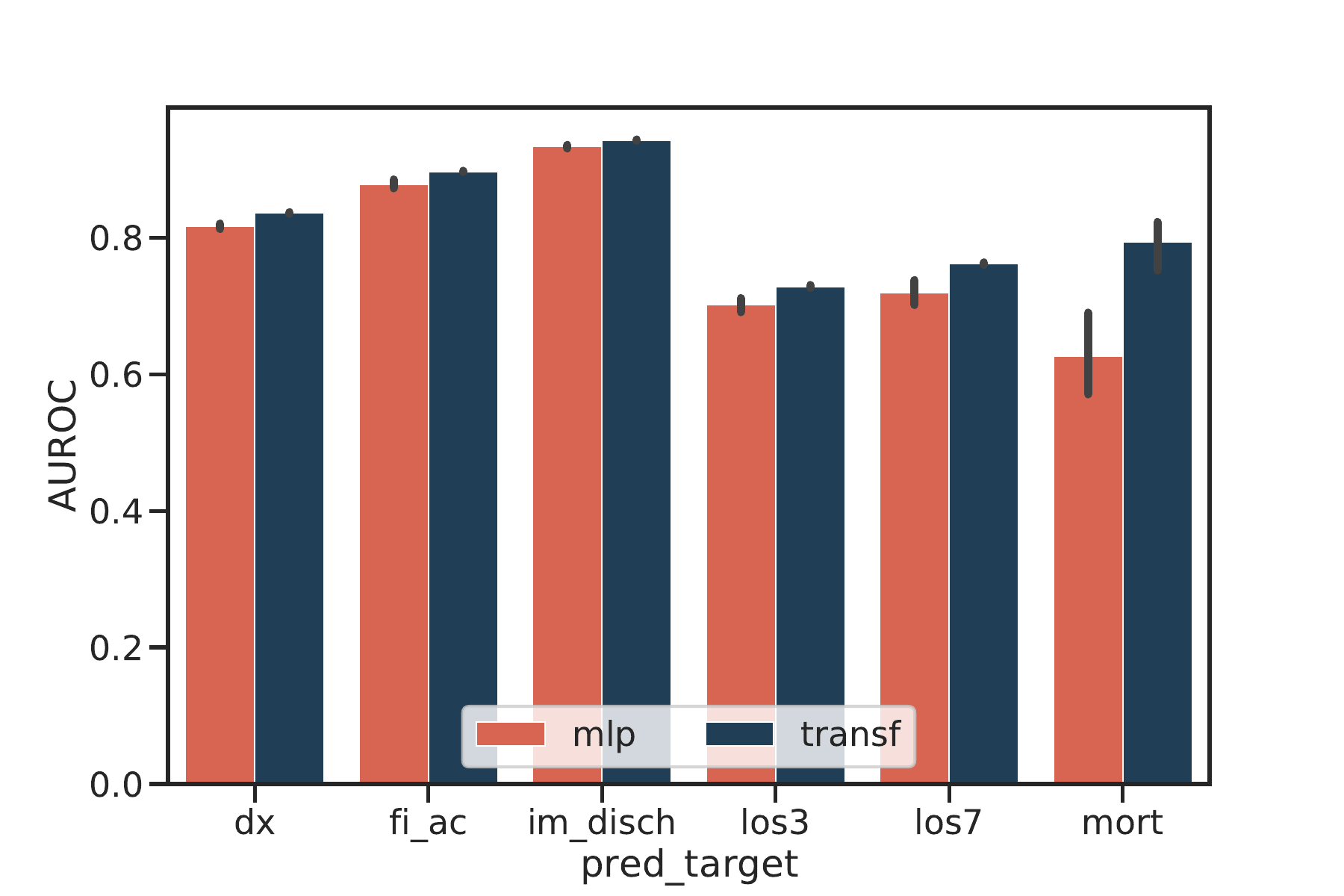}
    \centering
    \vspace*{-8mm}
    \caption{Prediction results according to the classifier backbone when $\textbf{z}\in \mathbb{R}^{2048}$ is compressed by a CNN-based one-stage encoder.}
    \label{fig:clsf}
\end{figure}

As a result, we choose the Transformer-based classifier instead of the MLP classifier.
\figureref{fig:clsf} also shows that the CNN-based encoder using Transformer as a classifier has a higher AUROC than MLP, proving that prediction with permutation invariant backbone is a better choice.
\vspace*{-5mm}
\vspace*{-5mm}

\begin{table}[]
\caption{Example of CNN-based encoder compressing $\textbf{E}_{input} \in \mathbb{R}^{8192 \times 256}$ to $\textbf{E}_{output} \in \mathbb{R}^{64 \times 8}$ following \algorithmref{alg:CNNlayernum} and \ref{alg:CNNlayerorder}.}
\vspace*{-3mm}
\label{tab:algCNN}
\centering
\begin{tabular}{c|cc}
\toprule
\textbf{Encoding} & \multicolumn{2}{c}{\textbf{CNN-based}} \\ \hline
Input & \multicolumn{2}{c}{\begin{tabular}[c]{@{}c@{}}$n=8192, d=256$\\ $\textbf{n'=64, d'=8}$\end{tabular}} \\ \midrule
\begin{tabular}[c]{@{}c@{}}Build\\ Layers\end{tabular} & \multicolumn{2}{c}{\begin{tabular}[c]{@{}c@{}}$r_n = log_2(8192/64)=7$\\ $r_d = log_2(256/8)=5$\\ $n_l = 7$\end{tabular}} \\ \midrule
Layer & \multicolumn{1}{c|}{Layer type} & Output shape \\ \midrule
1 & \multicolumn{1}{c|}{$L_{nd}$} & (4096,128) \\
2 & \multicolumn{1}{c|}{$L_{nd}$} & (2048,64) \\
3 & \multicolumn{1}{c|}{$L_{nd}$} & (1024,32) \\
4 & \multicolumn{1}{c|}{$L_{nd}$} & (512,16) \\
5 & \multicolumn{1}{c|}{$L_{n}$} & (256,16) \\
6 & \multicolumn{1}{c|}{$L_{nd}$} & (128,8) \\
7 & \multicolumn{1}{c|}{$L_n$} & \textbf{(64,8)} \\
\bottomrule
\end{tabular}
\end{table}

\begin{table}[]
\caption{Example of Transformer-based encoder compressing $\textbf{E}_{input} \in \mathbb{R}^{8192 \times 256}$ to $\textbf{E}_{output} \in \mathbb{R}^{64 \times 8}$ according to \algorithmref{alg:Transformer}.}
\vspace*{-3mm}
\label{tab:algTransf}
\centering
\begin{tabular}{c|cc}
\toprule
\textbf{Encoding} & \multicolumn{2}{c}{\textbf{Transformer-based}} \\ \hline
Input & \multicolumn{2}{c}{\begin{tabular}[c]{@{}c@{}}$n=8192, d=256$\\ $\textbf{n'=64, d'=8}$, $n_l=4$\end{tabular}} \\ \midrule
\begin{tabular}[c]{@{}c@{}}Build\\ Layers\end{tabular} & \multicolumn{2}{c}{\begin{tabular}[c]{@{}c@{}}$r_d = log_2(256/8)=5$\\ $q=1, r=1$\\ $L_{d_1}$: decrease $d$ by $2^2$ \\ $L_{d_2}$: decrease $d$ by $2^1$ \end{tabular}} \\ \midrule
Layer & \multicolumn{1}{c|}{Layer type} & Output shape \\ \midrule
1 & \multicolumn{1}{c|}{$L_{d_1}$} & (8192,64) \\
2 & \multicolumn{1}{c|}{$L_{d_2}$} & (8192,32) \\
3 & \multicolumn{1}{c|}{$L_{d_2}$} & (8192,16) \\
4 & \multicolumn{1}{c|}{$L_{d_2}$} & (8192,8) \\
- & \multicolumn{1}{c|}{$P_{adpt}$} & \textbf{(64,8)} \\
\bottomrule
\end{tabular}
\end{table}

\section{VQ-VAE}
\label{apd:VQVAE}
\paragraph{Stage 1. Learning a codebook}
With the conventional VQ-VAE method, latent vector $\textbf{z} \in \mathbb{R}^{t \times c}$ consists of $t$ fibers, and each fiber $\textbf{z}^i \in \mathbb{R}^c$ is mapped to the nearest code.
However, in order to improve the representation of each fiber, we divide each fiber into four pieces and replace each piece $\textbf{z}^{i,j} \in \mathbb{R}^{c/4}$ $(j \in 1,...,4)$ with its closest code from the codebook $\{\textbf{e}_k\}_{k=1}^K \in \mathbb{R}^{c/4}$ as follows:
$$\textbf{z}_q^{i,j} = \textbf{e}_k \ \text{where} \ k = \text{arg min}_l ||\textbf{z}^{i,j}-\textbf{e}_l||$$
As a result, $\textbf{z}$ is mapped into $\textbf{z}_q$ with $4t$ codes, and $\textbf{z}_q$ is fed to the decoder to reconstruct $\tilde{\textbf{x}}$.

The encoder, decoder, and codebook are trained in an end-to-end manner to minimize the distance between $\textbf{x}$, $\tilde{\textbf{x}}$ and $\textbf{z}$, $\textbf{z}_q$ respectively:
$$\mathcal{L}_{VQ}= ||\textbf{x}-\tilde{\textbf{x}}||^2_2 + ||\text{sg}[\textbf{z}]-\textbf{z}_q||^2_2 + \beta ||\text{sg}[\textbf{z}_q]-\textbf{z}||^2_2$$
where  sg$[\cdot]$ stands for the stop gradient operation, which supplements the non-differentiable quantization operation, and $\beta$ is a weight hyperparameter. 
In our case, we replace the second loss term with an exponential moving average for the codebook.

\paragraph{Stage 2. Learning a prior over discrete latents}
We train the Transformer-based autoregressive model to learn the prior distribution over the discrete latent space:
$$p(\textbf{z}_q)=\prod_{i}^{4t} p(\textbf{z}_{q,i}|\textbf{z}_{q,1}, ..., \textbf{z}_{q,i-1})$$
The autoregressive (AR) model predicts the next code based on past codes on every step to maximize the log-likelihood of the joint distribution of $p(\textbf{z}_q)$:
$$\mathcal{L}_{AR}=\mathbb{E}_{\textbf{x} \sim p(\textbf{x})}[-\log{p(\textbf{z}_q)}]$$
Finally, the sampled latent code sequence from $p(\textbf{z}_q)$ is decoded to generate $\mathbf{\hat{x}}$.

\section{Qualitative Evaluation Method for Synthetic Data}
\label{apd:SyntheticDataScoring}
Real data consists of samples representing patient data, and each sample consists of multiple events $\mathcal{E}_R$.
Each event $e_R \in \mathcal{E}_R $,  can be expressed as follows: $ (\texttt{table}, (\texttt{column name}, \texttt{cell}) \times \textit{n}, \texttt{timegap})$, where \textit{n} represents the number of columns for the event.
Likewise, synthetic data consists of samples with multiple events $\mathcal{E}_G$. 
We evaluate the quality of synthetic data based on table syntax preservation and semantics consistency by comparing triples (\texttt{table}, \texttt{column name}, \texttt{cell}) in $\mathcal{E}_G$ to $\mathcal{E}_R$.
The two-step procedure is as follows: (1) Definition of a set of triples (\texttt{table}, \texttt{column name}, \texttt{content}) based on $\mathcal{E}_R$, (2) Synthetic data evaluation based on pre-defined triples.

\paragraph{Definition of a set of triples}
Each event $e_R$ is first split into triples of (\texttt{table}, \texttt{column name}, \texttt{cell}).
Each \texttt{cell} can be categorized as either numeric or text-type.
For the (\texttt{table}, \texttt{column name}) combinations of all resulting triples from $\mathcal{E}_R$, we remove duplicates and extract only the minimum and maximum values for numeric-type \texttt{cell}s while performing tokenization for text-type $\texttt{cell}$s.
As a result, we build a refined set of triples (\texttt{table}, \texttt{column name}, \texttt{content}).

\paragraph{Synthetic data evaluation}

\begin{algorithm2e}[t]
\SetKw{kwSet}{set}
\SetKw{kwTo}{to}
\DontPrintSemicolon
\caption{Qualitative Evaluation Metric for Synthetic Data}
\label{alg:Metric}
\For {${e}_G \in \mathcal{E}_G$} {
    \tcp{ syntax check}
    \If {not (\texttt{order}($e_G$) and \texttt{column\_pair}($e_G$) and \texttt{table\_column}($e_G$))}{
        \Return incorrect \;
    }
    \tcp{ semantics check}
    \For{(column, content) in $e_G$} {
        \uIf {type(column) is numeric}{
            \uIf{not min\_max(content)}{
            \Return incorrect \;
            }
        }
        \uElse{
        \uIf{not sub\_word(content)}{
            \Return incorrect\;
            }
        }
    }
    \Return correct\;
}
\end{algorithm2e}

We define the functions of \algorithmref{alg:Metric} as follows:

To measure syntactic consistency, \textbf{\textit{order}} checks whether $e_G$ starts with a table name.
\textbf{\textit{column\_pair}} checks whether columns and contents consist in a pair.
\textbf{\textit{table\_column}} checks whether the table and column names are in the pre-defined set.

To evaluate semantic consistency,
for numeric-type cells, \textbf{\textit{min\_max}} checks whether the content lies within the minimum and maximum value of $(\texttt{table}, \texttt{column name})$ in the pre-defined set.
For the text-type cells, \textbf{\textit{sub\_word}} splits the cell values into subwords and then checks whether the split content is in the pre-defined set of $(\texttt{table}, \texttt{column name})$.

Based on evaluation of each $e_G$ with \algorithmref{alg:Metric},  we use the 3 metrics below to report the synthetic data scores:
\setlist{nolistsep}
\begin{itemize}[noitemsep]
\item \textbf{RCE}: ratio of correct events to total events
\item \textbf{RUE}: ratio of correct unique events to total unique events
\item \textbf{RCS}: ratio of correct samples to total samples
\end{itemize}
Specifically, RCE and RUE evaluate the number of correct events given all generated events on a non-unique and unique basis, respectively.
RCS evaluates the number of correct patient samples out of all generated patient samples, in which each patient sample is a batch of events.
By measuring the event-level and sample-level accuracy of synthetic data, we can analyze the consistency of synthetic data with real data.

\vspace*{-4mm}
\section{Additional results of qualitative evaluation (RUE)}
\label{apd:RUE}
\begin{figure}[!h]
\floatconts
  {fig:qual_eval_RUE}
  {\caption{Quantitative evaluation on MIMIC-III and eICU datasets. RUE represents the ratio of correct unique events to total unique events.}
  \label{fig:RUE}
  }
  {%
    \subfigure[RUE/mimic-III]{\label{20}%
      \includegraphics[width=0.45\linewidth]{     
      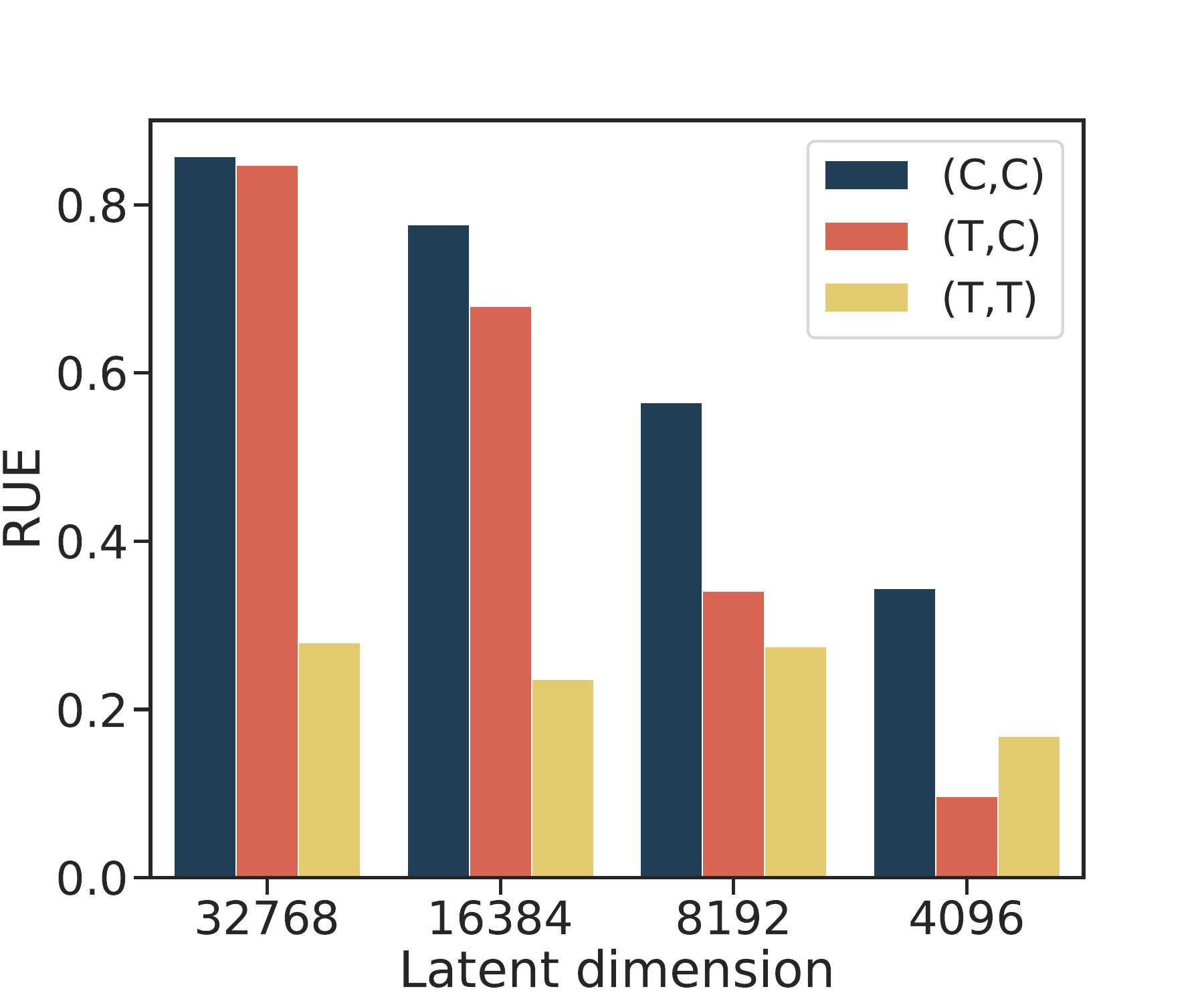}}
    \subfigure[RUE/eicu]{\label{}%
      \includegraphics[width=0.45\linewidth]{
      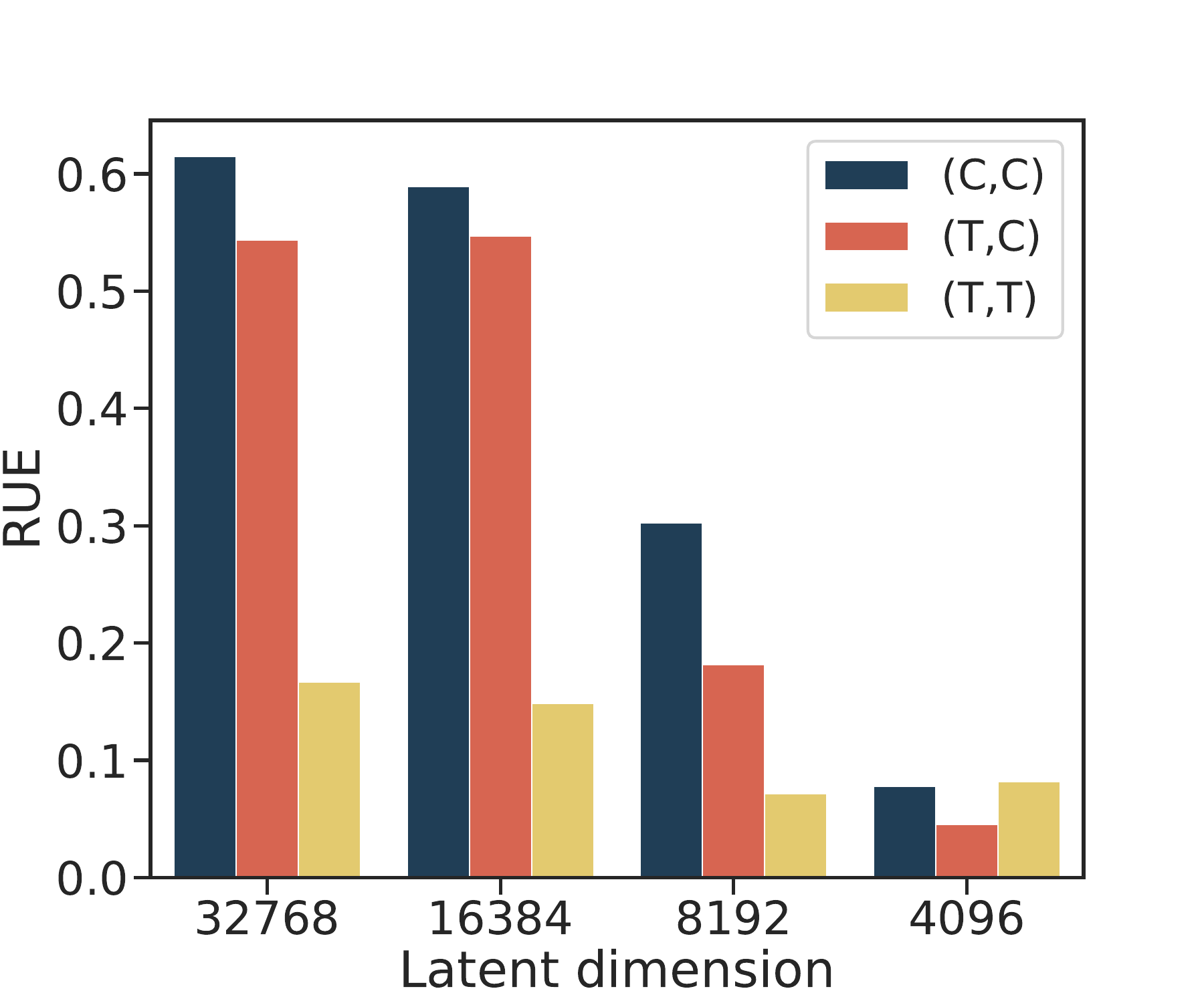}}
  }
\end{figure}

We measured RUE metrics for synthetic data. 
Since RUE ignores duplicates and only considers unique events, it can measure the generative data from a different perspective than RCE.
However, as shown in \figureref{fig:RUE}, it matches other metrics in \sectionref{section:Gen}.

\begin{figure}[!ht]
    \includegraphics[width=\linewidth]{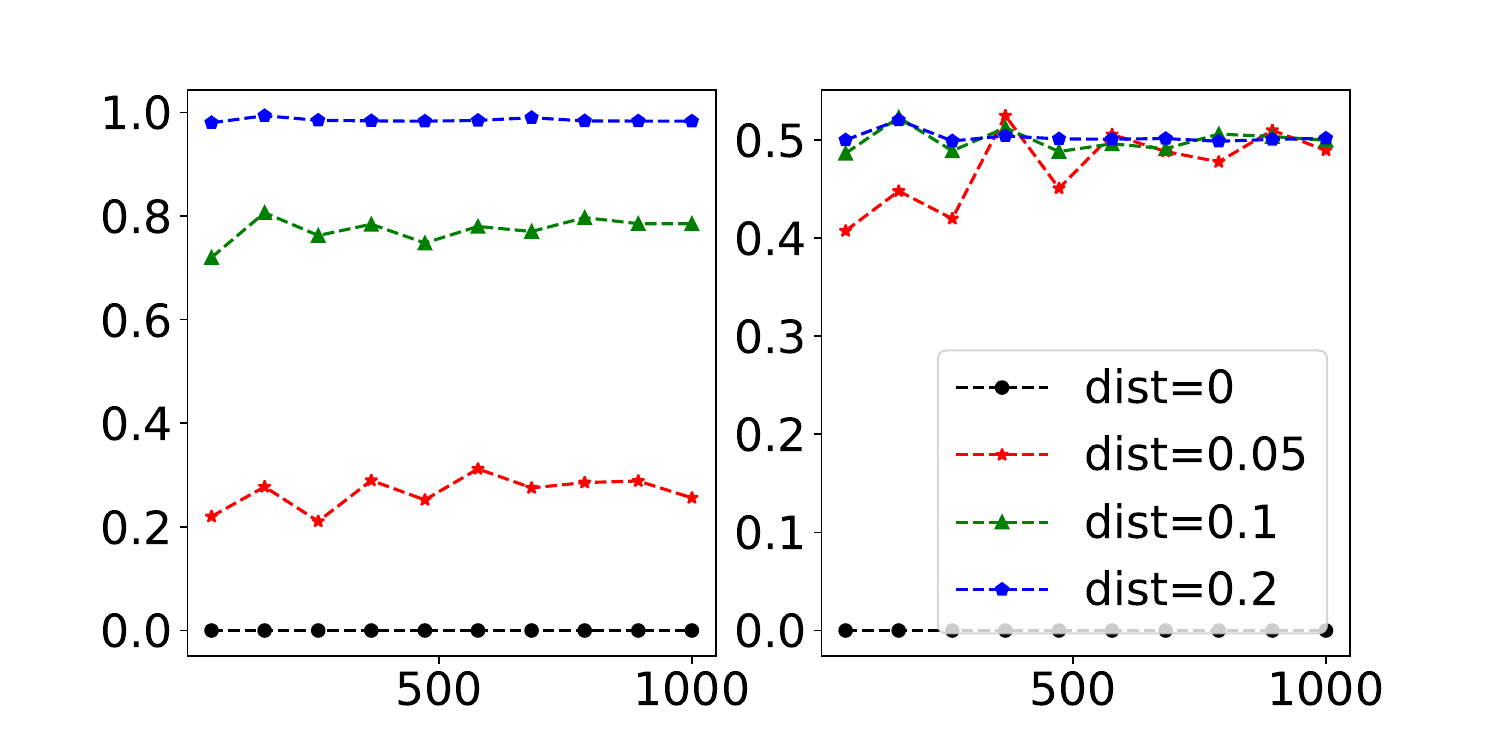}
    \vspace*{-8mm}
    \centering
    \caption{Recall (left) and precision (right) according to the number of real data that the attacker can access. The legend expressed Hamming distance threshold divided by the number of tokens.}
    \label{fig:MI}
\end{figure}
\vspace*{-8mm}

\section{Privacy evaluation}
\label{apd:MI}
Privacy evaluation is essential for assessing the quality of synthetic medical data.
We conducted membership inference that measures the privacy leakage risk by determining whether the data generated by the model is similar enough to be considered a member of the training dataset.

We generated 1,000 records and sampled $n_r$ records each for train data and test data.
The attacker has access to the real records of $2n_r$ and determines if each record was used for synthetic data creation via Hamming distance threshold between the two records. 
\figureref{fig:MI} shows recall and precision of membership inference for CNN-based VQ-VAE. 
Recall is the ratio from which members of a training set are correctly inferred as members by the attacker.
Precision represents how many of the samples inferred as members are actually in training data set.
Precision and recall are zero when Hamming distance threshold is zero, indicating that the synthetic data is not memorized by the model nor copied from the training data.
A precision around 0.5, regardless of the $n_r$, suggests the attacker has poor membership inference performance.
\vspace*{-4mm}

\section{t-SNE visualization for different perplexity values}
\label{apd:tsne_ppl}
\begin{figure*}[]
\floatconts
  {fig:tsneppl_mimic3}
  {
  \caption{t-SNE visualization in latent space on MIMIC-III dataset according to perplexiy (PPL).}
  \label{fig:tsneppl}
  }
  {%
    \subfigure[(C,C) (PPL=1)]{\label{11}%
      \includegraphics[width=.33\linewidth]{     
      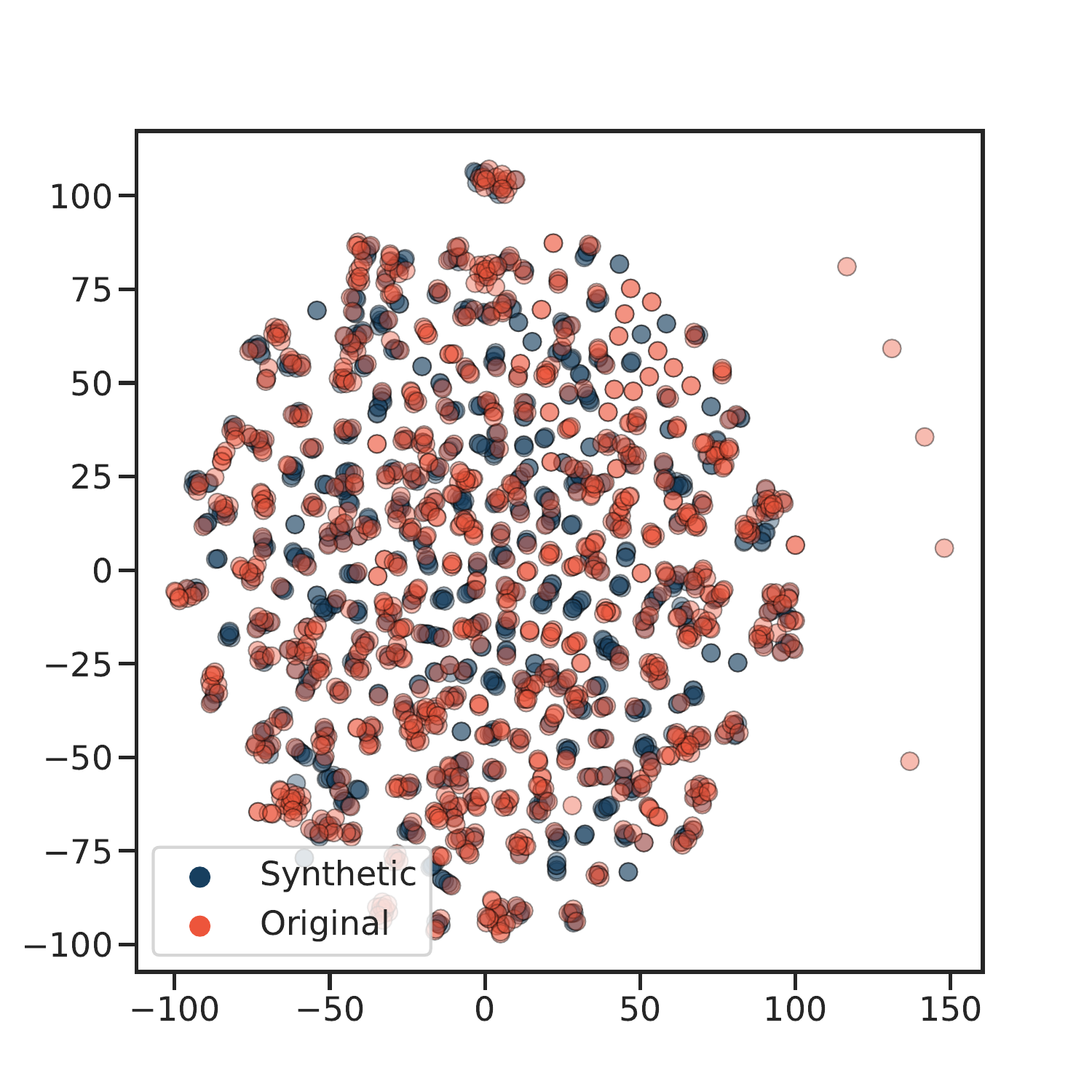}}%
    \subfigure[(T,C) (PPL=1)]{\label{12}
      \includegraphics[width=.33\linewidth]{     
    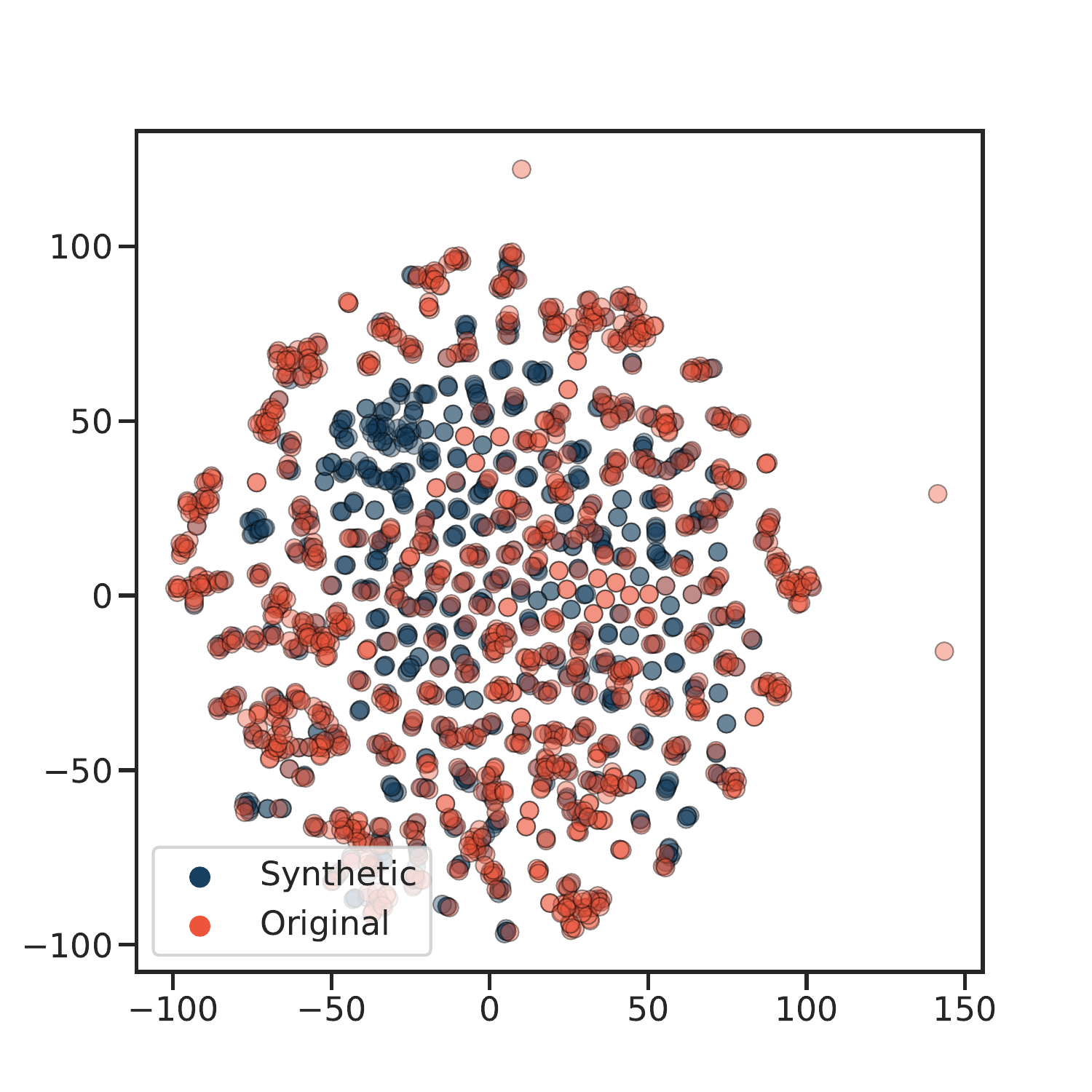}}%
    \subfigure[(T,T) (PPL=1)]{\label{}%
      \includegraphics[width=.33\linewidth]{
    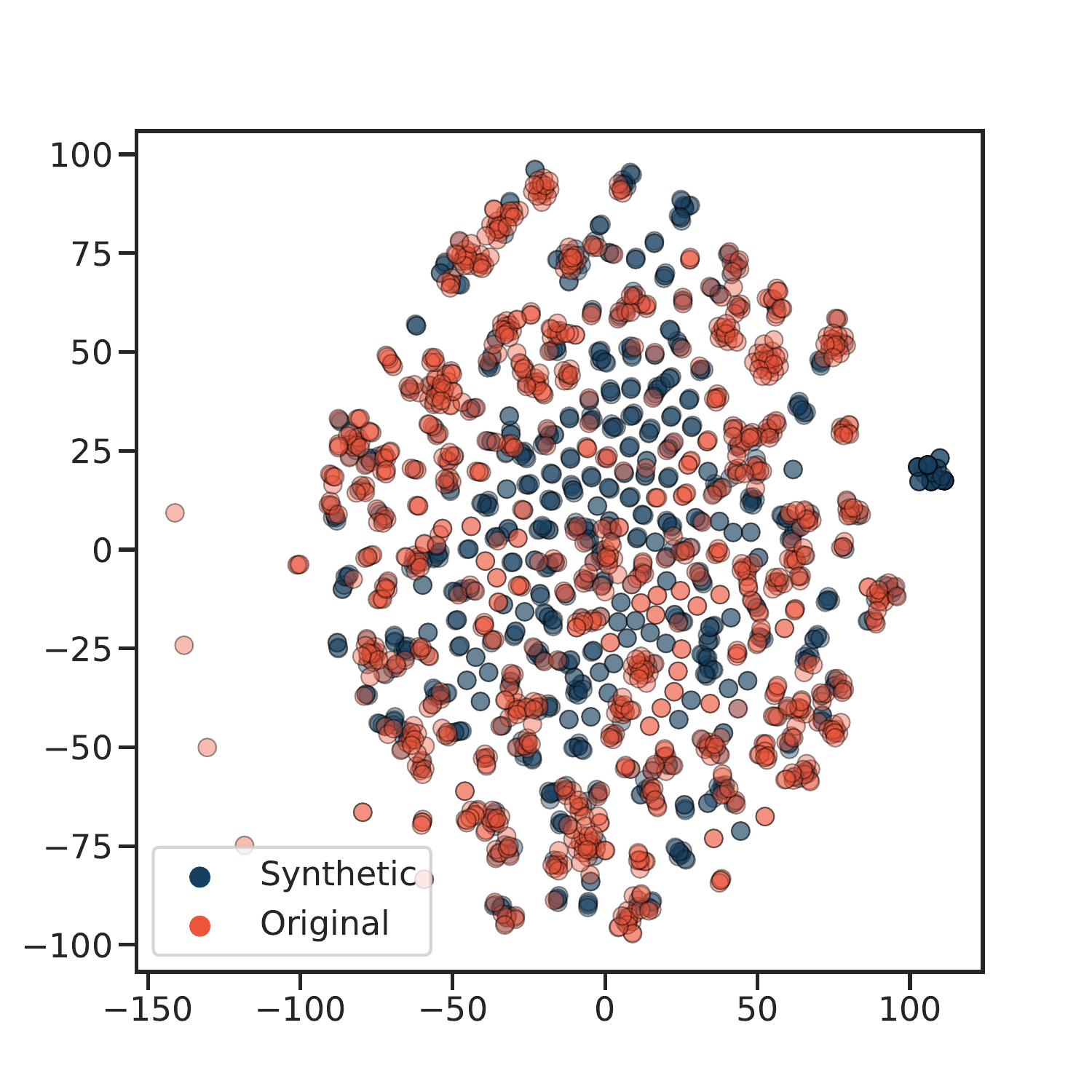}} \\
    \subfigure[(C,C) (PPL=5)]{\label{14}%
      \includegraphics[width=.33\linewidth]{     
      images/Method/genTsne/0_5_mimic3_2022_enc_conv_None_32768_3_3d.pdf}}%
    \subfigure[(T,C) (PPL=5)]{\label{15}
      \includegraphics[width=.33\linewidth]{     
    images/Method/genTsne/0_5_mimic3_2022_enc_transf_dec_conv_32768_3_3d}}%
    \subfigure[(T,T) (PPL=5)]{\label{}%
      \includegraphics[width=.33\linewidth]{
    images/Method/genTsne/0_5_mimic3_2022_enc_transf_None_32768_3_3d.pdf}} \\
    \subfigure[(C,C) (PPL=10)]{\label{17}%
      \includegraphics[width=.33\linewidth]{     
      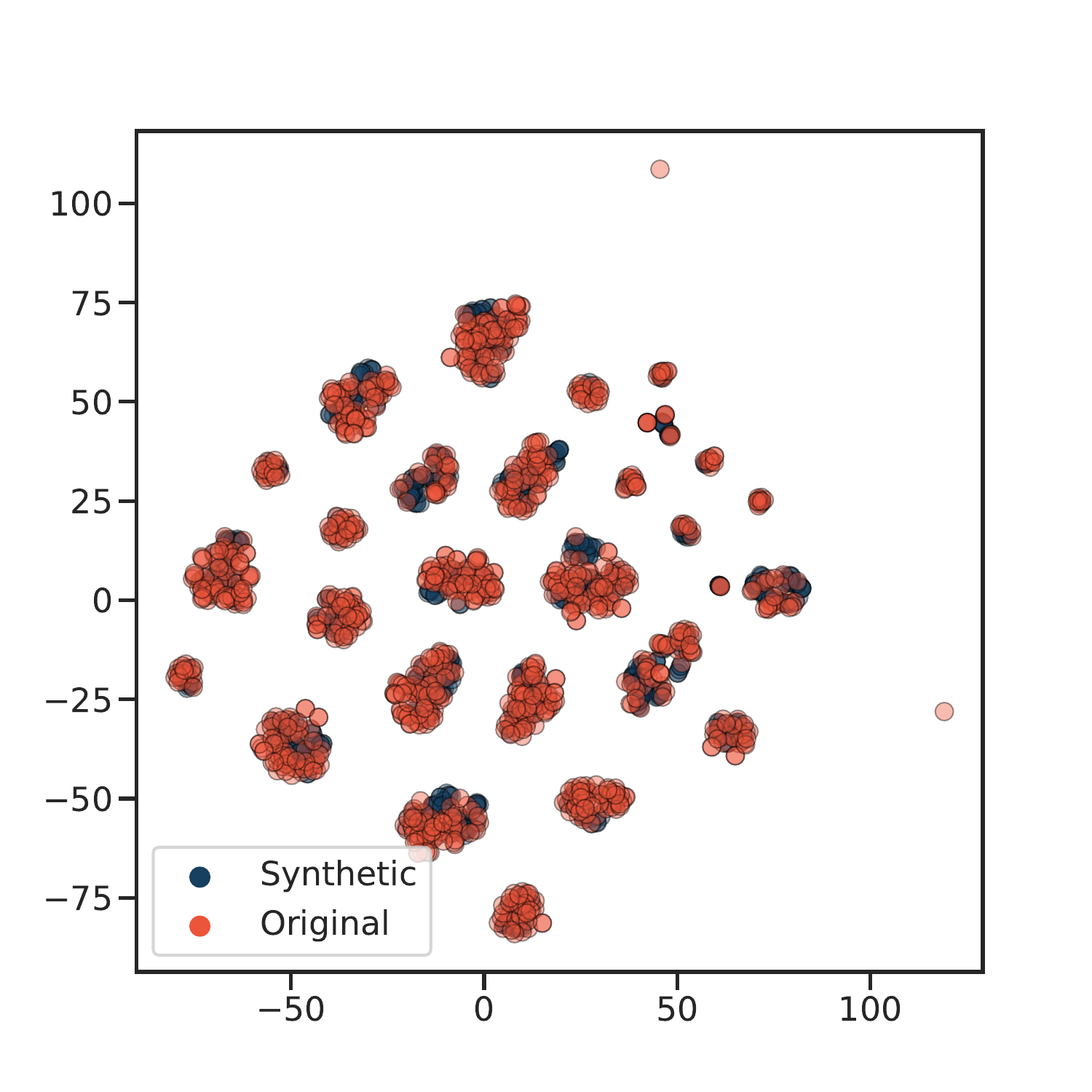}}%
    \subfigure[(T,C) (PPL=10)]{\label{18}
      \includegraphics[width=.33\linewidth]{     
    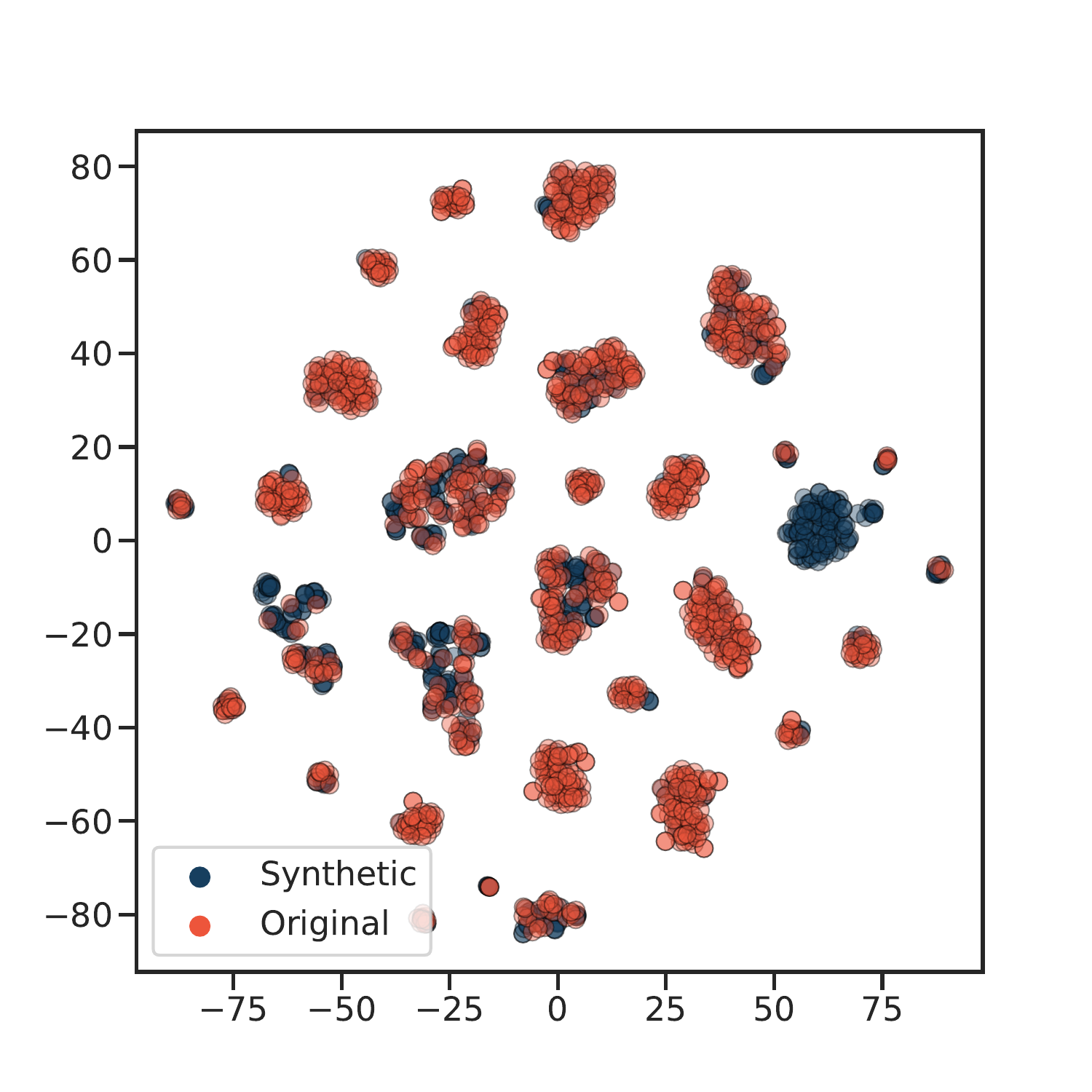}}%
    \subfigure[(T,T) (PPL=10)]{\label{}%
      \includegraphics[width=.33\linewidth]{
    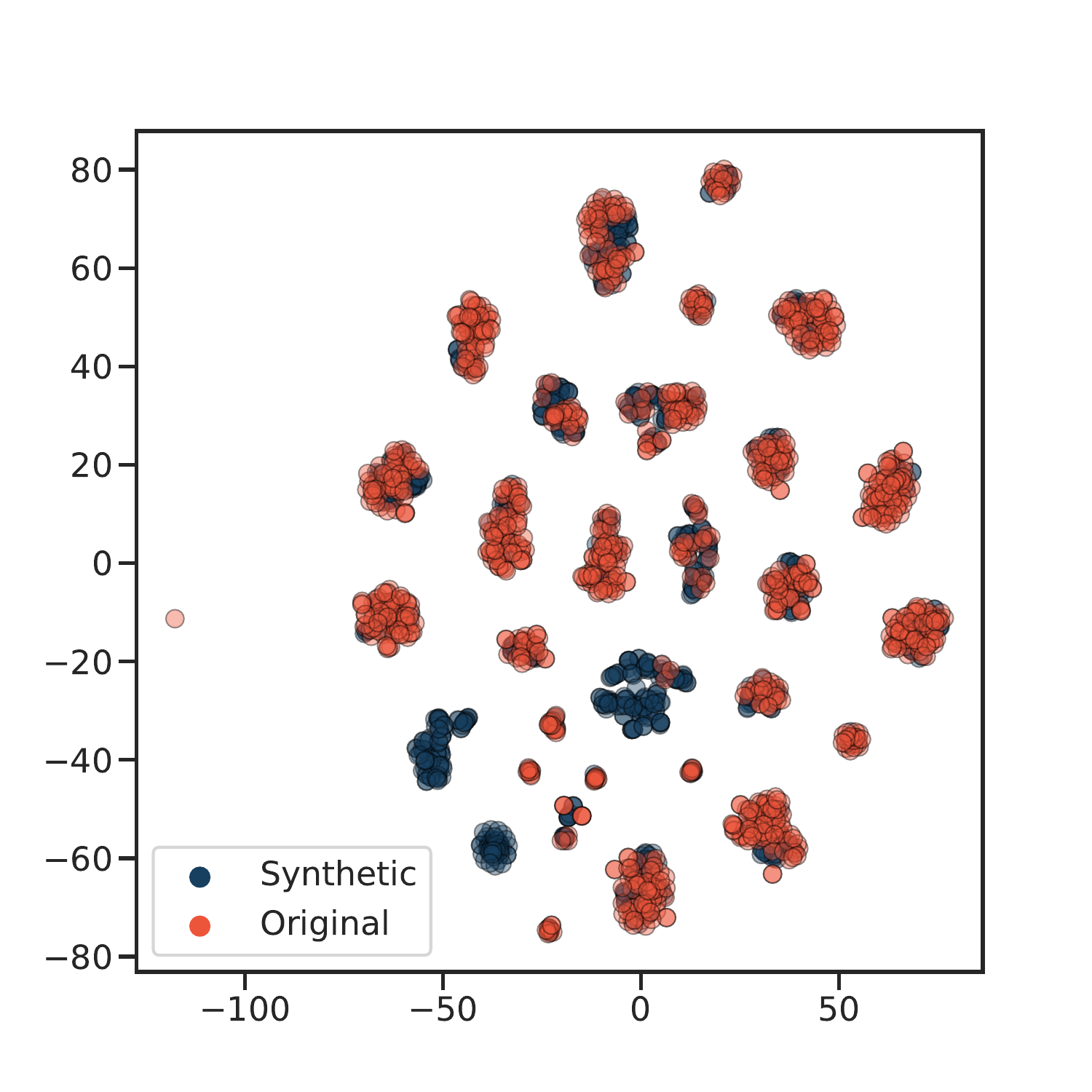}} 
  }
\end{figure*}

When using t-SNE for a qualitative evaluation, the interpretation may vary depending on the perplexity value, so we performed visualizations at various perplexities.
The results displayed in \figureref{fig:tsneppl} demonstrate that increasing perplexity led to tighter clustering of the data points.
Notably, the analysis conducted with multiple perplexity values yielded consistent results with those obtained in Section \ref{section:Gen}, which relied on a single perplexity value.

\begin{figure*}
\floatconts
  {fig:TrnasfDecDesign}
  {
  \caption{Transformer decoding scheme designs}
  \label{fig:TransfDec}
  }
  {%
    \subfigure[AR \& CA]{
    \label{fig:AR_CA}%
    \includegraphics[width=.3\textwidth]{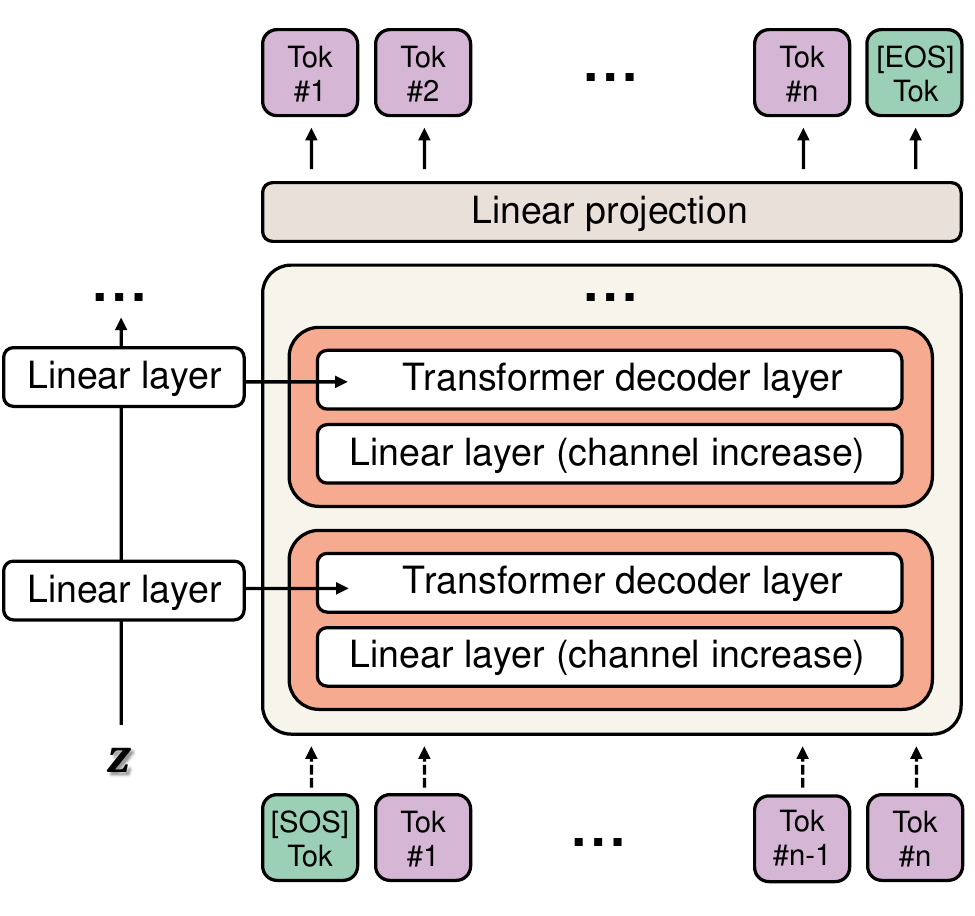}
    }%
    \subfigure[non-AR \& SA]{
    \label{fig:nonAR_SA}%
    \includegraphics[width=.3\textwidth]{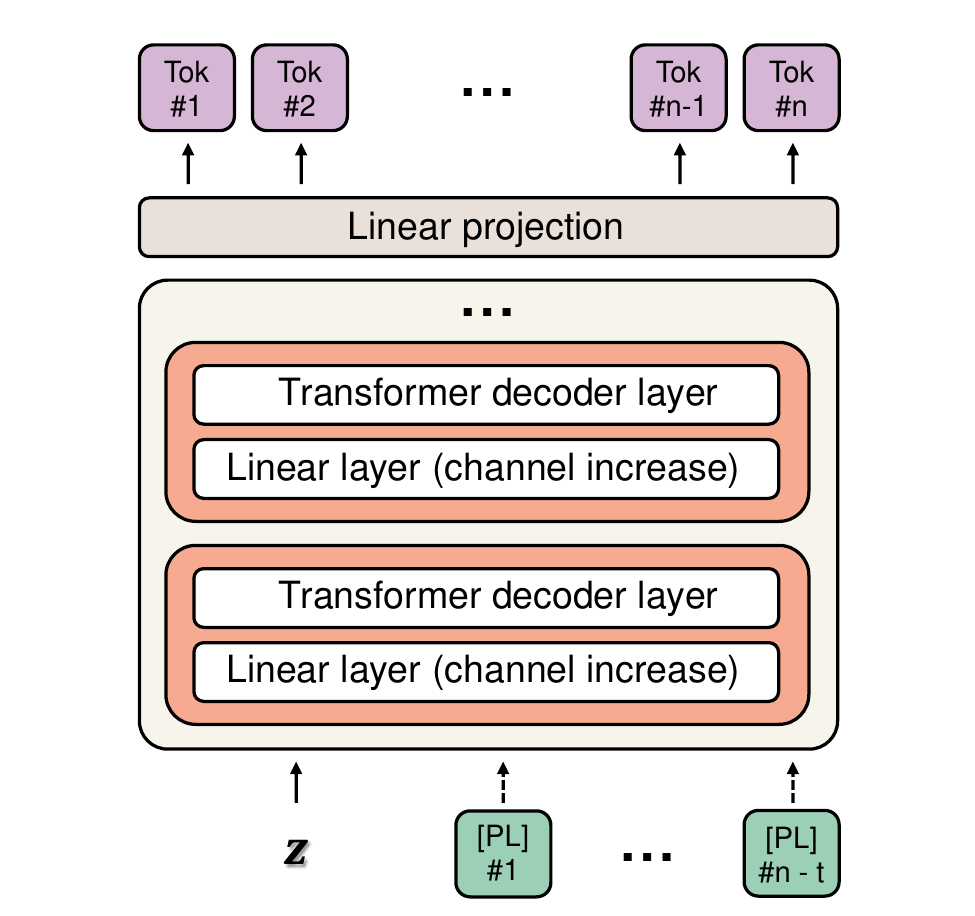}
    }
    \subfigure[non-AR \& SA+Unpool]{
    \label{fig:nonAR_SAU}
    \includegraphics[width=.3\textwidth]{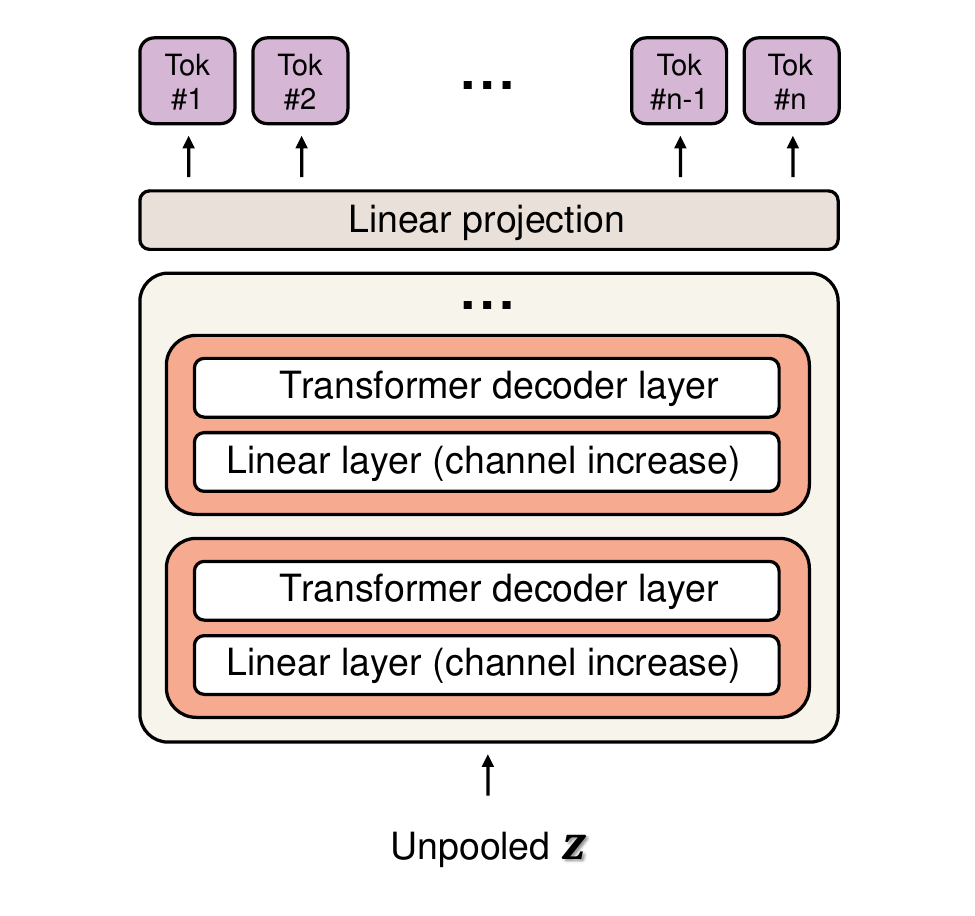}
    }%
    \vspace*{-3mm}
  }
\end{figure*}

\section{Transformer encoding schemes}
\label{apd:TransfEncoding}
As shown in \algorithmref{alg:Transformer}, our proposed Transformer encoding method gradually reduces the channel dimension as input $\textbf{x}$ is passed through encoding layers.
However, for Transformer-based encoding, it is common only to compress the temporal dimension (using [CLS] token or mean pooling) while leaving the channel dimension unchanged.
Thus, we experiment and compare the temporal dimension compression to our channel compression method for reconstruction and prediction with $\textbf{z}$ in 256 and 2048 dimensional spaces.

For $\textbf{z} \in \mathbb{R}^{1 \times 256}$, the Transformer encoder layers process $\textbf{x}$ and aggregate the temporal dimension into one by either extracting the [CLS] tokens or by mean pooling.

For $\textbf{z} \in \mathbb{R}^{1 \times 2048}$, the output of the Transformer encoder is linearly projected into 2048 dimensions and then temporally compressed.

For $\textbf{z} \in \mathbb{R}^{8 \times 256}$, we apply [CLS]-based and mean pooling-based compression with the following processes. 
For [CLS]-based compression, we insert 7 [CLS] tokens to the start of the sequence before encoding and subsequently use the 8 [CLS] tokens (i.e., encoded [CLS] output) for compression.
We apply adaptive average pooling to the encoded outputs for mean pooling-based compression.

\begin{figure*}[h]
\floatconts
  {fig:comparisonTrnasformerEncdoing}
  {\caption{Comparison of Transformer encoding schemes.}
  \label{apd:TransfDecoding}}
  {%
    \subfigure[Reconstruction ($l=256$)]{\label{30}%
      \includegraphics[width=.45\textwidth]{     
      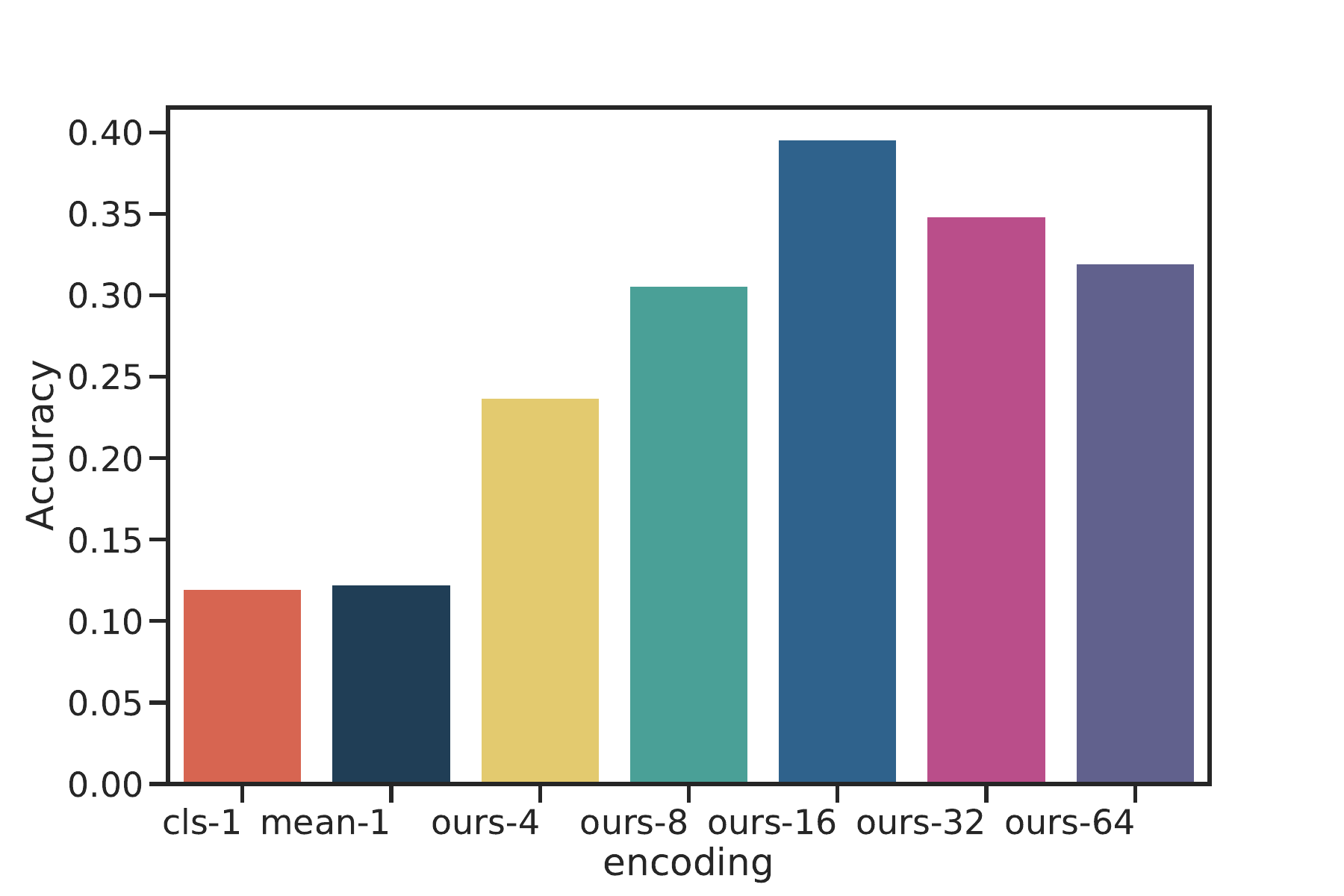}}%
    \subfigure[Prediction ($l=256$)]{\label{}%
      \includegraphics[width=.45\textwidth]{
      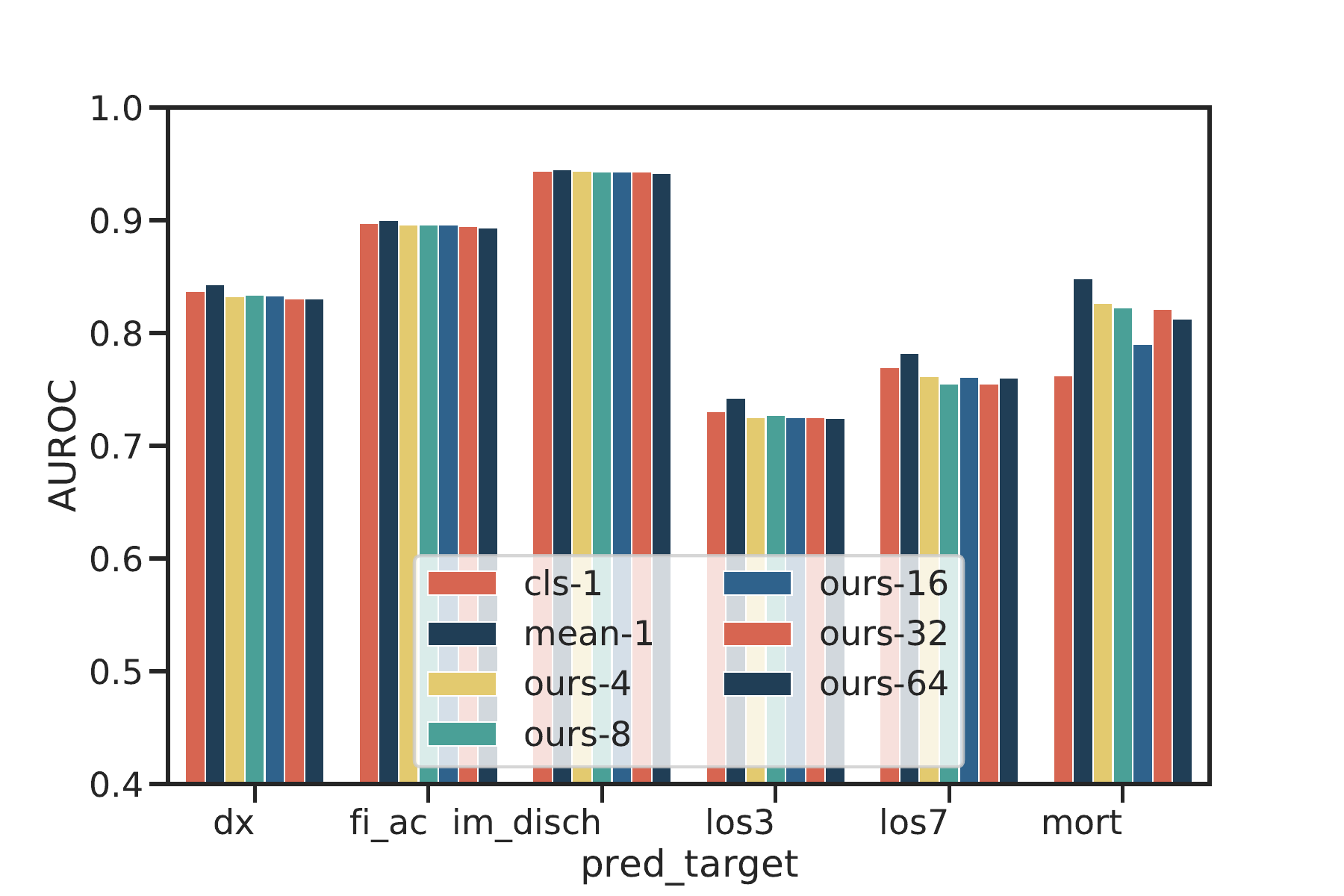}}
    \subfigure[Reconstruction ($l=2048$)]{\label{32}%
      \includegraphics[width=.45\textwidth]{     
      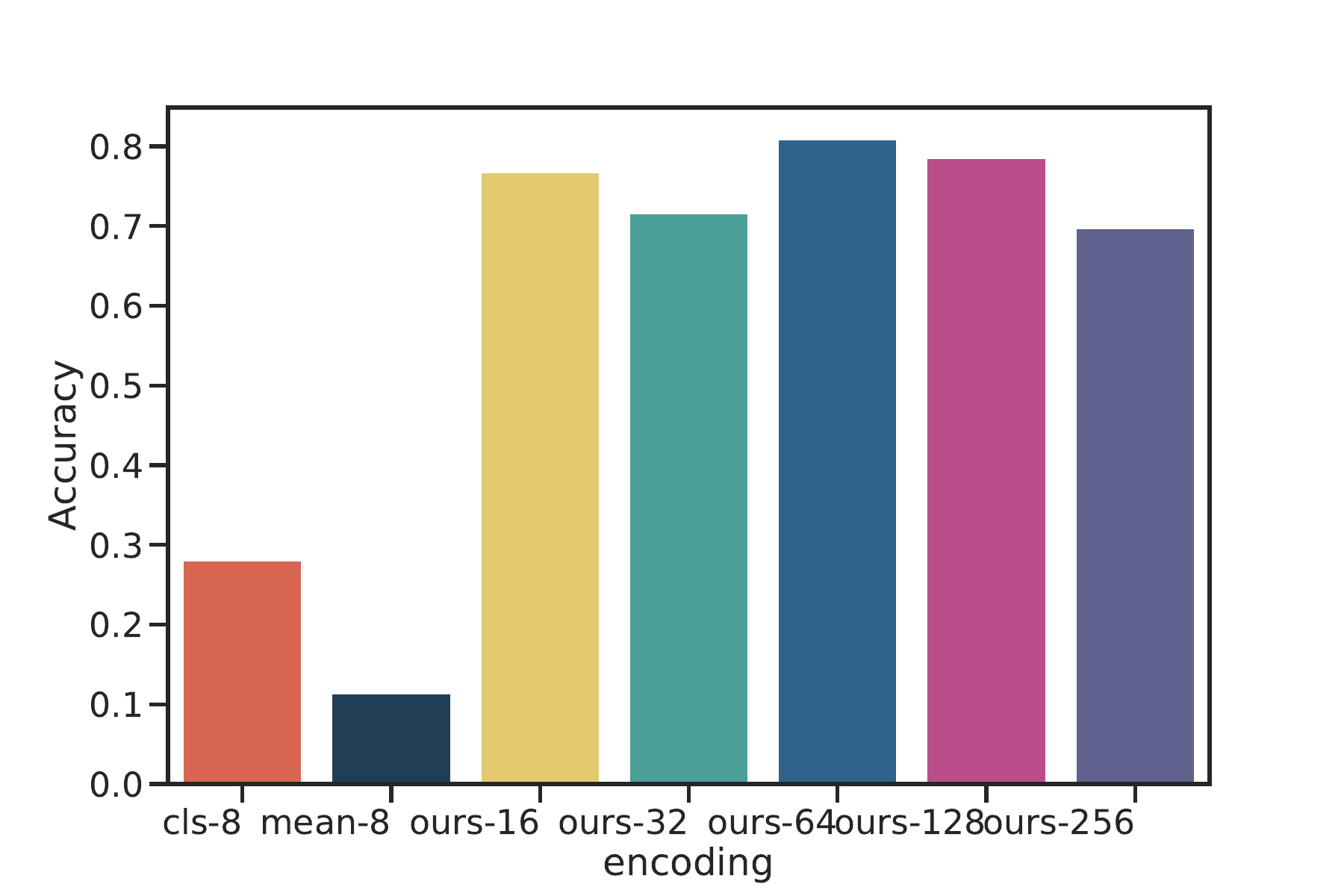}}%
    \subfigure[Prediction ($l=2048$)]{\label{}%
      \includegraphics[width=.45\textwidth]{
      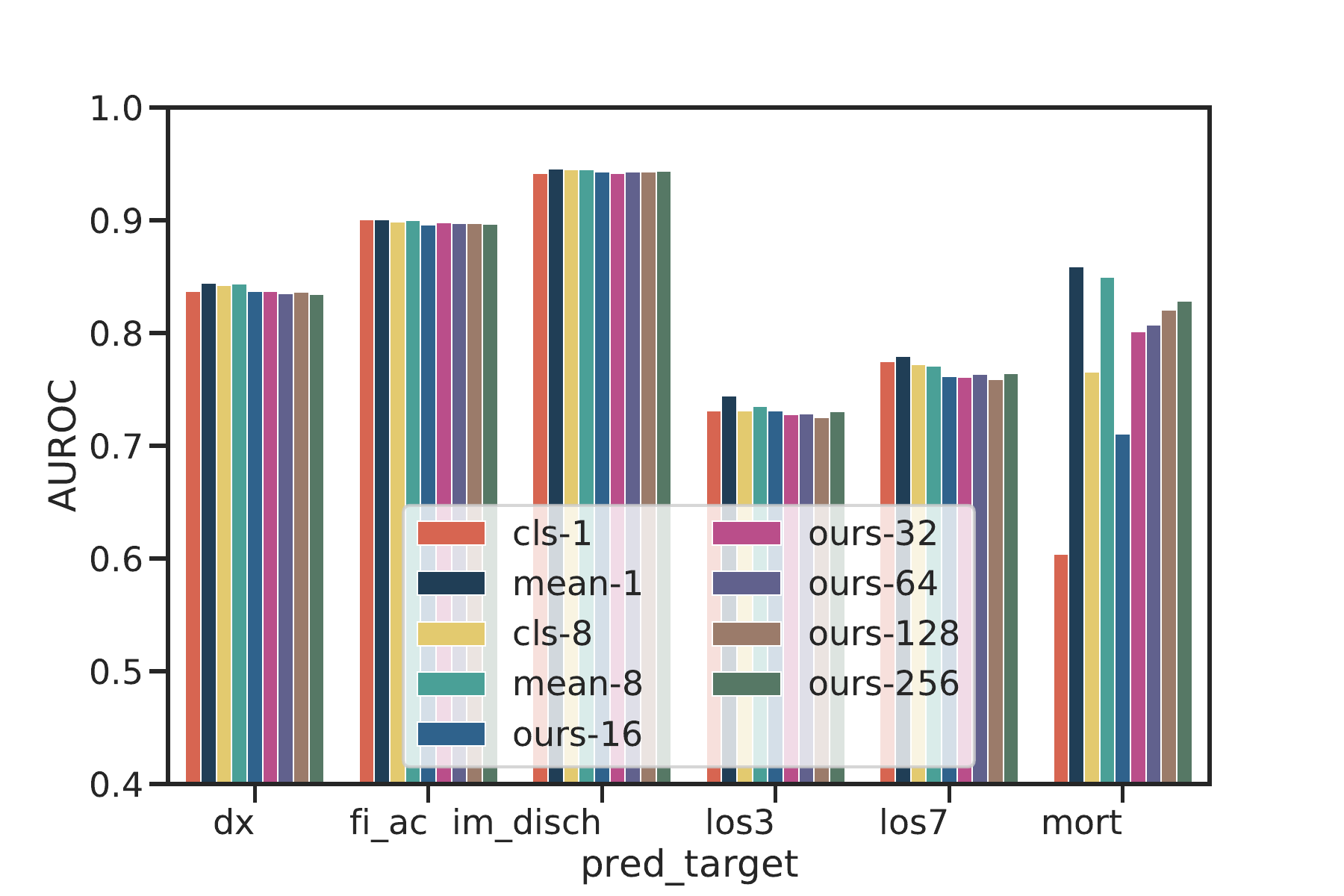}}
  }
\end{figure*}

\figureref{apd:TransfDecoding} shows the results of reconstruction and prediction. 
For the reconstruction tasks, our encoding scheme outperforms all existing methods with a significant gap in accuracy for both $\textit{l}$ = 256 and $\textit{l}$ = 2048. 
Our method also shows comparable AUROC but slightly lower than mean pooling and higher AUROC than using [CLS].

\section{Transformer decoding schemes}
\label{apd:TransfDec}

We experiment on four different Transformer decoding schemes (shown in \figureref{fig:TransfDec}), including the setting in \figureref{fig:transformerdecoder}, where both the encoder and decoder are Transformers.
After $\textbf{z}$ is generated from the encoder, the decoder reconstructs input $\textbf{x}$ by using either autoregressive or non-autoregressive methods.
The autoregressive method generates $\textbf{x}$ sequentially based on tokens from previous time steps, while the non-autoregressive method generates $\textbf{x}$ at once.

For non-autoregressive decoding, $\textbf{z}$ can be passed to the decoder either indirectly as a key for cross-attention (\sectionref{subsec:TaskSpecificModels}) or directly for self-attention.

We use two different types of decoder inputs for self-attention.
The first input type is a vector obtained by unpooling $\textbf{z}$ to the length of input $\textbf{x}$, followed by adding positional encoding.
Unpooling replicates adaptive mean pooling values for each pooling window by the size of the window.
Also, positional encoding was added to overcome Transformer's inability to distinguish tokens with identical values of different positions.
The other input type is a vector of length $\textbf{x}$, created by concatenating placeholder tokens to $\textbf{z}$.

\begin{figure}
    \includegraphics[width=0.7\linewidth]{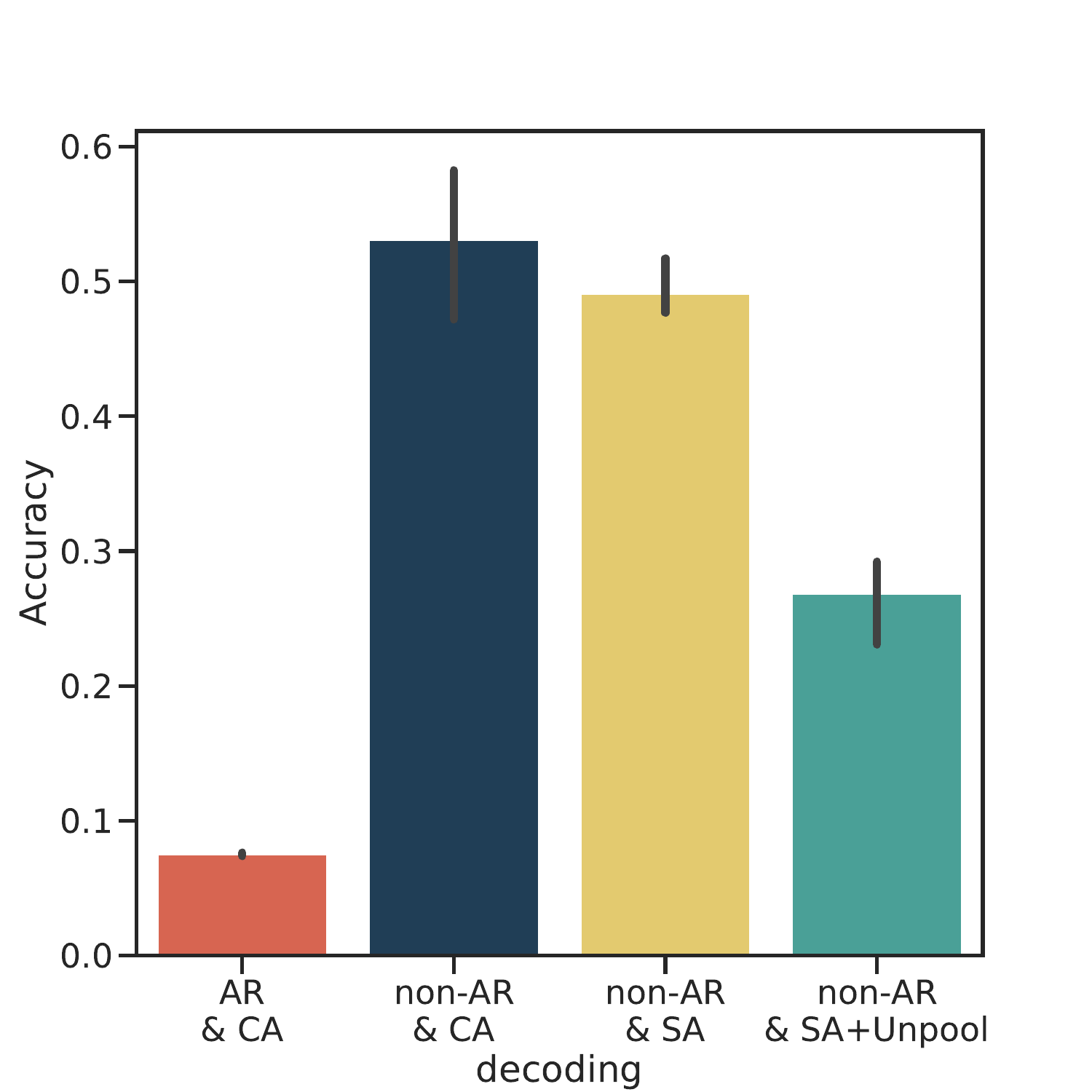}
    \vspace*{-5mm}
    \centering
    \caption{Reconstruction accuracy of various Transformer decoding schemes.}
    \label{fig:TransfDecoding}
\end{figure}

As shown in \figureref{fig:TransfDecoding}, autoregressive decoding shows the lowest reconstruction accuracy and does not appear to be suitable for reconstructing long sequences.
Among the non-autoregressive decoding methods, the method of passing information of $\textbf{z}$ from the encoder to the decoder through cross-attention shows the most efficient reconstruction.

\section{Visualization and analysis of self-attention maps in Transformer}
\label{apd:LocalEHR}

\begin{figure*}[]
\floatconts
  {fig:selfattentionmap}  
  {%
    \subfigure[Reconstruction]{\label{fig:sarecon}
      \includegraphics[width=\textwidth]{     
      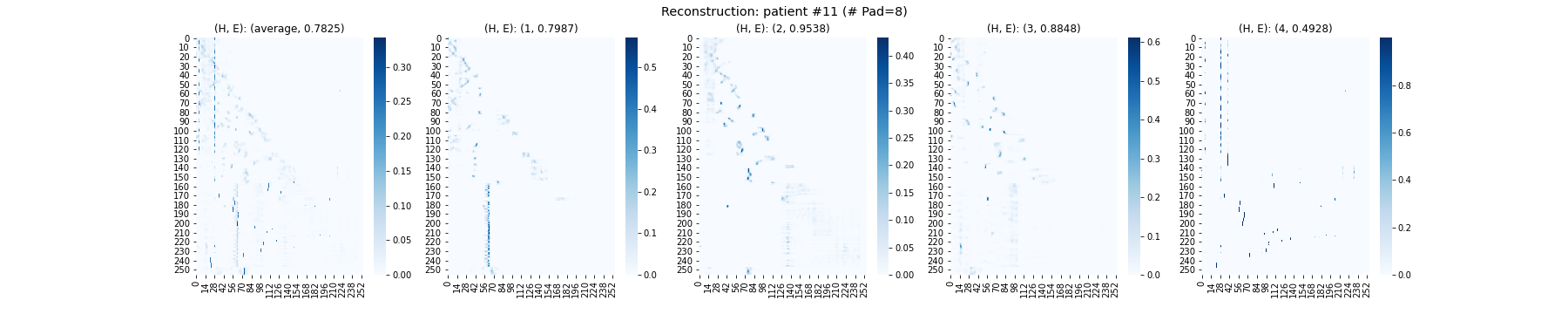}
      }
    \subfigure[Prediction-dx]{\label{fig:sapreddx}
      \includegraphics[width=\textwidth]{
      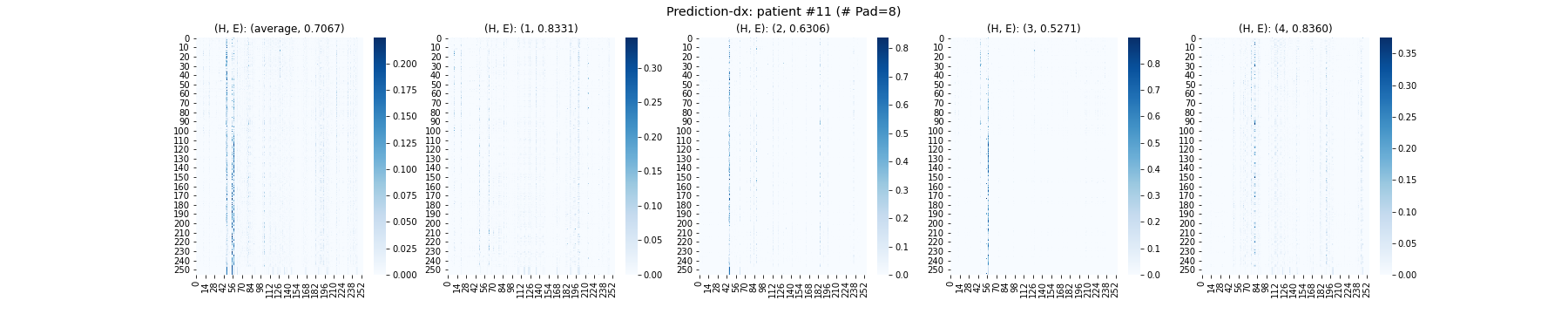}
      }
    \subfigure[Prediction-mort]{\label{fig:sapredmort}
      \includegraphics[width=\textwidth]{
      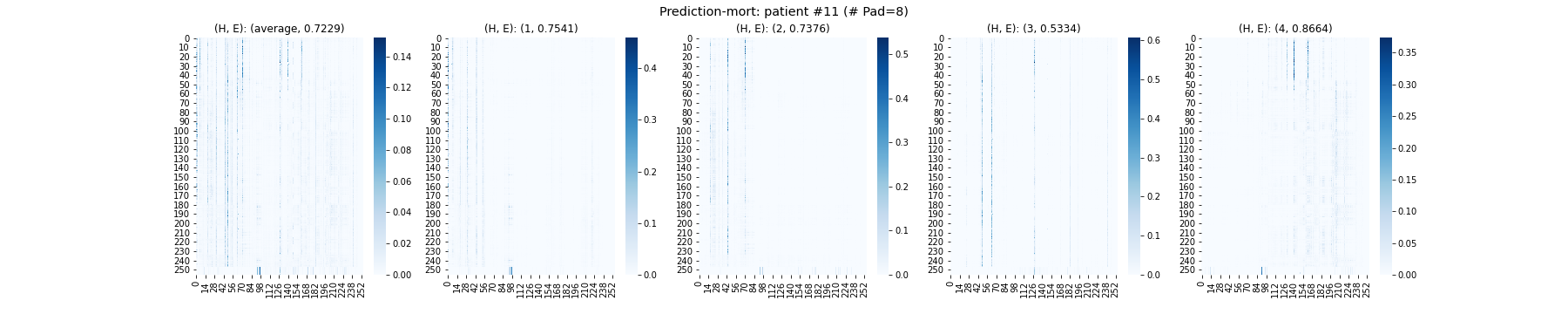}
      }
  }
    {
  }
  \caption{Self-attention heat-map of the $\Ence$ in the hierarchically-structured Transformer.
  In $\Ence$, the input for the patient has 256 medical events as query and key vectors. 
  y and x axes in the figure respectively mean key and query event vector in each head.
  Query event vectors shows completely different pattern in reconstruction and prediction case.
  \label{fig:sa}
  }
\end{figure*}

By visualizing the self-attention map of the Transformer layer having a hierarchical structure, we analyze how the reconstruction and prediction task differ from each other.
Specifically, we choose a patient who has less than ten pad-events, which are fully filled with pad tokens, out of $n_{e}=256$ events.
Then, we draw a heat-map for the first self-attention layer of the $\Ence$ in the autoencoder $(T,T)$ and the predictive model $\mathtt{Classifier}(\mathtt{Enc}_{struc}(\mathbf{x}))$. 
The patient's key and query medical events respectively correspond to the y and x axes in the \figureref{fig:sa}.

As shown in the figure, Transformer shows an entirely different pattern in cases of reconstruction and prediction.
For the reconstruction task illustrated in Figure \ref{fig:sarecon}, most of the attention heads in the Transformer mainly attend temporally proximal events represented by the elements near the diagonal line.
However, almost every query event attends specific events out of all $n_{e}$ events to perform the clinical prediction task, as depicted both in \figureref{fig:sapreddx} and \ref{fig:sapredmort}.

\section{Results of each prediction task}
\label{apd:Pred6Tasks}

\begin{figure*}[!hbt]
\floatconts
  {fig:prediction_arranged_by_latent}
  {\caption{Prediction performances on MIMIC-III arranged by the latent dimension $l$}
  \label{fig:pred6Tasks-mimic-latent}}
  {%
    \subfigure[Diagnosis]{\label{idle}
      \includegraphics[width=.45\textwidth]{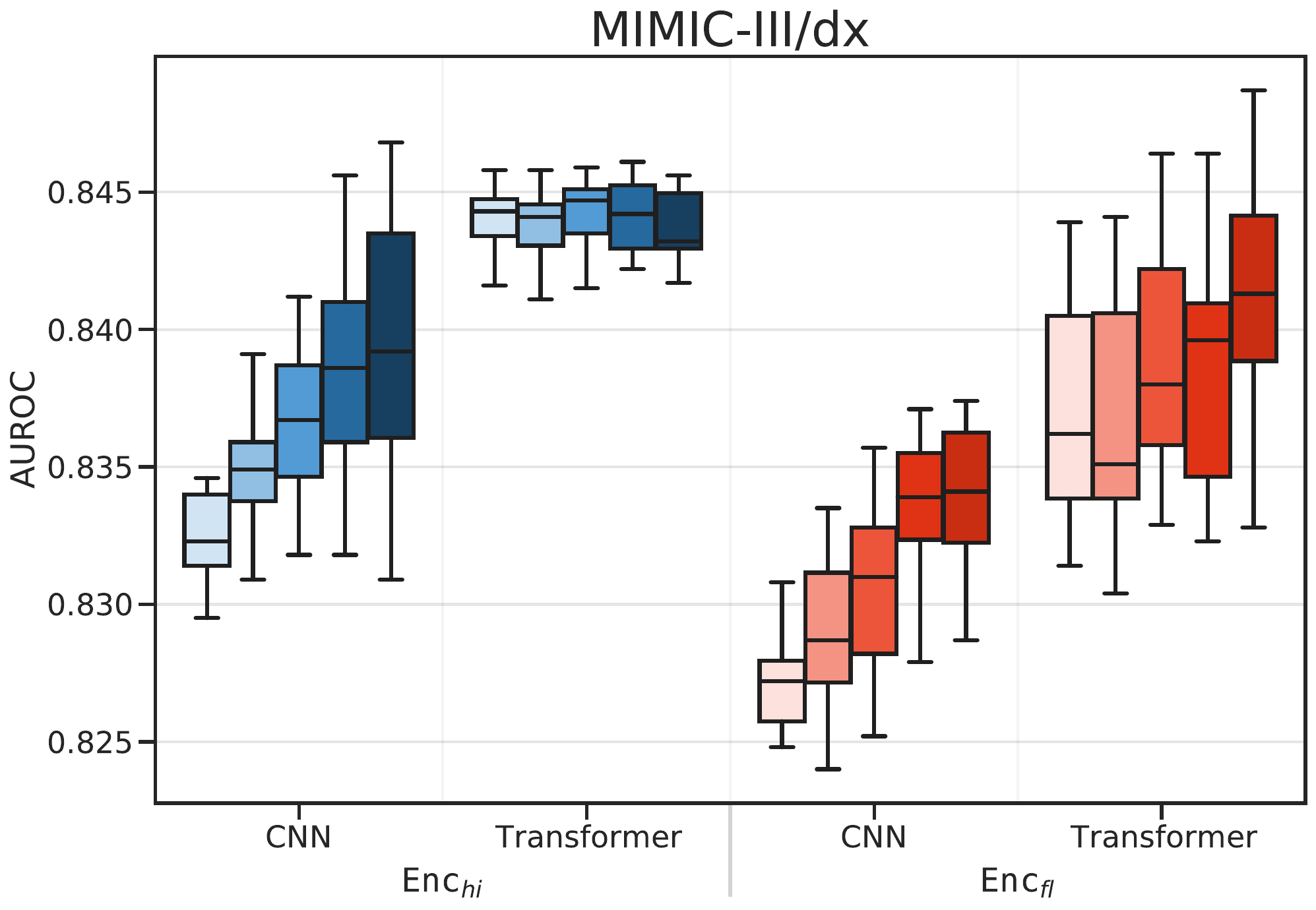}}%
    \subfigure[Final acuity]{\label{idle}
      \includegraphics[width=.45\textwidth]{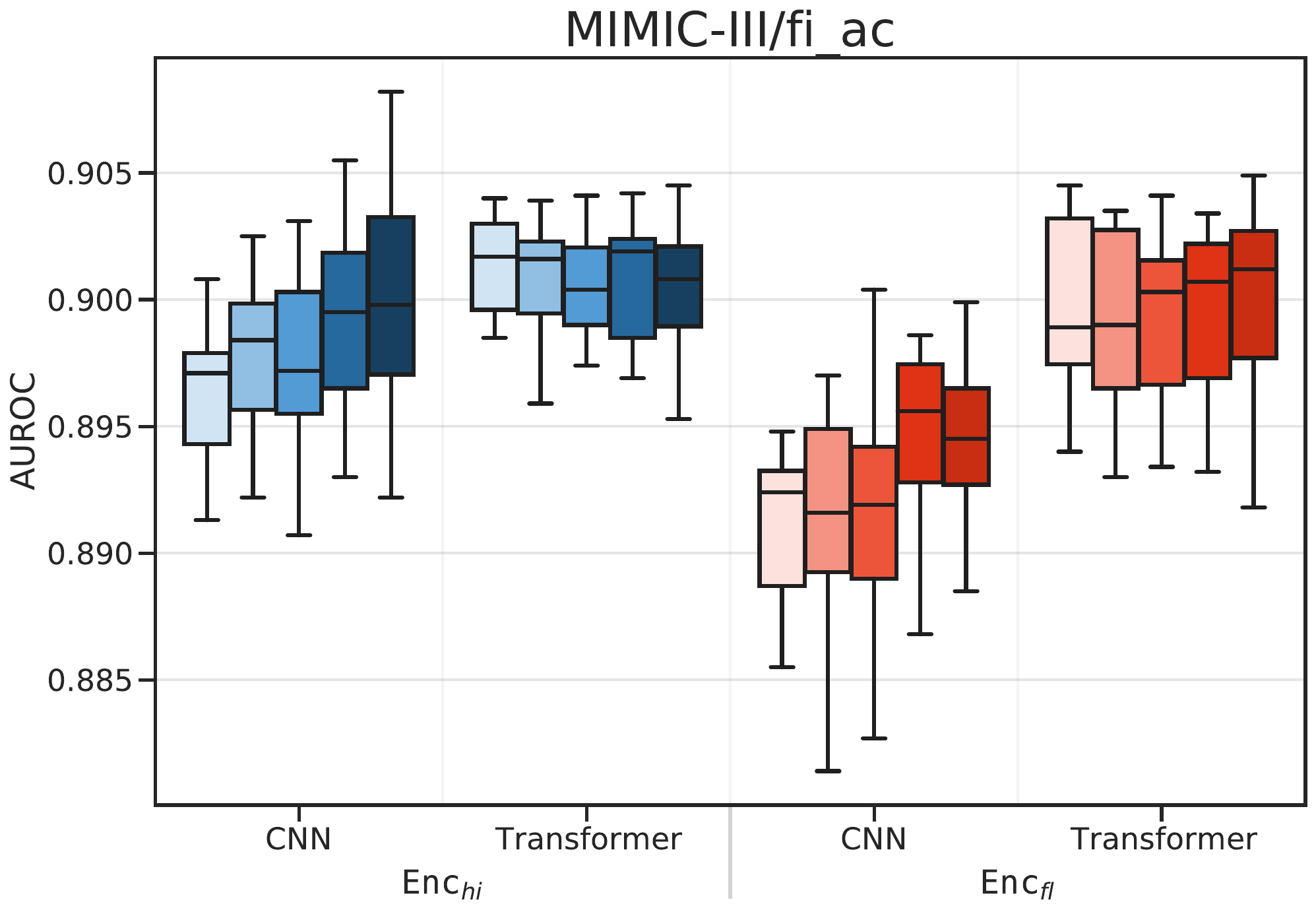}}
    \subfigure[Imminent Discharge]{\label{idle}
      \includegraphics[width=.45\textwidth]{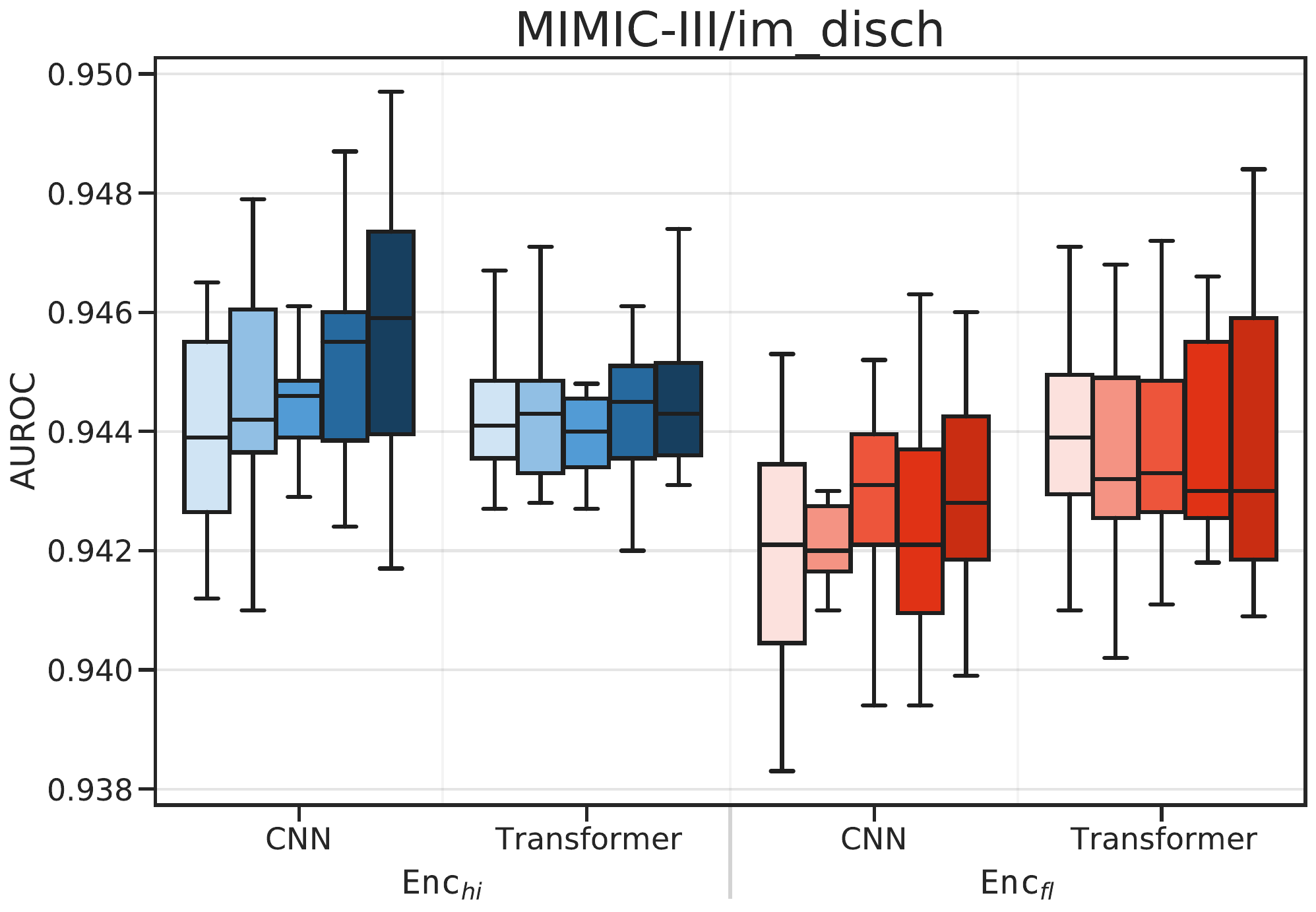}}%
    \subfigure[Length-Of-Stay for case of three days]{\label{idle}
      \includegraphics[width=.45\textwidth]{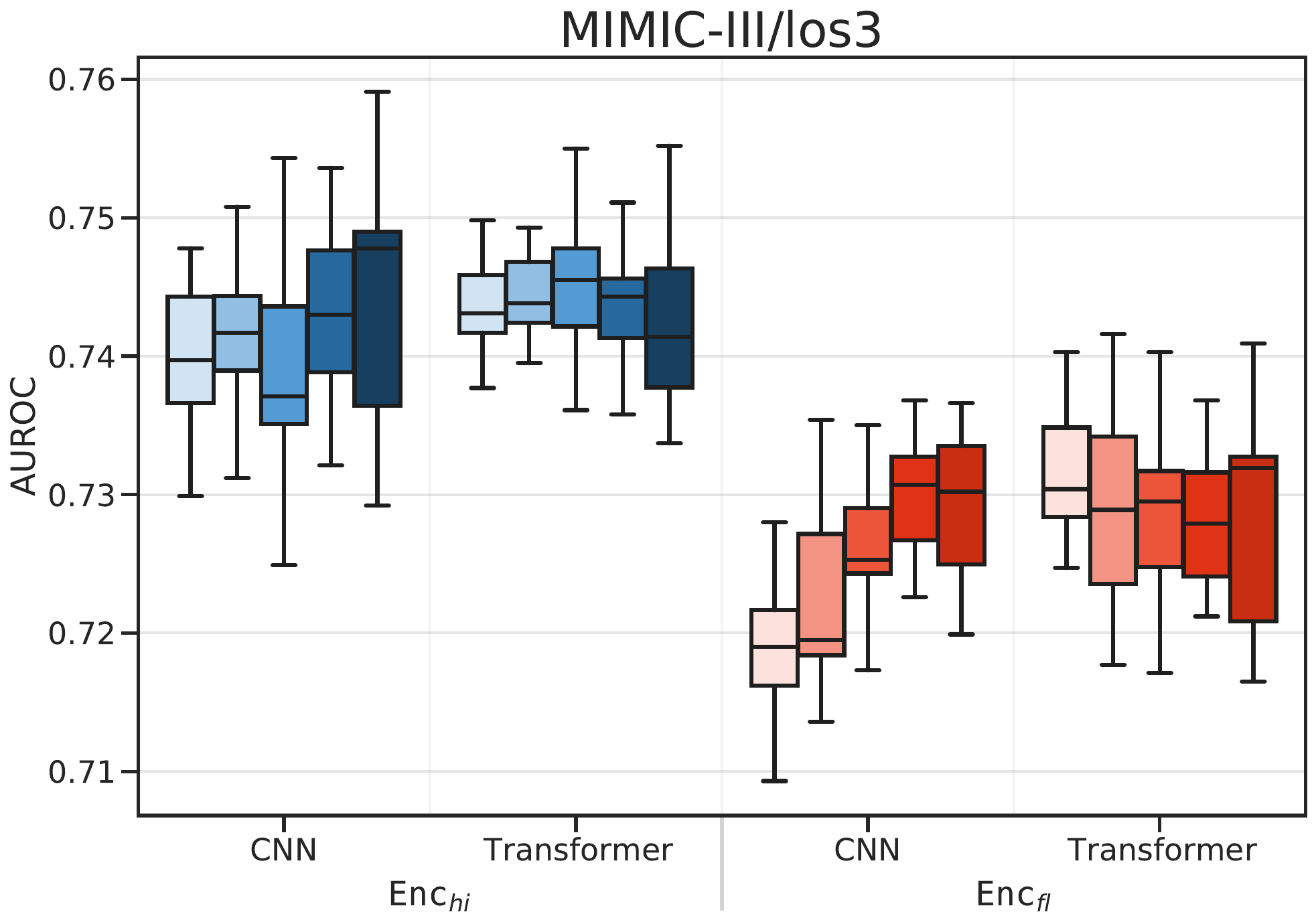}}
    \subfigure[Length-Of-Stay for case of seven days]{\label{idle}
      \includegraphics[width=.45\textwidth]{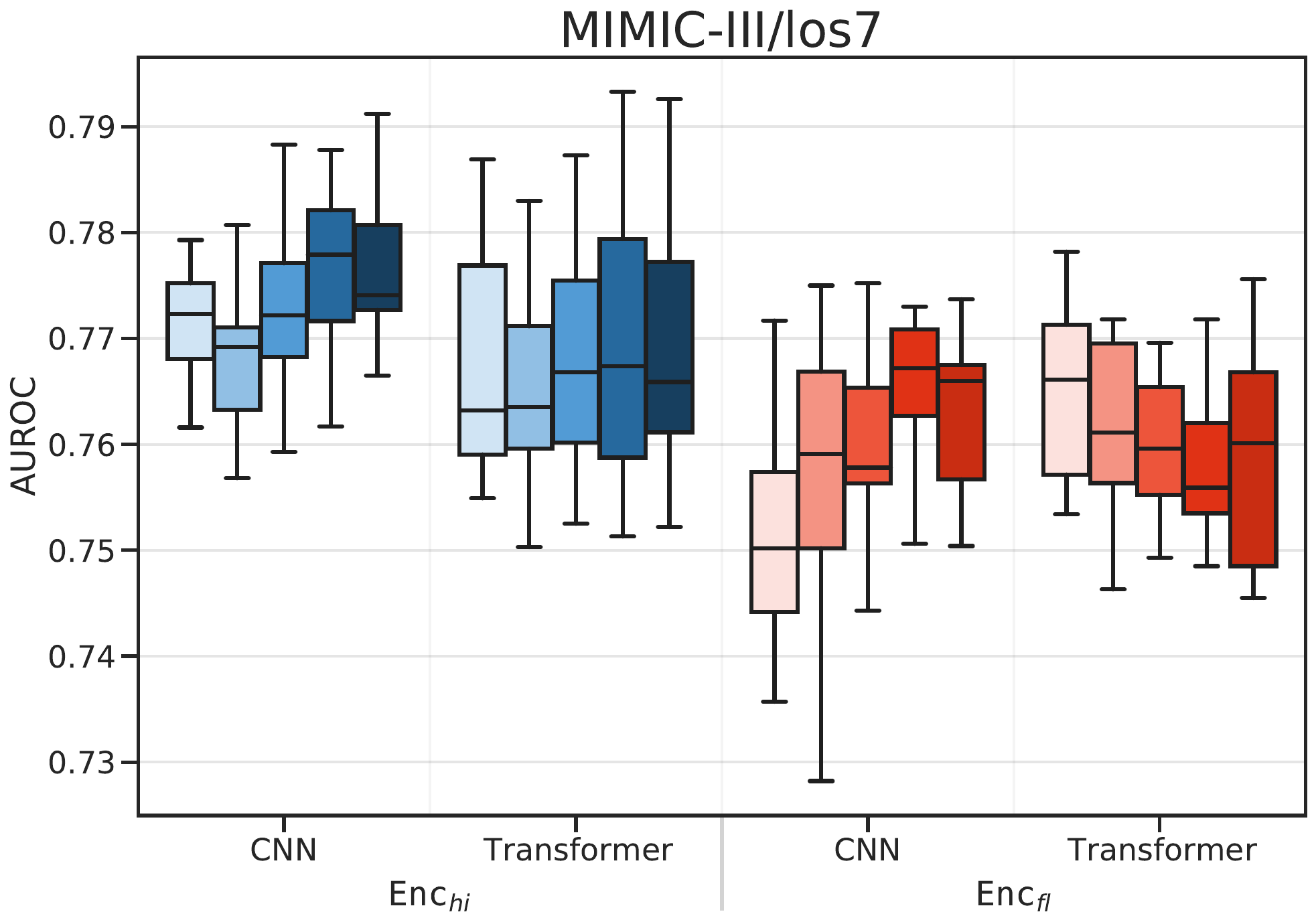}}%
    \subfigure[Mortality]{\label{idle}
      \includegraphics[width=.45\textwidth]{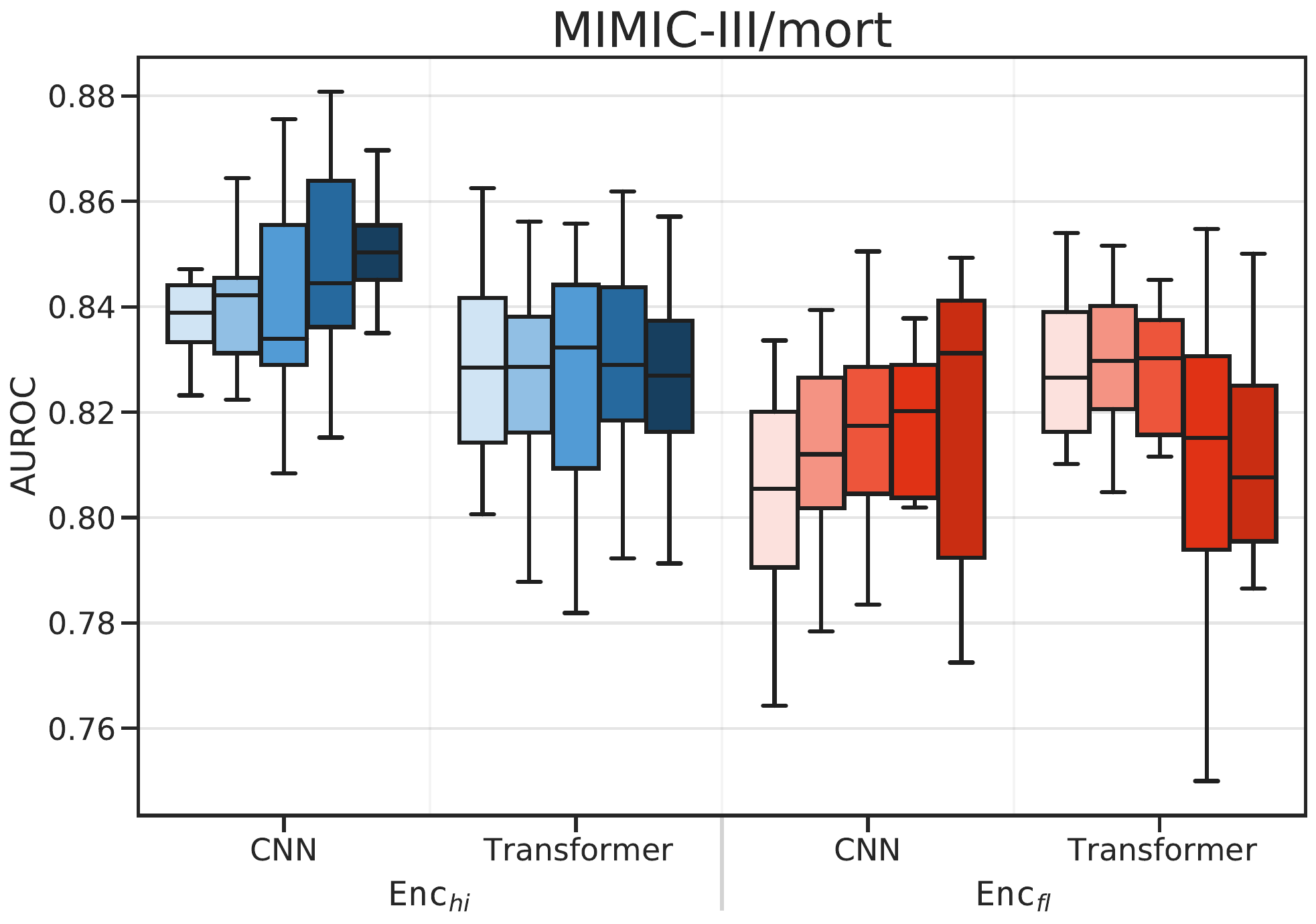}
      }
  }
\end{figure*}

\begin{figure*}[!h]
\floatconts
  {fig:prediction_aranged_by_temporal}
  {\caption{Prediction performances on MIMIC-III arranged by the temporal dimension $t$}
  \label{fig:pred6Tasks-mimic-temporal}}
  {%
    \subfigure[Diagnosis]{\label{}%
      \includegraphics[width=.45\textwidth]{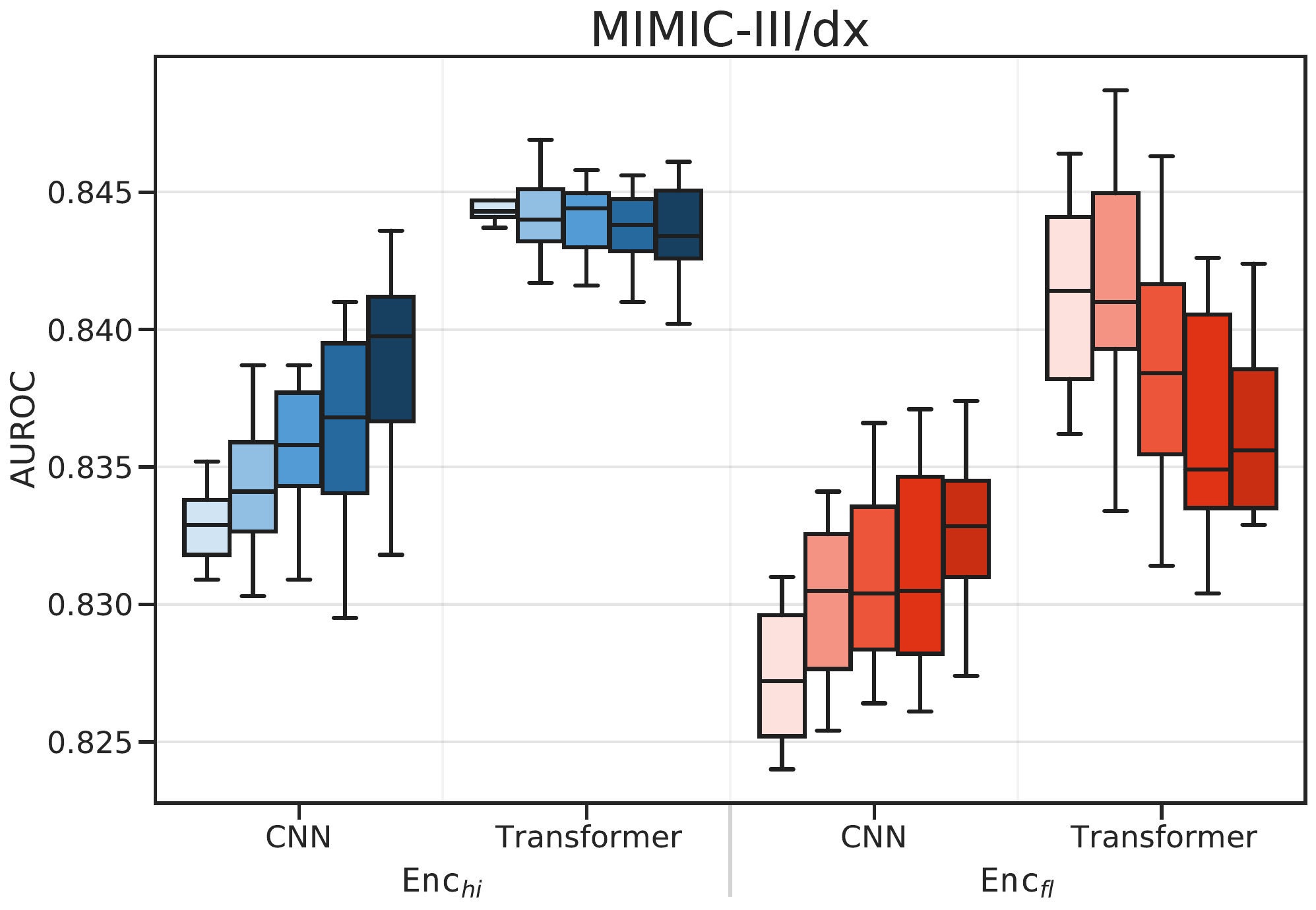}}%
    \subfigure[Final acuity]{\label{}%
      \includegraphics[width=.45\textwidth]{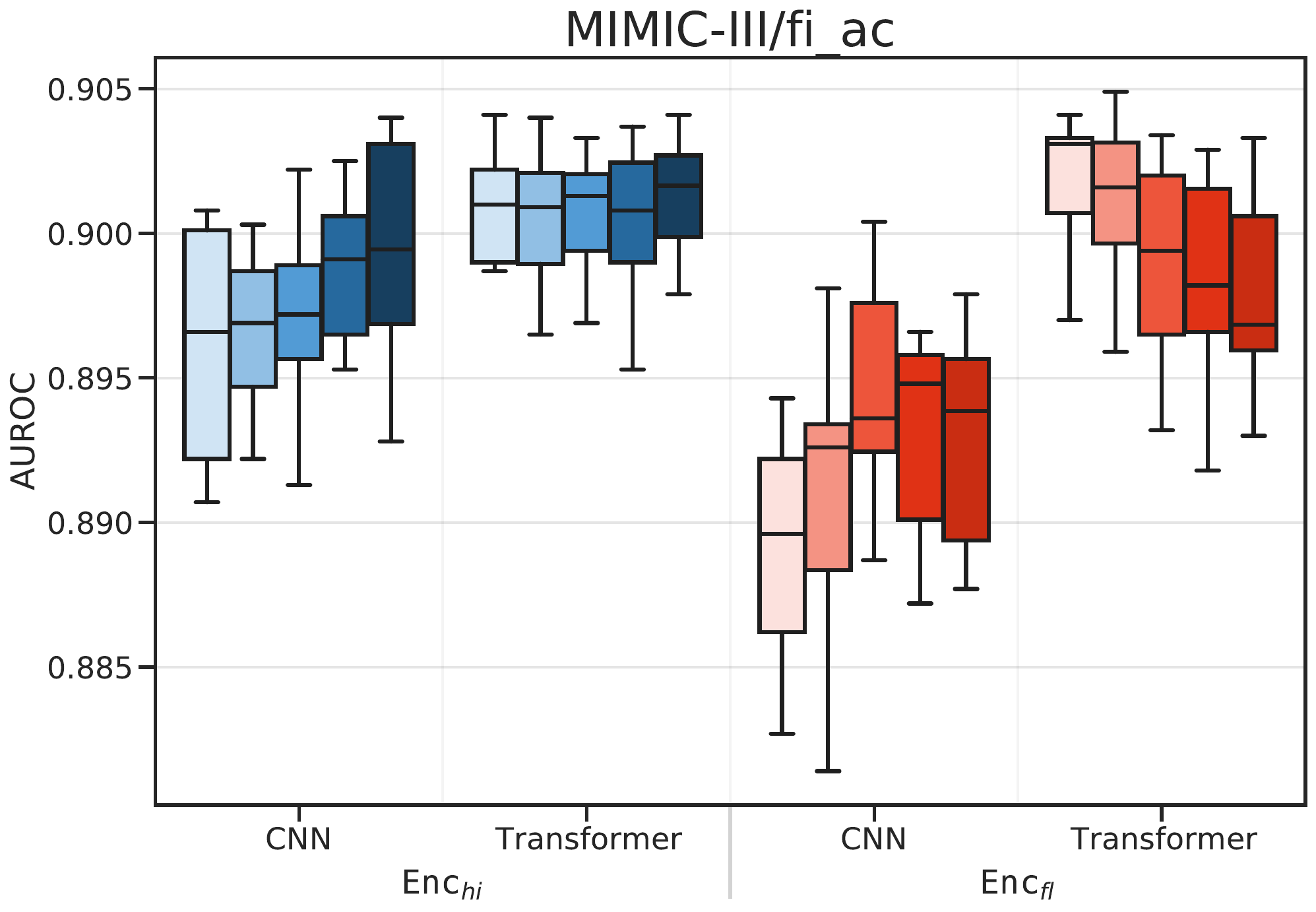}}
    \subfigure[Imminent Discharge]{\label{}%
      \includegraphics[width=.45\textwidth]{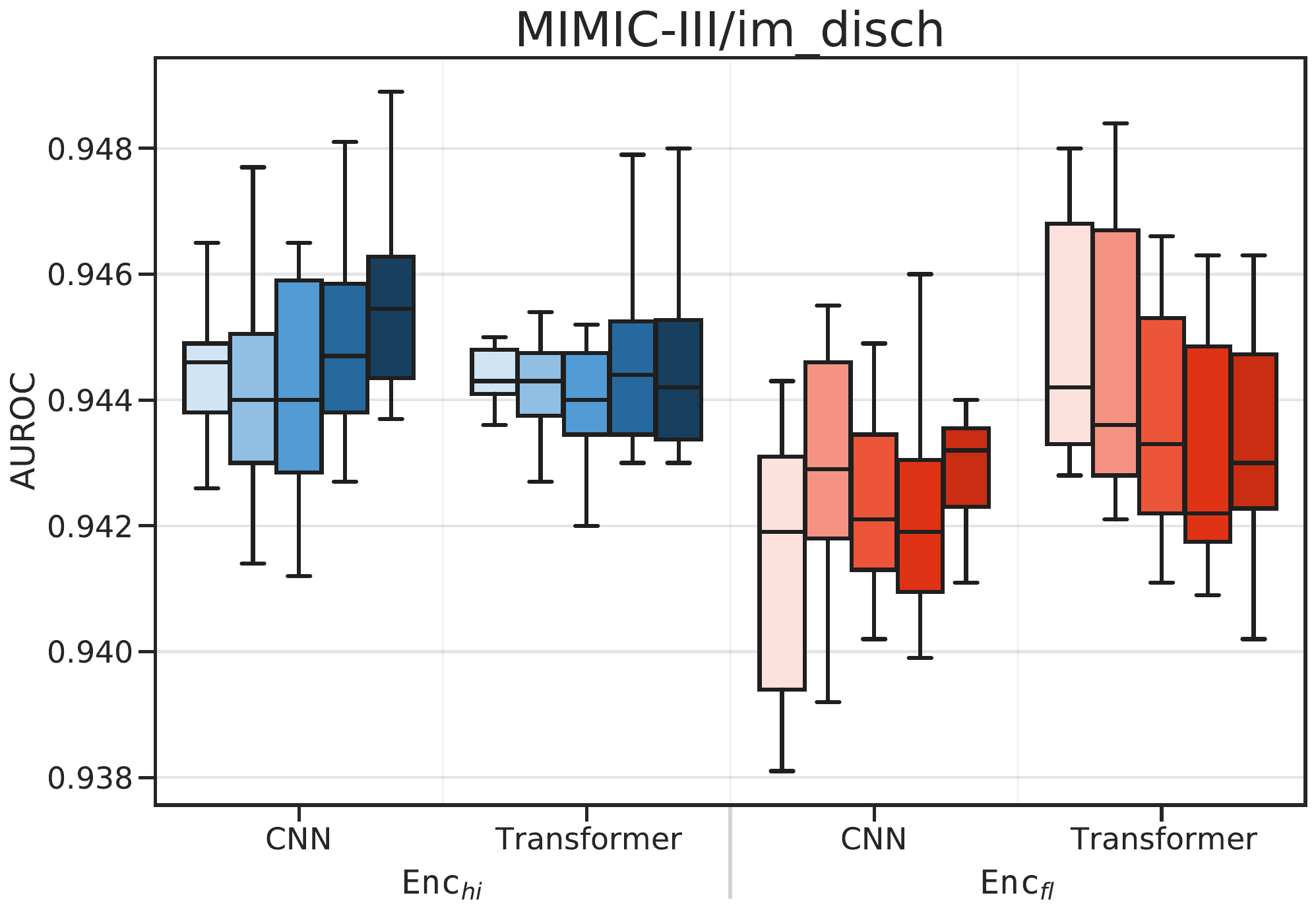}}%
    \subfigure[Length-Of-Stay for case of three days]{\label{}%
      \includegraphics[width=.45\textwidth]{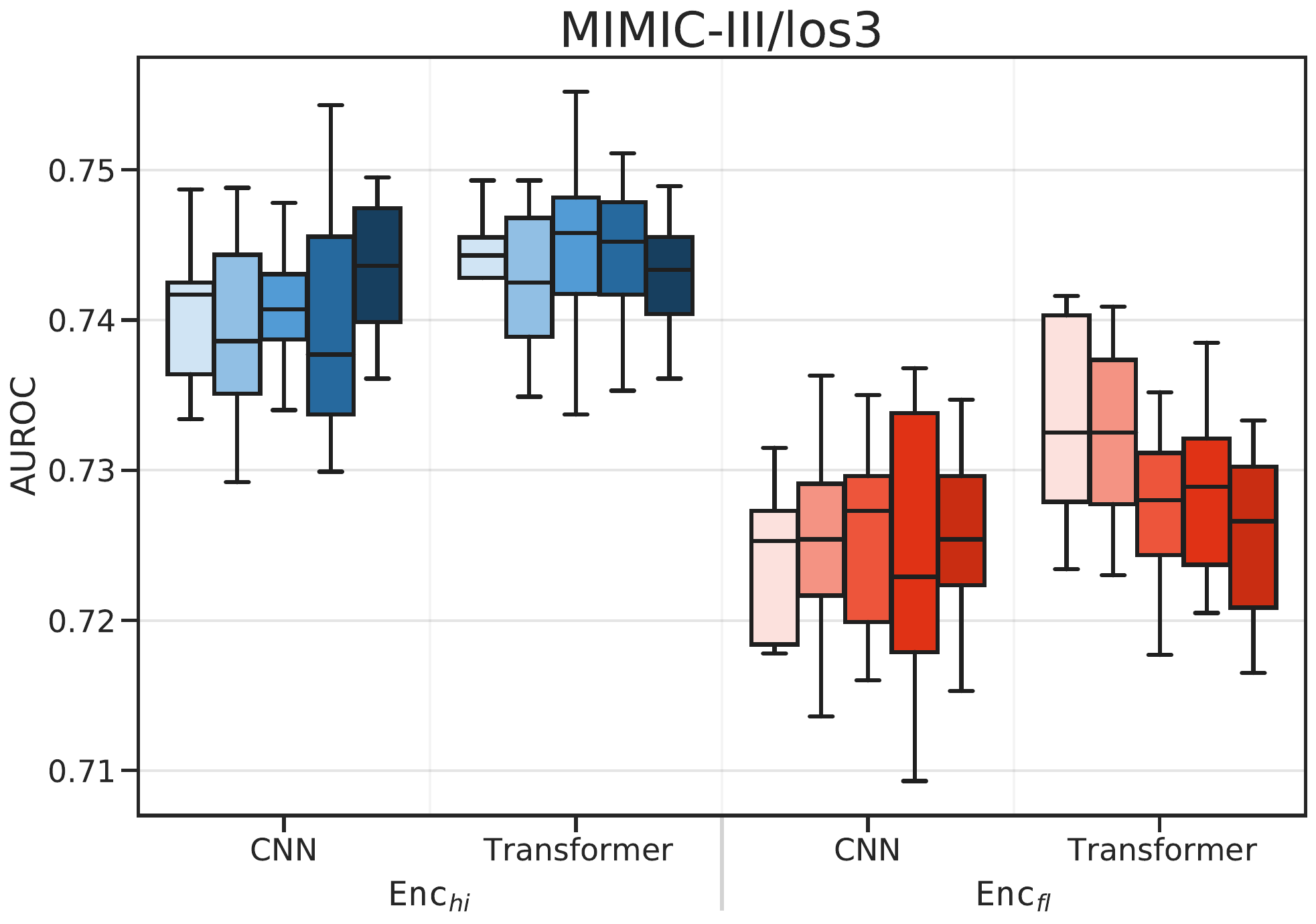}}
    \subfigure[Length-Of-Stay for case of seven days]{\label{}%
      \includegraphics[width=.45\textwidth]{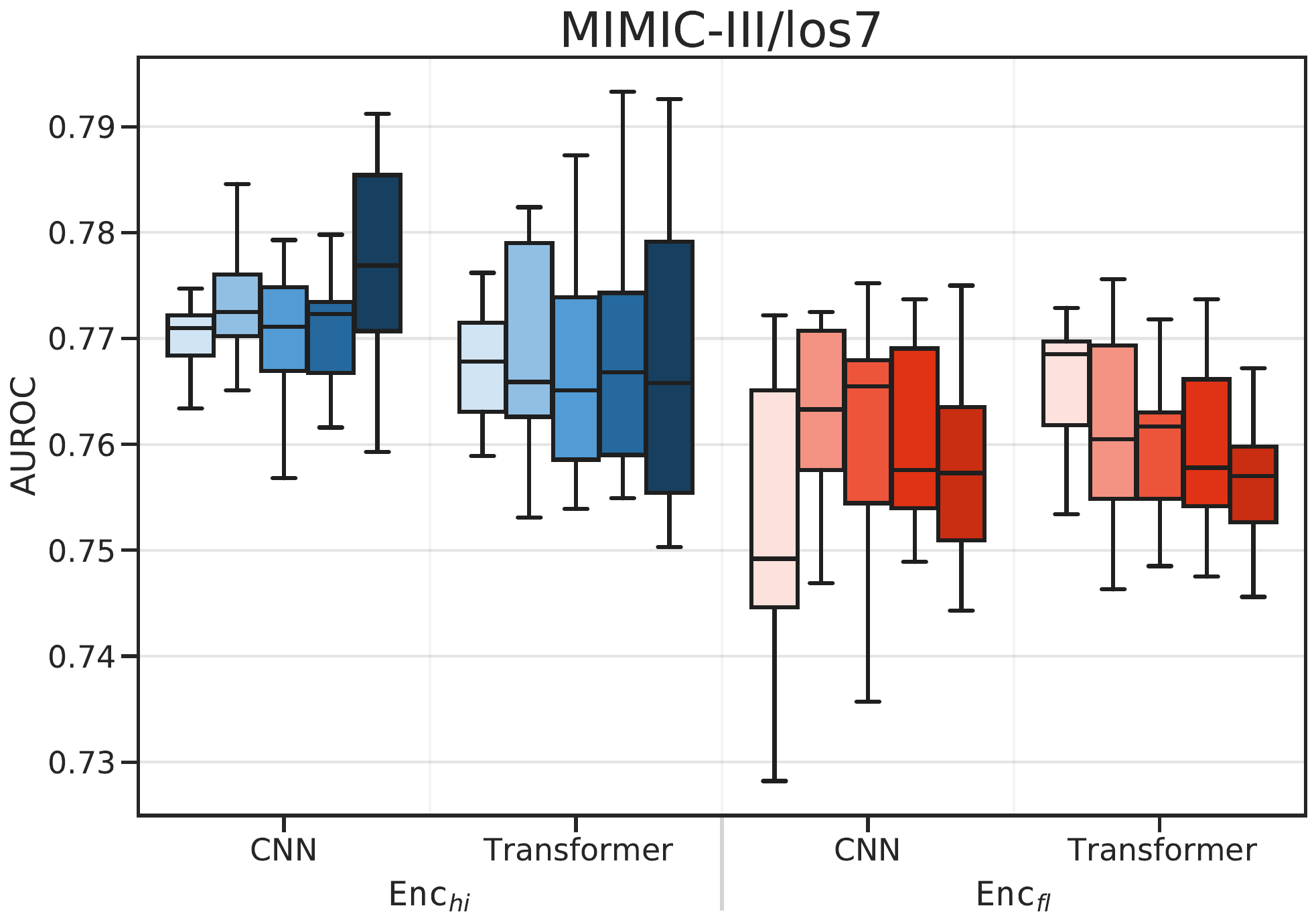}}%
    \subfigure[Mortality]{\label{}%
      \includegraphics[width=.45\textwidth]{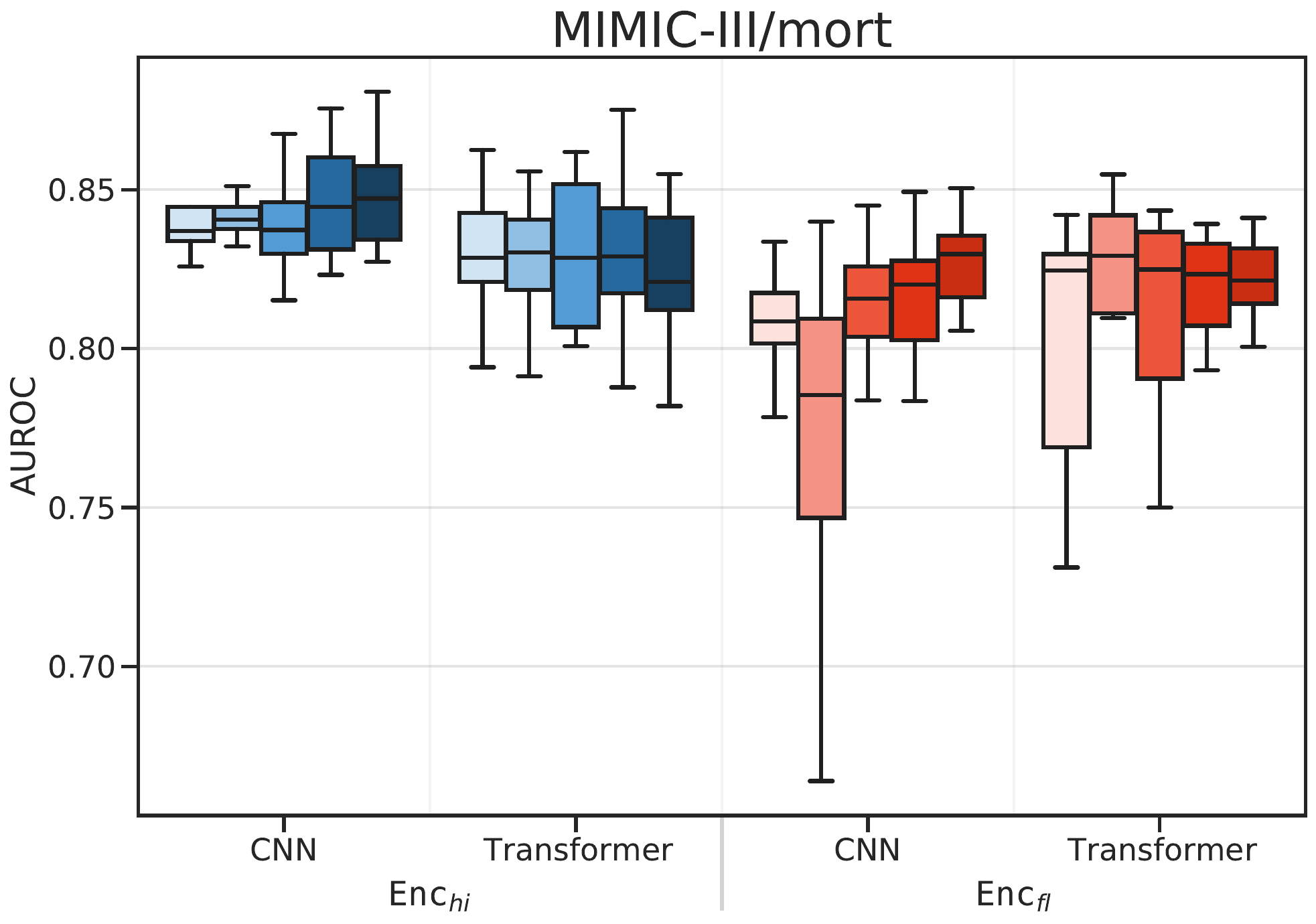}}
  }
\end{figure*}

\begin{figure*}[!h]
\floatconts
  {fig:prediction_eicu_latnet}
  {\caption{Prediction performances on eICU arranged by the latent dimension $l$}
  \label{fig:pred6Tasks-eicu-latent}}
  {%
    \subfigure[Diagnosis]{\label{}%
      \includegraphics[width=.45\textwidth]{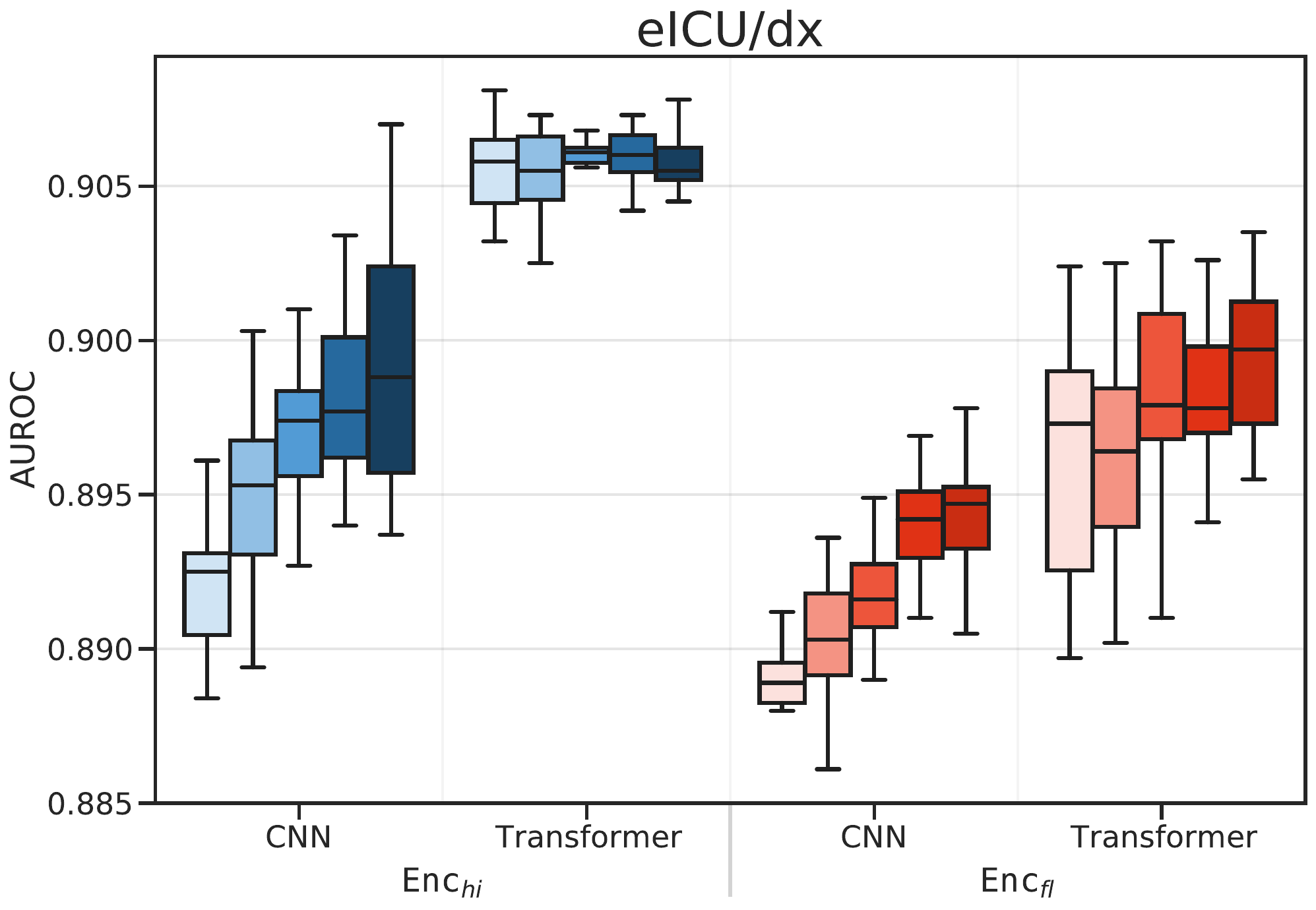}}%
    \subfigure[Final acuity]{\label{}%
      \includegraphics[width=.45\textwidth]{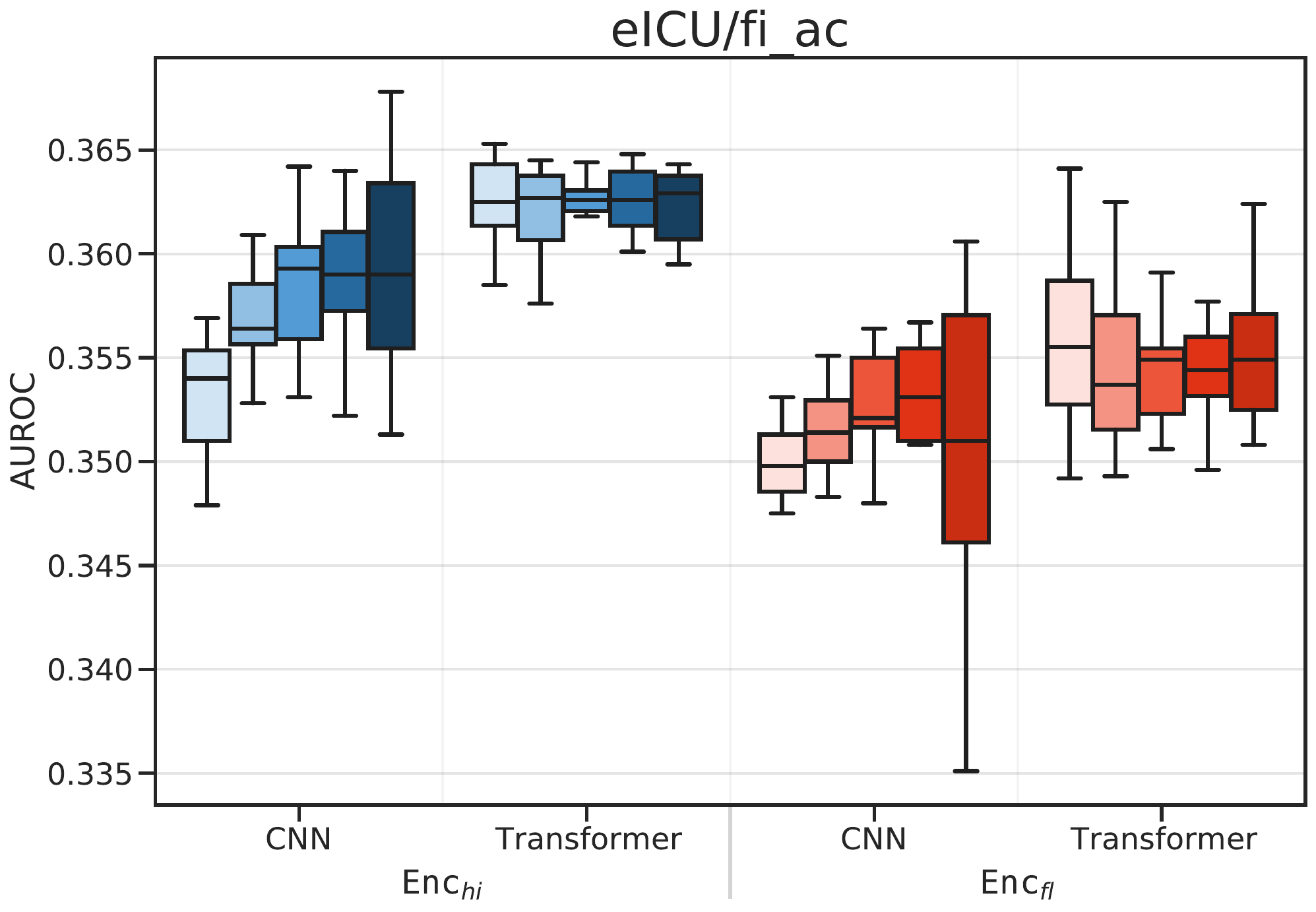}}
    \subfigure[Imminent Discharge]{\label{}%
      \includegraphics[width=.45\textwidth]{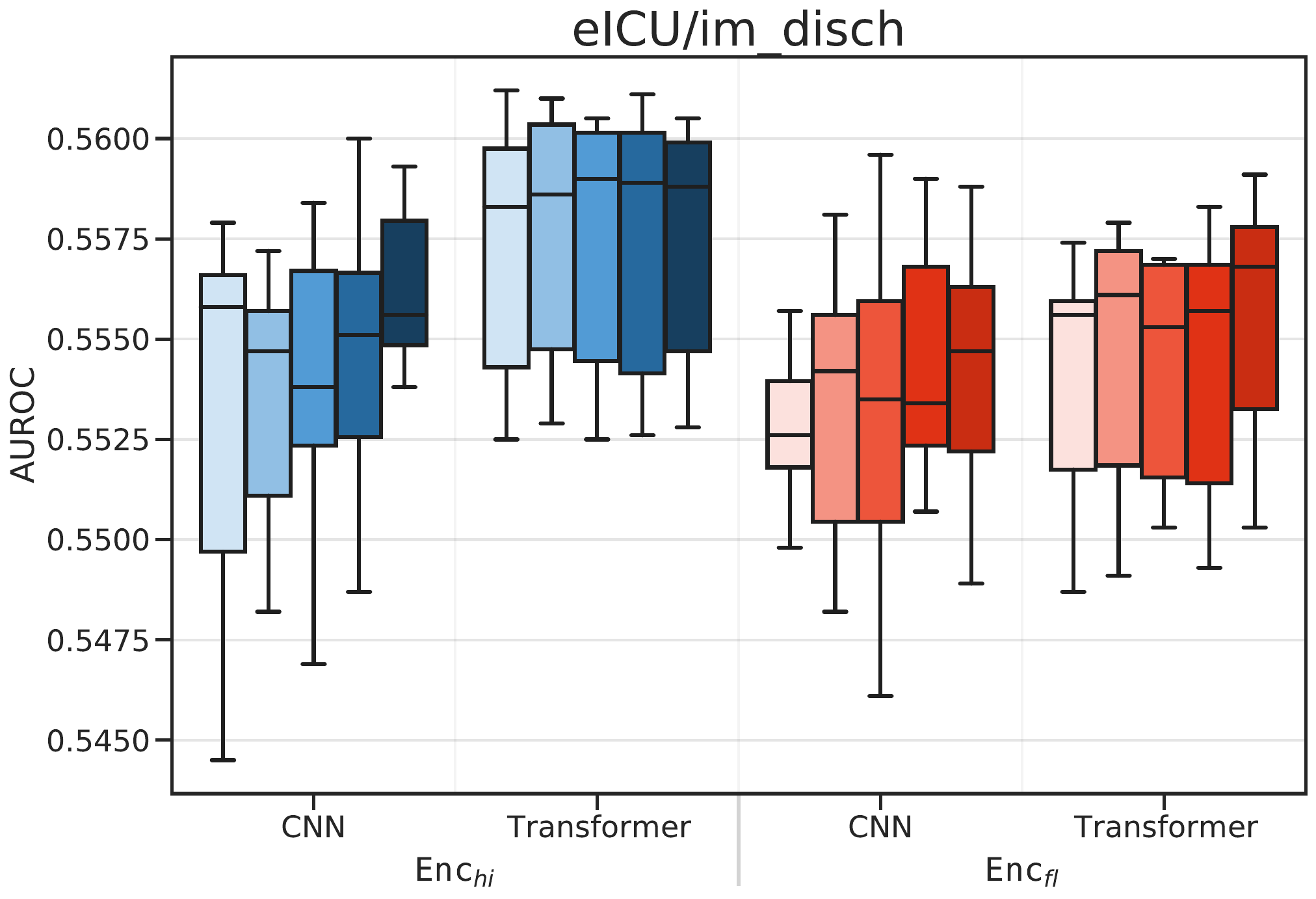}}%
    \subfigure[Length-Of-Stay for case of three days]{\label{}%
      \includegraphics[width=.45\textwidth]{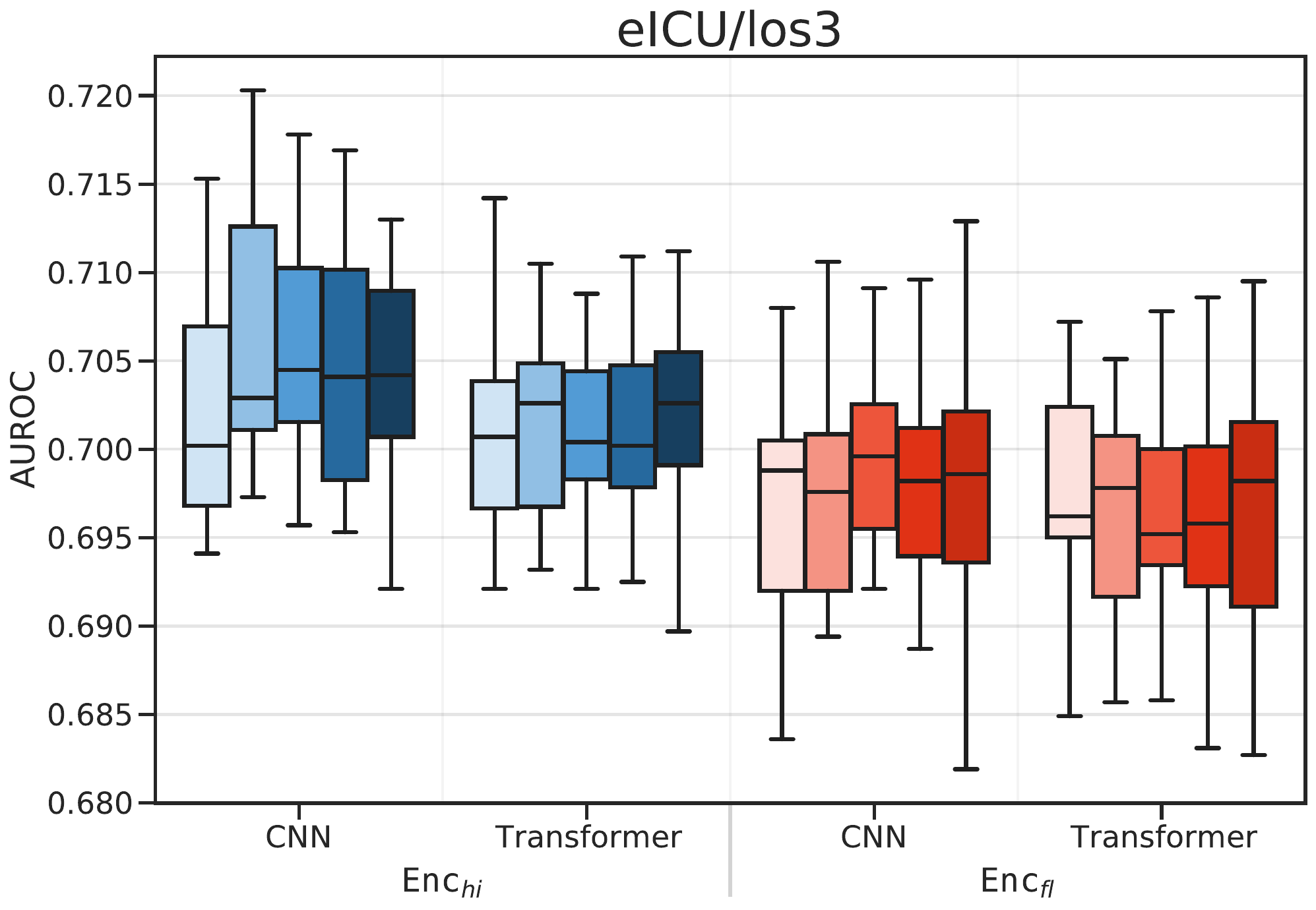}}
    \subfigure[Length-Of-Stay for case of seven days]{\label{}%
      \includegraphics[width=.45\textwidth]{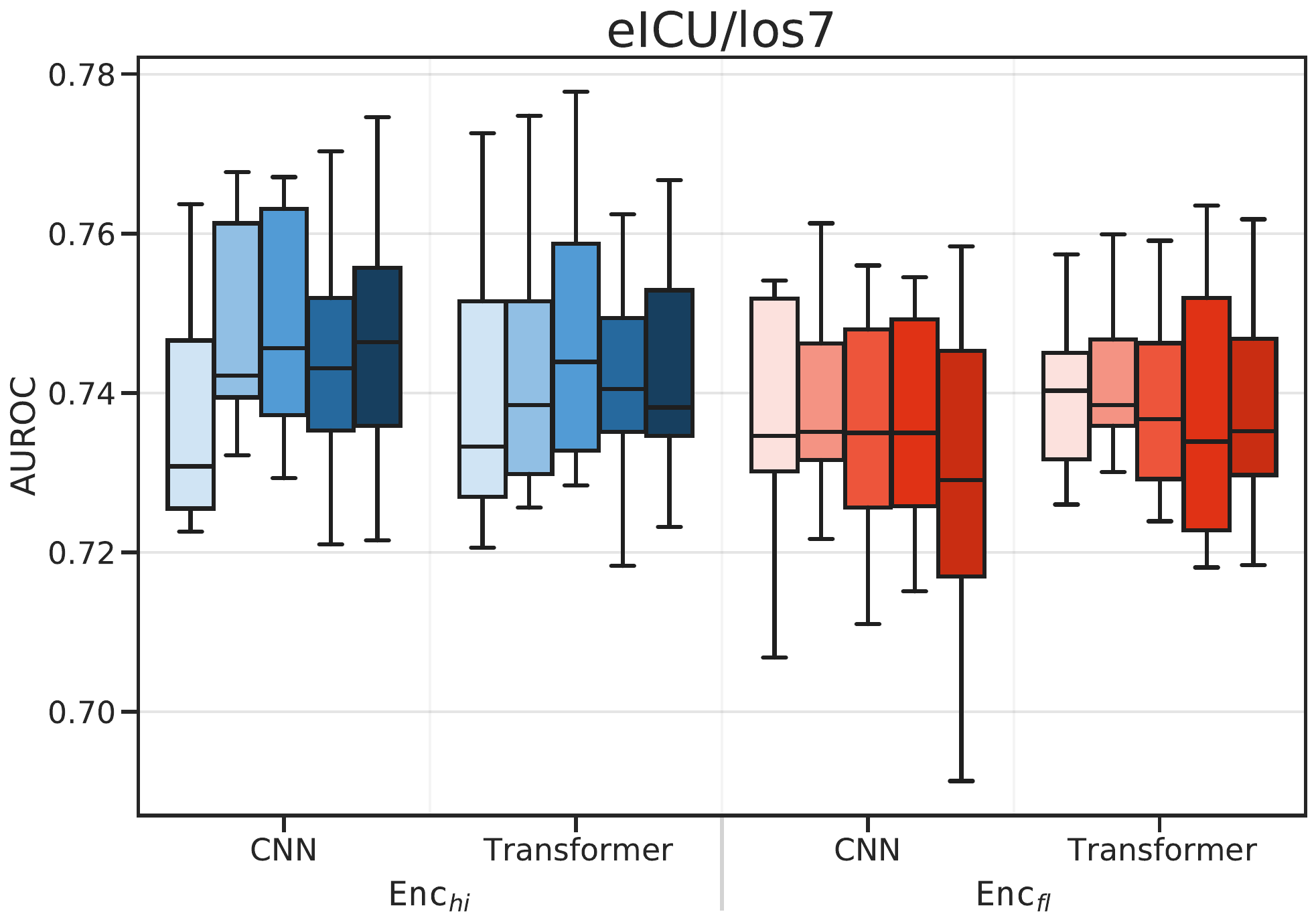}}%
    \subfigure[Mortality]{\label{}%
      \includegraphics[width=.45\textwidth]{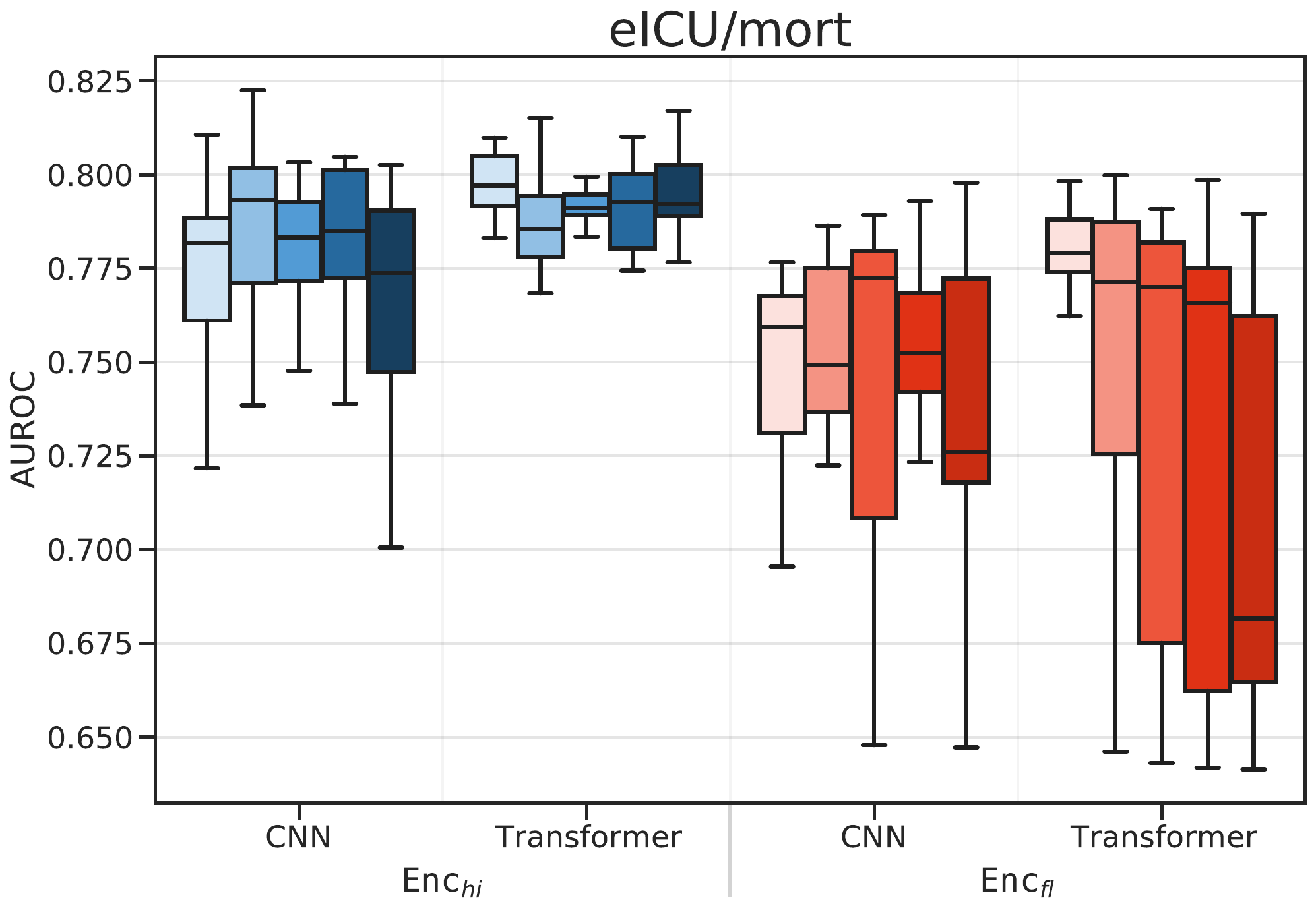}}
  }
\end{figure*}

\begin{figure*}[!h]
\floatconts
  {fig:prediction_eicu_temporal}
  {\caption{Prediction performances on eICU arranged by the temporal dimension $t$}
  \label{fig:pred6Tasks-eicu-temporal}}
  {%
    \subfigure[Diagnosis]{\label{}%
      \includegraphics[width=.45\textwidth]{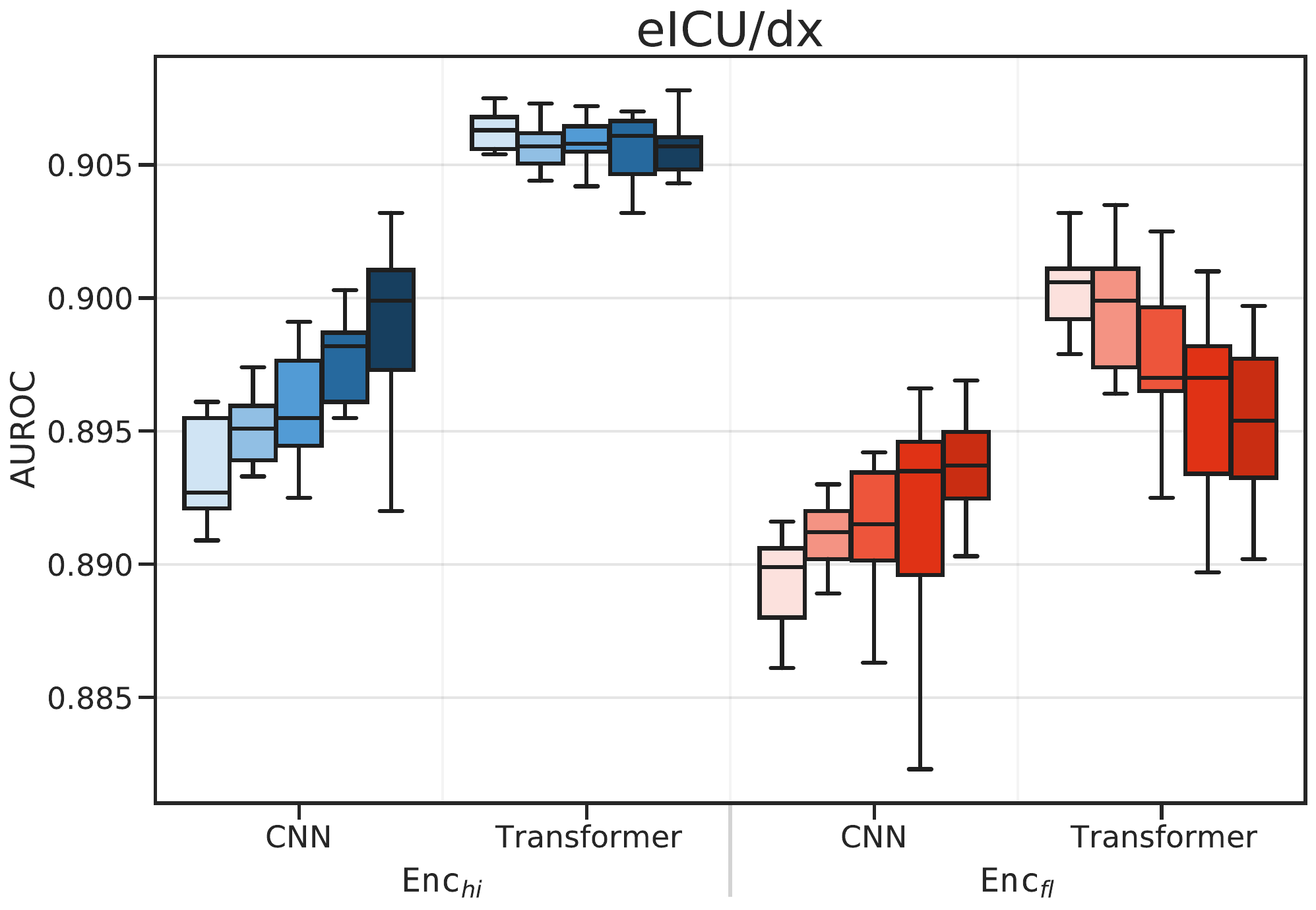}}%
    \subfigure[Final acuity]{\label{}%
      \includegraphics[width=.45\textwidth]{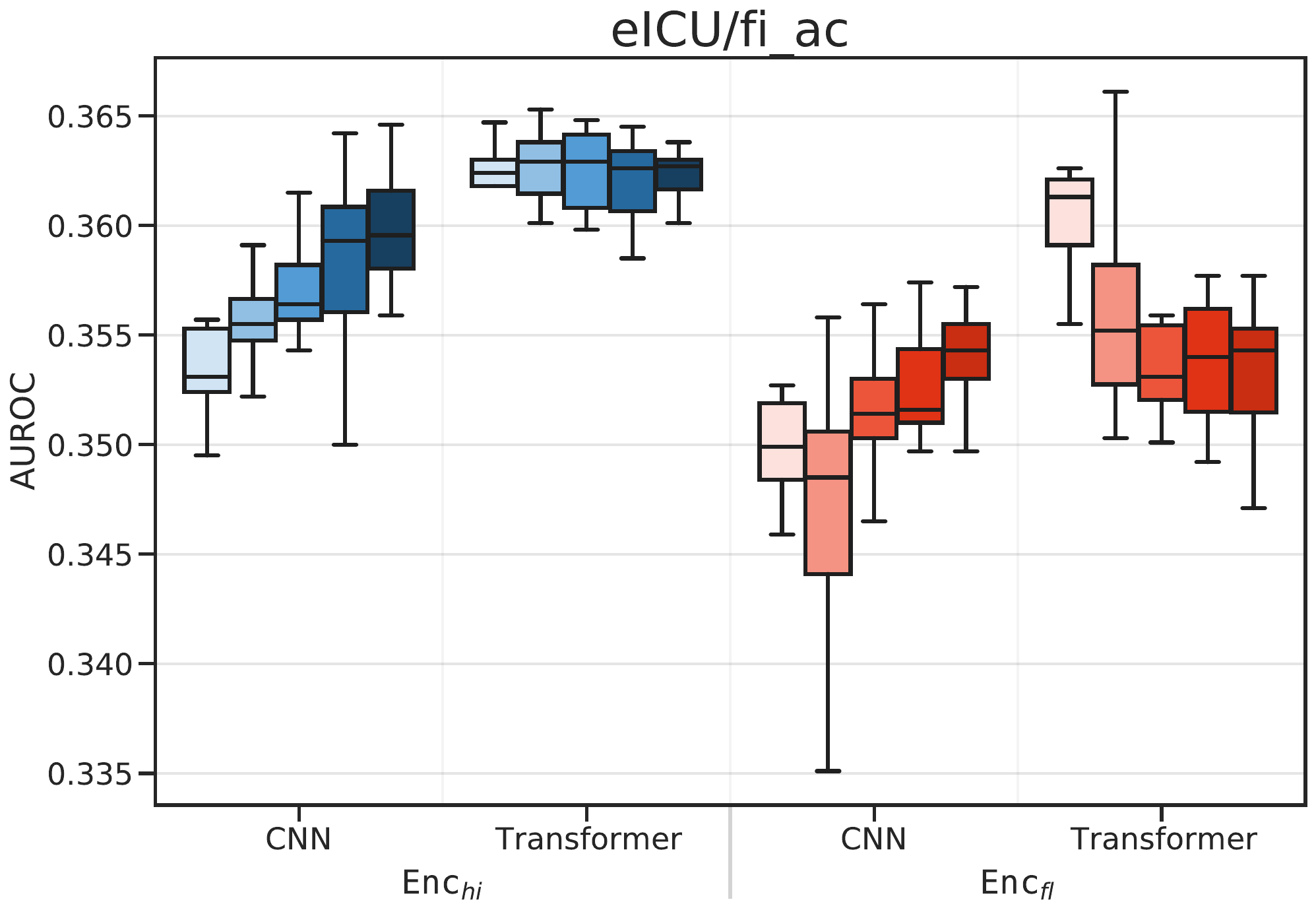}}
    \subfigure[Imminent Discharge]{\label{}%
      \includegraphics[width=.45\textwidth]{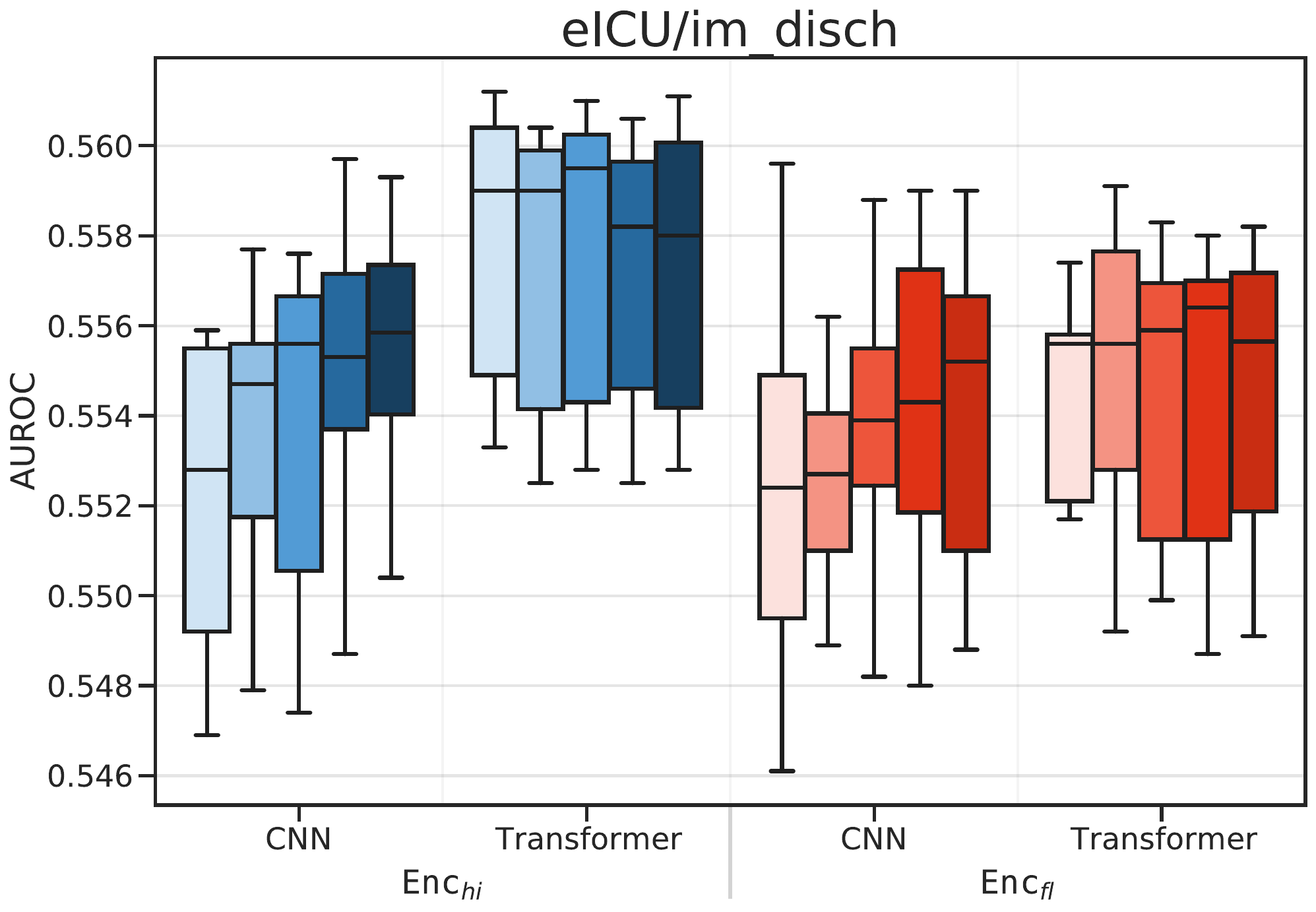}}%
    \subfigure[Length-Of-Stay for case of three days]{\label{}%
      \includegraphics[width=.45\textwidth]{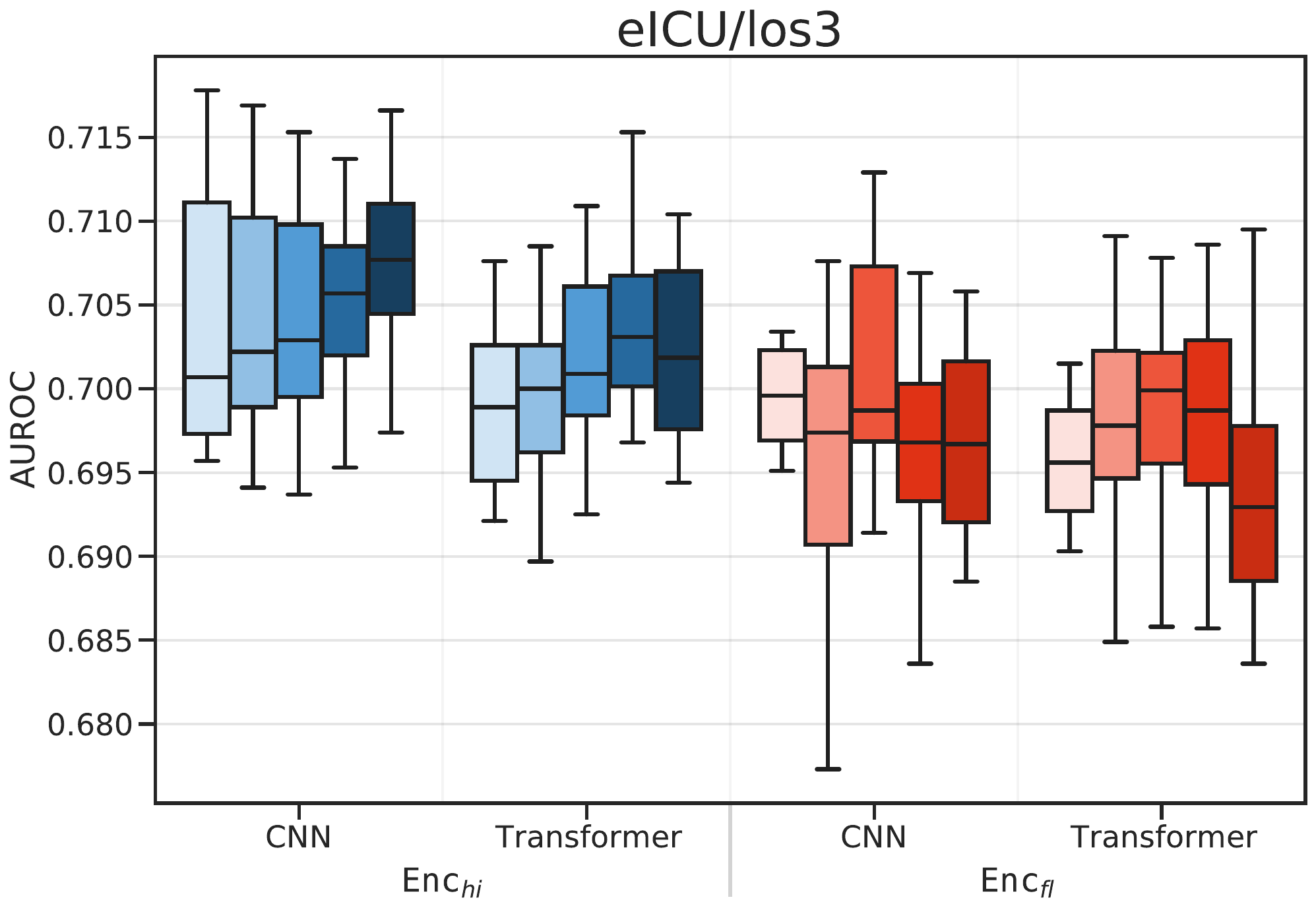}}
    \subfigure[Length-Of-Stay for case of seven days]{\label{}%
      \includegraphics[width=.45\textwidth]{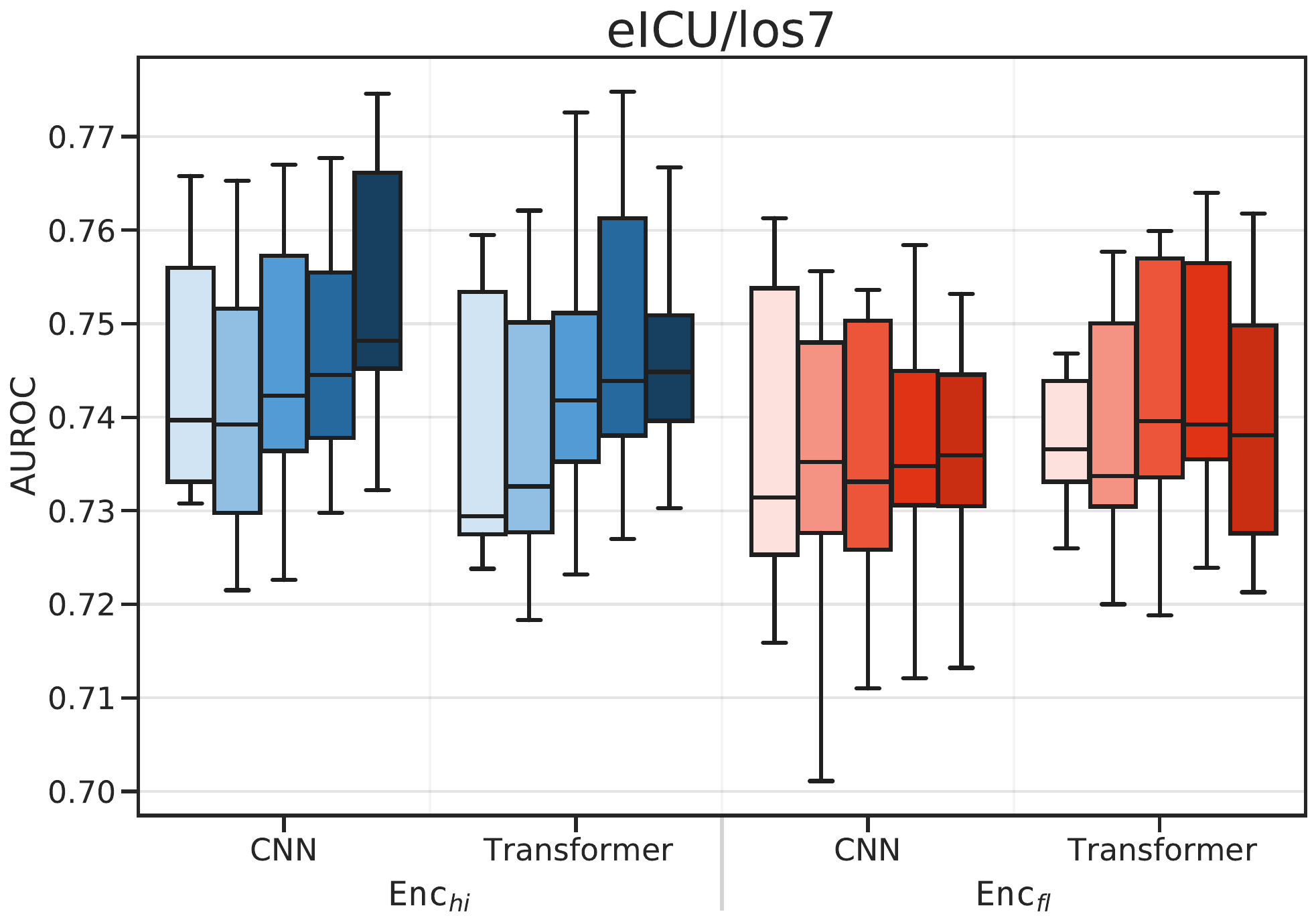}}%
    \subfigure[Mortality]{\label{}%
      \includegraphics[width=.45\textwidth]{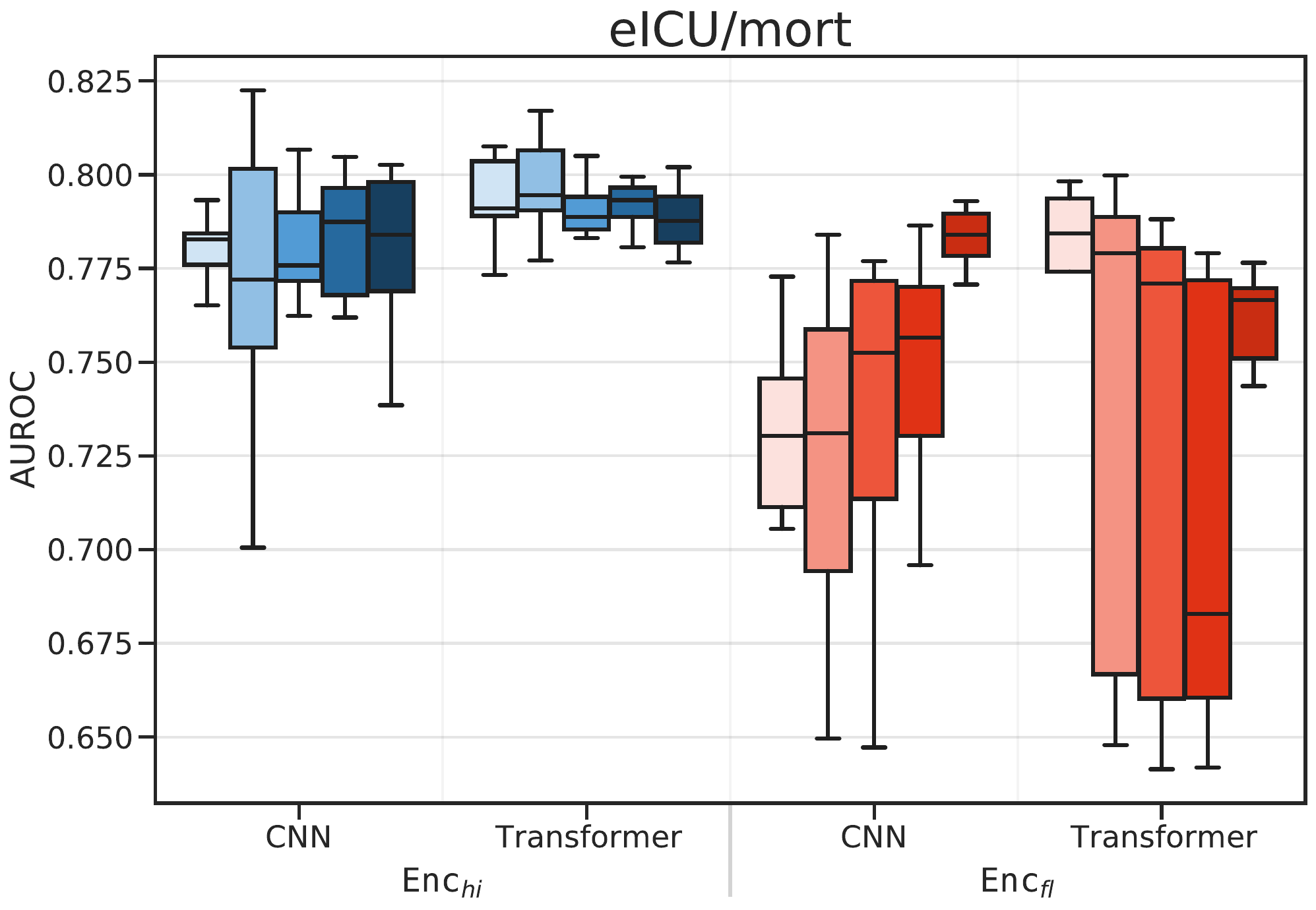}}
  }
\end{figure*}

\figureref{fig:pred6Tasks-mimic-latent}, \ref{fig:pred6Tasks-mimic-temporal}, \ref{fig:pred6Tasks-eicu-latent}, and
\ref{fig:pred6Tasks-eicu-temporal} show the prediction performances without averaging in a task-wise manner.
Prediction performances are arranged by the latent dimension $l$ in \figureref{fig:pred6Tasks-mimic-latent} and \ref{fig:pred6Tasks-eicu-latent}, and by the temporal dimension $t$ in \figureref{fig:pred6Tasks-mimic-temporal} and \ref{fig:pred6Tasks-eicu-temporal}.
We report these performances of each subtask on both MIMIC-III and eICU datasets.

\begin{figure}[!h]
\floatconts
  {fig:kde_plot_mimic}
  {\caption{Kernel Density Estimate (KDE) plot visualizing the distribution of the total number of tokens according to observation time in the MIMIC-III dataset.}
  \label{apd:tokenhist}
  }
  {%
    \subfigure[Total number of tokens of hierarchical input]{\label{1}%
      \includegraphics[width=\linewidth]{     
      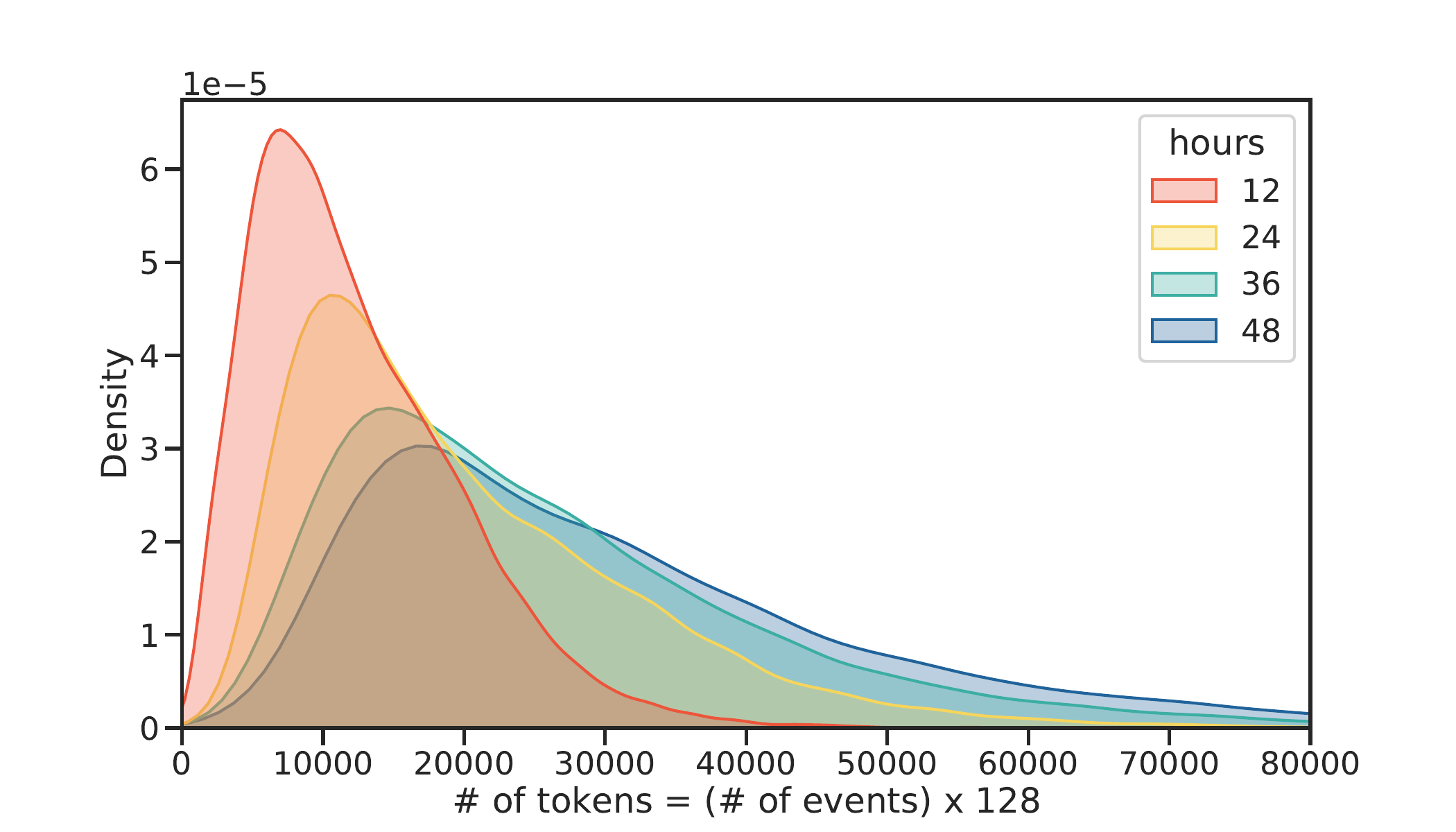}}
    \subfigure[Total number of tokens (excluding pad)]{\label{}%
      \includegraphics[width=\linewidth]{
      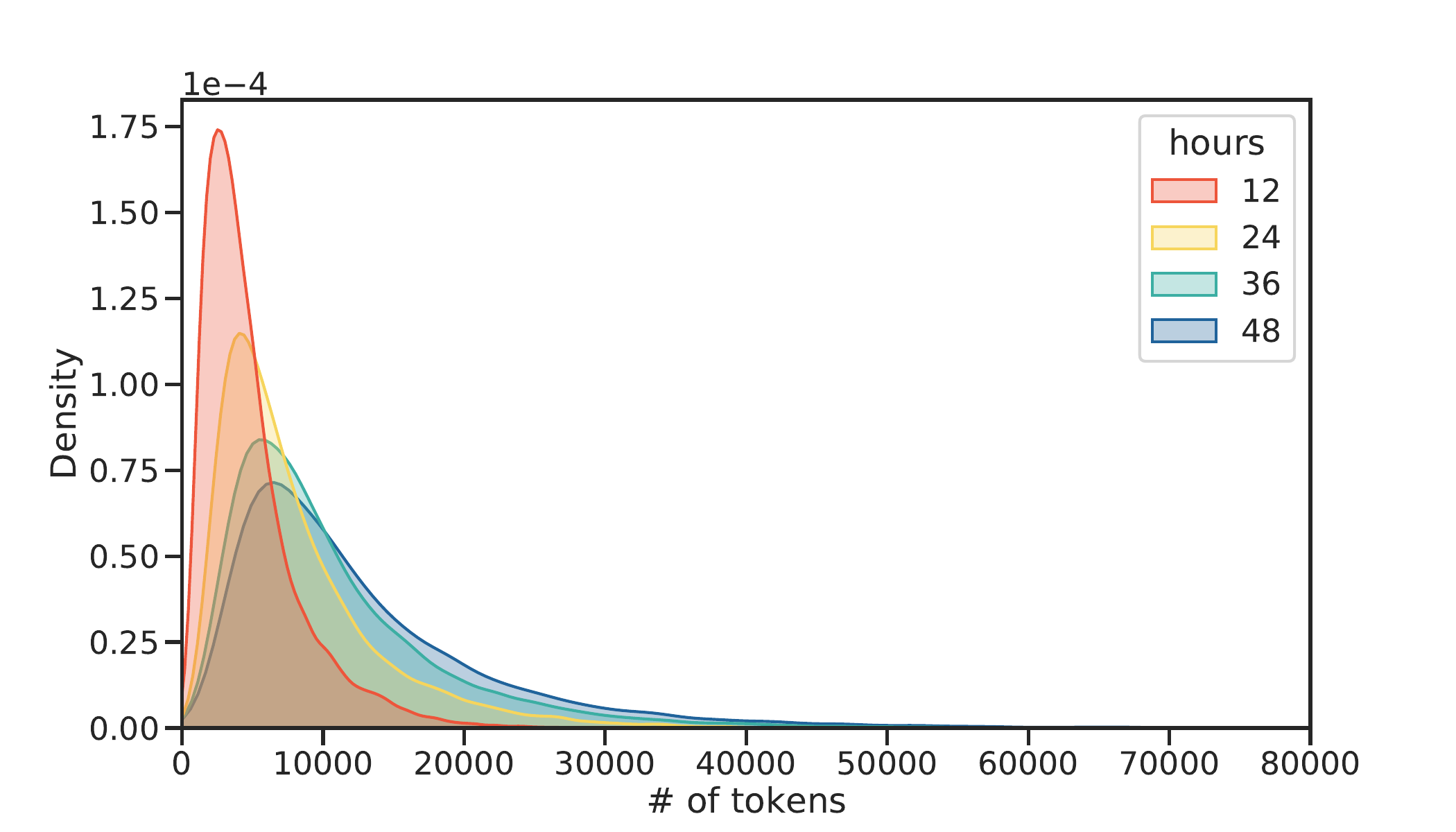}}
  }
\end{figure}

\begin{figure}[!h]
\floatconts
  {fig:pca_mimiciii}
  {\caption{Visualization of latent vector using PCA on MIMIC-III dataset. Both CNN-based and Transformer-based encoders use CNN-based decoders for reconstruction.}
  \label{apd:pca}}
  {%
    \subfigure[CNN-based (dx)]{\label{2}%
      \includegraphics[width=.5\linewidth]{     
      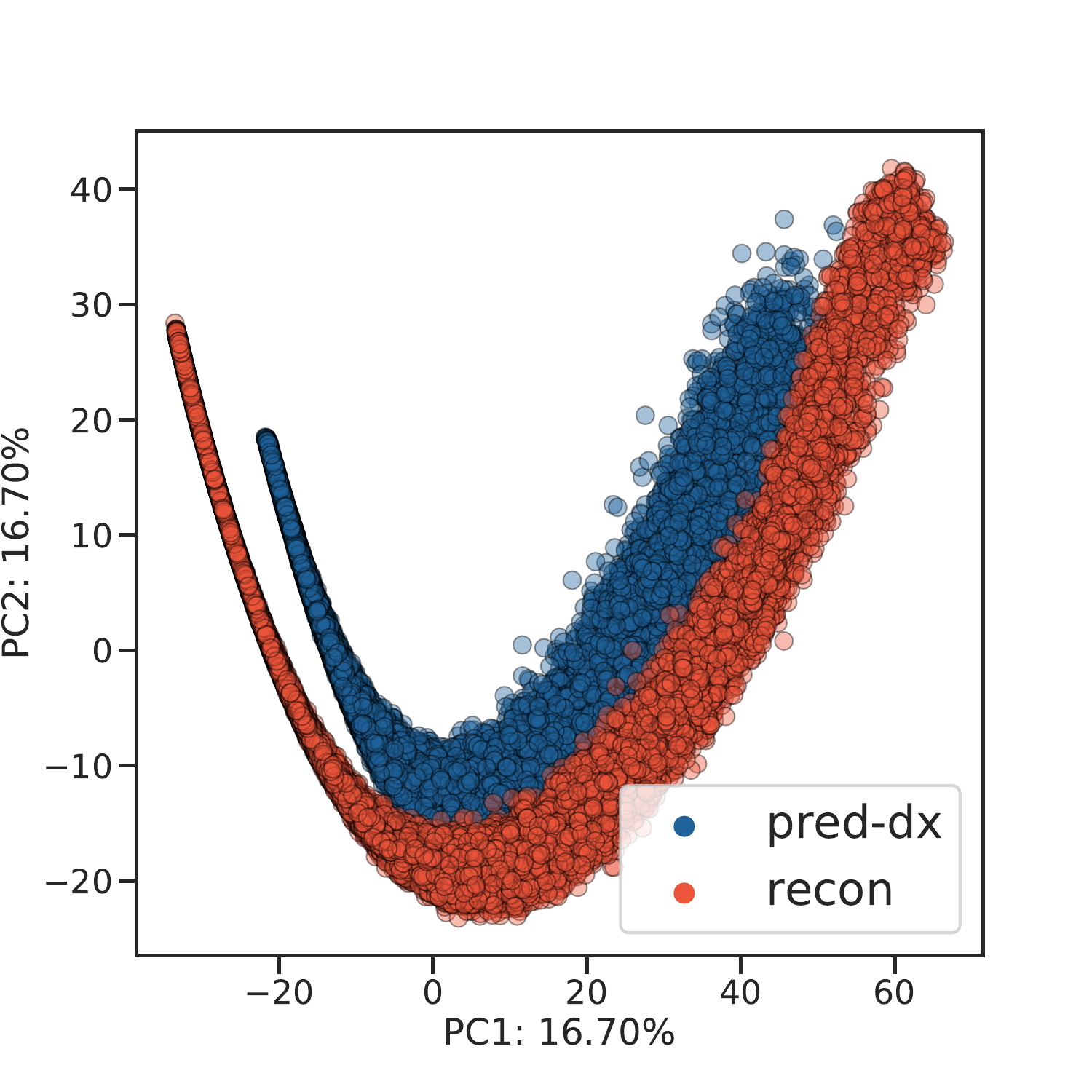}}%
    \subfigure[CNN-based (mort)]{\label{}%
      \includegraphics[width=.5\linewidth]{
      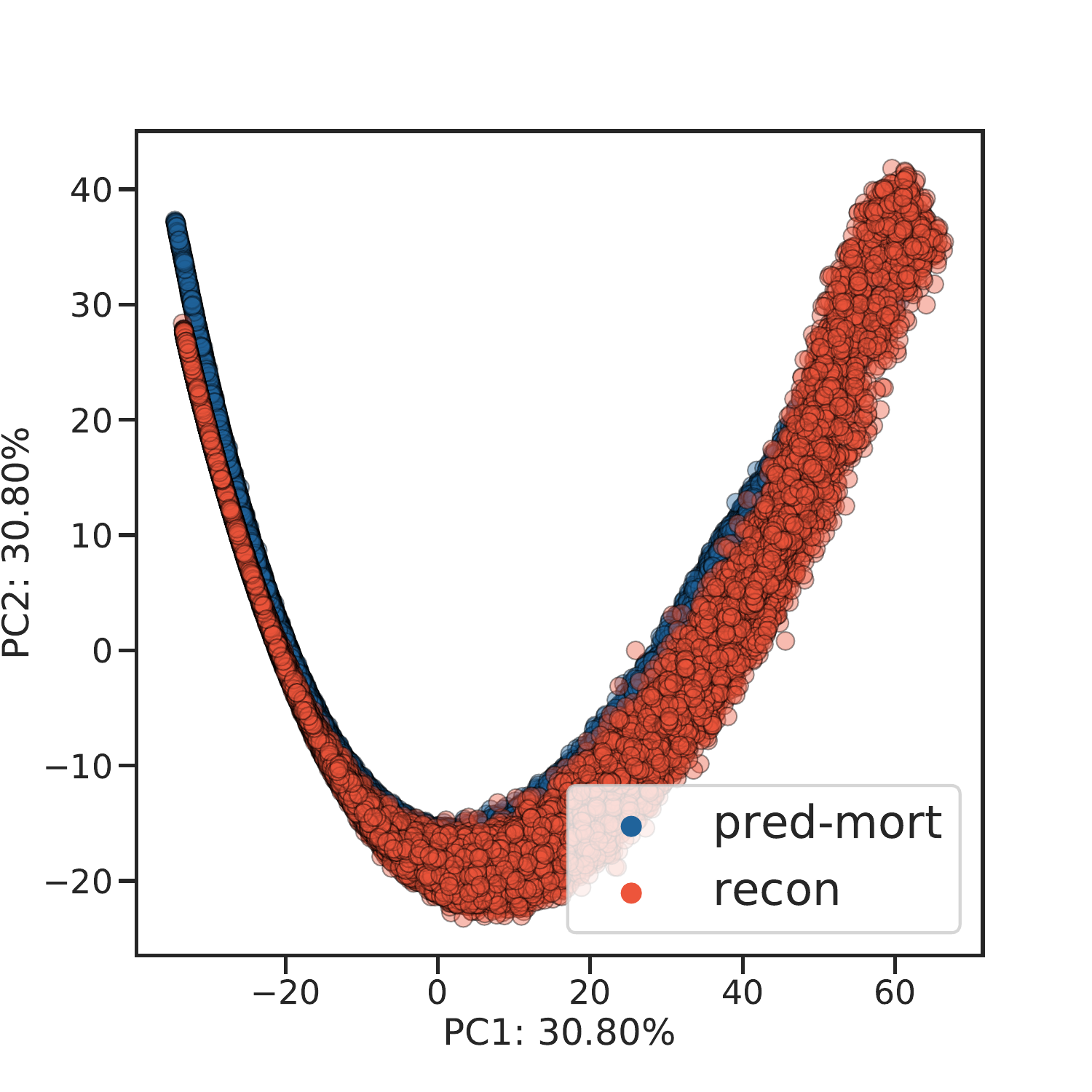}}
    \subfigure[Transformer-based (dx)]{\label{3}%
      \includegraphics[width=.5\linewidth]{     
      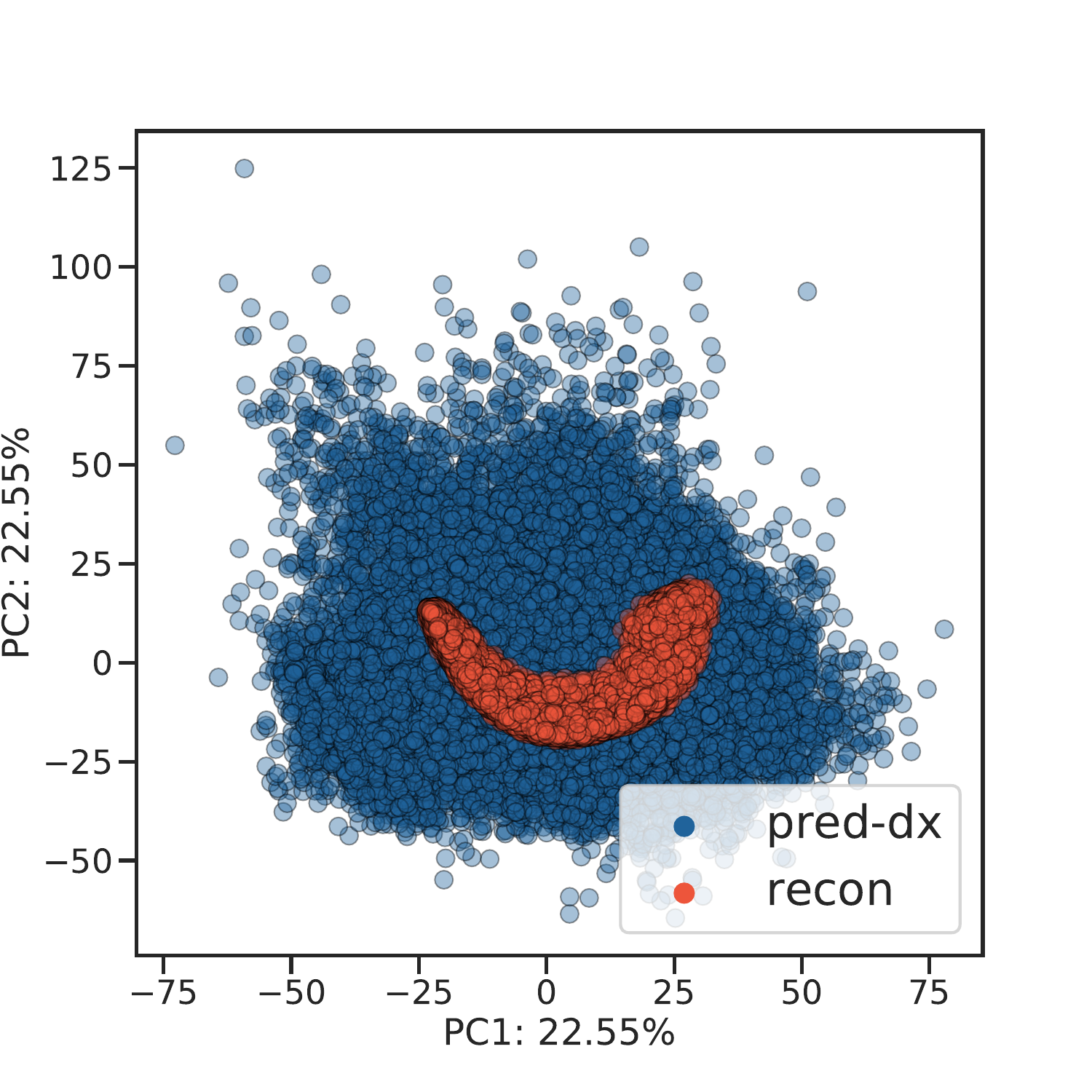}}%
    \subfigure[Transformer-based (mort)]{\label{}%
      \includegraphics[width=.5\linewidth]{
      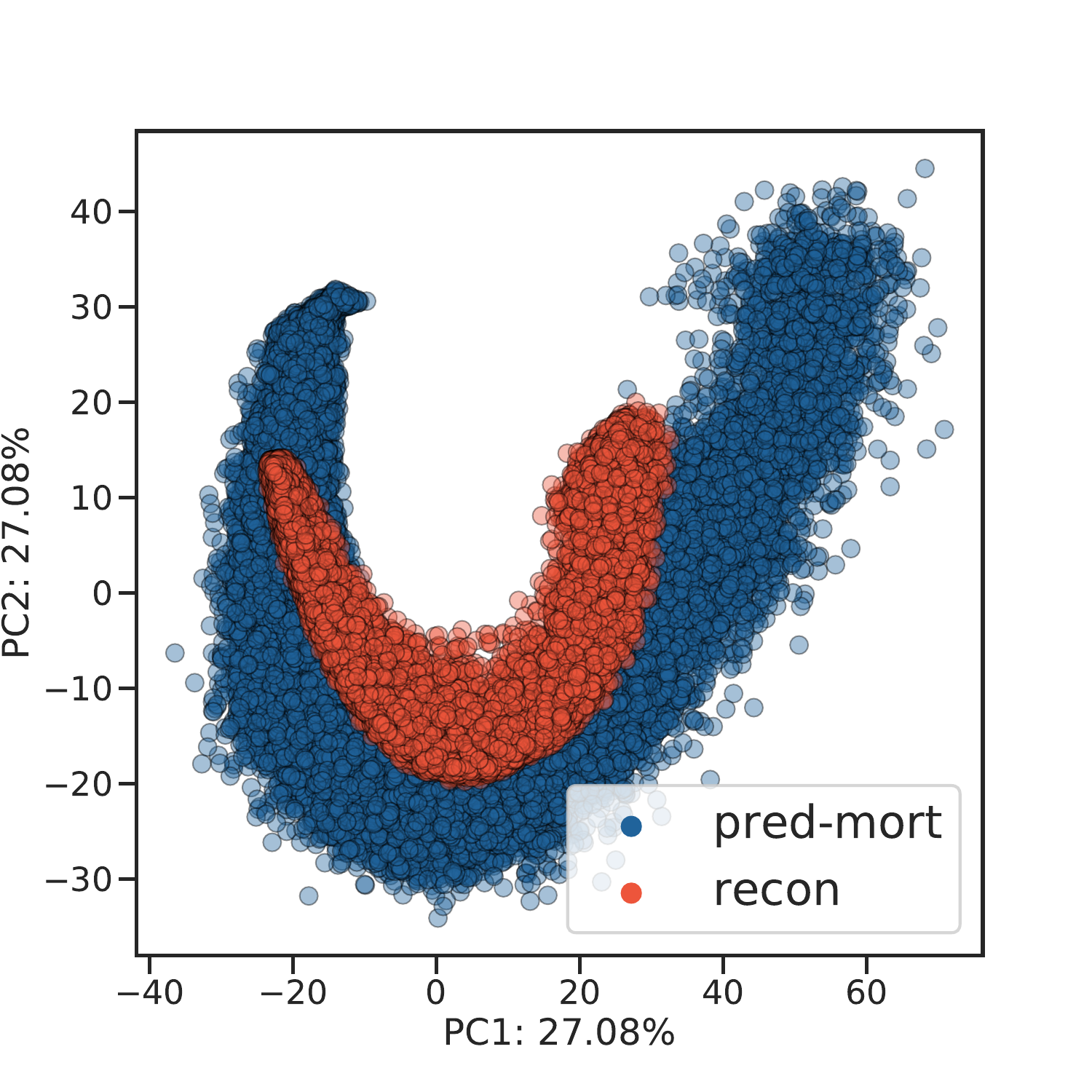}}
  }
\end{figure}

\end{document}